\documentclass[journal]{IEEEtran}
\usepackage{pgfplots}
\pgfplotsset{compat=1.14}

\usepackage{comment}
\usepackage{amsmath}
\usepackage[capitalise]{cleveref}
\usepackage{amssymb}
\usepackage{amsthm}
\usepackage{url}

\usepackage{tikz}
\usetikzlibrary{arrows,positioning,shapes.symbols,shapes.callouts,patterns}

\ifCLASSINFOpdf
\else
\fi
\usepackage[subrefformat=parens]{subcaption}

\def\R{\mathbb{R}}
\def\S{\mathcal{S}}
\def\Z{\mathbb{Z}}
\def\etal{{\it et al.\ }}

\renewcommand{\tilde}{\widetilde}
\renewcommand{\epsilon}{\varepsilon}

\begin{document}
%
\title{Three-Dimensional Fourier Scattering Transform and Classification of Hyperspectral Images}
%
%
%

\author{Ilya~Kavalerov,
        Weilin~Li,
        Wojciech~Czaja,
        and~Rama~Chellappa,~\IEEEmembership{Fellow,~IEEE}
\thanks{Manuscript Submitted February 13, 2020. This work was supported in part by the MURI from the Army Research Office under the Grant No. W911NF-17-1-0304. This is part of the collaboration between US DOD, UK MOD and UK Engineering and Physical Research Council (EPSRC) under the Multidisciplinary University Research Initiative. This work was also  supported in part by LTS through Maryland Procurement Office and the NSF DMS 1738003 grant. (\textit{Corresponding author: Ilya Kavalerov})}
\thanks{I. Kavalerov and R. Chellappa are with UMIACS, University of Maryland at College Park, College Park, MD 20742 USA (e-mail: ilyak@umiacs.umd.edu; rama@umiacs.umd.edu)}
\thanks{W. Li was with the Norbert Wiener Center, University of Maryland at College Park, College Park, MD 20742 USA. He is now with the Courant Institute of Mathematical Sciences, New York University, NY 10012 USA (e-mail: weilinli@cims.nyu.edu)}
\thanks{W. Czaja is with the Norbert Wiener Center, University of Maryland at College Park, College Park, MD 20742 USA (e-mail: wojtek@math.umd.edu)}
}

\markboth{Three-Dimensional Fourier Scattering Transform and Classification of Hyperspectral Images}%
{Kavalerov \MakeLowercase{\textit{et al.}}: Three-Dimensional Fourier Scattering Transform and Classification of Hyperspectral Images}

%



\maketitle

\begin{abstract}

Recent developments in machine learning and signal processing have resulted in many new techniques that are able to effectively capture the intrinsic yet complex properties of hyperspectral imagery. Tasks ranging from anomaly detection to classification can now be solved by taking advantage of very efficient algorithms which have their roots in representation theory and in computational approximation. Time-frequency methods are one example of such techniques. They provide means to analyze and extract the spectral content from data. On the other hand, hierarchical methods such as neural networks incorporate spatial information across scales and model multiple levels of dependencies between spectral features. Both of these approaches have recently been proven to provide significant advances in the spectral-spatial classification of hyperspectral imagery. The 3D Fourier scattering transform, which is introduced in this paper, is an amalgamation of time-frequency representations with neural network architectures. It leverages the benefits provided by the Short-Time Fourier Transform with the numerical efficiency of deep learning network structures. We test the proposed method on several standard hyperspectral datasets, and we present results that indicate that the 3D Fourier scattering transform is highly effective at representing spectral content when compared with other state-of-the-art spectral-spatial classification methods.
\end{abstract}


%
\IEEEpeerreviewmaketitle

\section{Introduction}


Hyperspectral image sensors routinely collect hundreds of bands of different wavelength channels of the surface of the Earth \cite{hsi_history}.
Due to the rapidly growing amount of available hyperspectral imagery (HSI)  \cite{NASA}, there is much interest in the development of algorithms that can take advantage of these resources for a wide range of applications, from anomaly detection to automatic classification. However, several characteristics of HSI data make these tasks challenging: the high dimensionality of the data, the low spatial resolution, the resulting unfavorable signal-to-noise ratio, and the fact that labeled data is scarce and typically not transferable across different domains. This emphasizes the continued need for development of more efficient processing algorithms.

One potential source of needed advancements is the field of machine learning. As recently noted in the review by He \etal \cite{He_review}, neural networks (NNs) and deep learning, having already achieved breakthroughs in the traditional image classification or segmentation tasks, are now gaining popularity in HSI applications \cite{acquarelli_spectral-spatial_2018,chen_deep_2016,deng_hyperspectral_2018,ma_hyperspectral_2015,liang_hyperspectral_2016}. These hierarchical networks built for feature extraction at multiple levels have a potential to produce highly informative data features, which cannot be achieved by manually-designed feature extractors. However, the cost of these improvements is the increase in computational complexity of learning algorithms, which stems from the fact that there is now a great number of parameters to train. This predicament is often resolved by combining the learning scheme with an appropriately chosen representation transform which maximizes the information content in the first layer, consecutively leading to a reduction of the time needed for training the algorithm. It is thus in this context that we note that time-frequency representations for HSI have recently proven to provide both meaningful and high quality results \cite{He_review,Bau,jia_gabor_2015}, especially when the filters are specifically designed for the HSI data \cite{he_discriminative_2017}. At the same time, these time-frequency representations are an underutilized tool when compared with the more popular wavelet based methods \cite{cao_integration_2017,franchi_deep_2016,jia_gabor_2016,shen_three-dimensional_2011,qian_hyperspectral_2013,tang2015hyperspectral}.

The {\it Fourier scattering transformation} (FST), introduced in \cite{czaja2017analysis,czaja2017rotationally}, can be viewed as a modern machine learning-inspired approach to time-frequency analysis. It unifies deep learning architectures with time-frequency generated filters in order to capture higher order correlations between different time-frequency coefficients. One notable difference from deep learning is that the FST uses fixed filters instead of adaptable or learned ones such as those in NNs. However, the advantage is that the FST does not require computationally expensive training and enjoys theoretical guarantees that NNs lack \cite{czaja2017analysis}. In particular, the FST is invariant under small diffeomorphic nonlinearities or perturbations.

The {\it three-dimensional Fourier scattering transform} (3D FST) algorithm, which is introduced in this paper, is inspired by the mathematical Fourier scattering transformation for square integrable functions on a continuum. The novelty of this algorithm is the deviation from the original continuous time one-dimensional setting designated for traditional function approximation. Instead, here we design and employ three dimensional time-frequency generated filters to deal simultaneously with the discrete spatial and spectral aspects of the data. Furthermore, we implement these filters in a purpose-built architecture which takes advantage of recent developments in convolutional NNs. When used on HSI data, it provides a multi-layer spectral-spatial decomposition. We argue that the spectra generated by standard material classes are more discriminable in the time-frequency domain when compared to other representations. We note that this argument was also made in \cite{he_discriminative_2017}, which showed that decomposing the signal using time-frequency filters provides informative features for HSI data. However, the 3D FST further refines this idea by integrating together the spectral and spatial information in a multi-layer setting, where deeper layers are able to capture more complex features. 

We demonstrate that the 3D FST provides state-of-the-art performance on HSI data.
We compare to results with neural network and wavelet scattering based methods \cite{tang2015hyperspectral,eap,dffn}.
One notable method that we compare with is the {\it three-dimensional wavelet scattering transform} (3D WST). The {\it wavelet scattering transform} (WST) was originally developed by Mallat \cite{mallat2012group} and the 3D WST was applied to HSI classification in \cite{tang2015hyperspectral}. Our results show that time-frequency Fourier features are more suitable than time-scale wavelet features for HSI discrimination and classification purposes. Among others, we obtain state of the art results on Indian Pines at 10\% and 5\% of training data, and on Pavia University at 1\% and 0.5\%.
We also provide an open source implementation of all code used for our algorithms and experiments.\footnote{\url{https://github.com/ilyakava/pyfst}}




The rest of this paper is organized as follows.
Section \ref{sec:prev_work} reviews related work on using neural networks, wavelets, and time-frequency bases for feature extraction and the classification of HSI data.
Section \ref{sec:scattering} provides background information on scattering transforms. 
\cref{sec:methodology} defines the 3D FST and how it provides a joint spectral-spatial representation suited for HSI data.
\cref{sec:experiments} introduces the datasets that 3D FST is evaluated on, explains the parameter choices in the 3D FST, and discusses the results on these datasets while comparing them to other competing methods.
\cref{sec:conc} concludes the paper.


\section{Background}

\subsection{Previous Work}

\label{sec:prev_work}

Not surprisingly, many methods available in the literature have concentrated solely on analyzing the spectrum content for the classification of HSI data.
More recently, to improve classification performance, spectral-spatial techniques which better exploit the properties of HSI data have become popular. Our review of deep learning, neural network, wavelet, and time-frequency methods can be roughly split into four categories: 1) purely spectral techniques 2) spectral methods that incorporate spatial pre/post processing, 3) purely spatial methods that may include spectral pre/post processing, and 4) those that integrate spectral-spatial information at once.

\begin{enumerate}
\item 
Pixelwise methods that extract wavelet, time-frequency, and neural network features solely in the spectral domain have been developed to address the challenges in HSI classification
\cite{guo_guided_2018,hu_deep_2015,zhong_learning_2017,hsu_evaluating_2016,gao_convolution_2018,our_spie}.
The neural network (NN) family of methods iteratively composes layers of matrix multiplications or linear convolutions with a pointwise non-linear function. The weights in these matrices, or convolution filters, are adapted using backpropagation during an initial training phase.
NNs consisting of 1D convolutions with spectra \cite{hu_deep_2015}, and 2D convolutions of reshaped 1D spectral vectors \cite{gao_convolution_2018}, as well as other Deep Belief Networks (DBN) \cite{zhong_learning_2017} have been evaluated.
Another family of methods are the wavelet and time-frequency methods which use a pre-determined basis to extract edge-like features.
1D Morlet wavelet features with trainable scale and translation parameters have been extracted and input to a 2 layer NN \cite{hsu_evaluating_2016,hsu_classification_2006}.

A compromise between the learned and potentially deep and complex neural networks, and classical wavelet features are scattering transforms, which are particular types of operators introduced by S. Mallat \cite{mallat2012group}. The coefficients are computed with a hierarchical network structure that captures several types of invariances in data.
1D Fourier scattering features coupled with an SVM have also proved to be effective, outperforming 1D wavelet scattering features \cite{our_spie}.
These purely spectral methods improve upon the performance of simpler machine learning algorithms, like the application of SVMs on the 1D spectra of each pixel, but ignore the significant spatial structure present in HSI data.

\item
A variety of methods have successfully used spatial information in the pre-processing steps (more rarely in post-processing, as well) to improve classification performance while still focusing on the spectral aspects of the data \cite{acquarelli_spectral-spatial_2018,ma_hyperspectral_2015,lee_going_2017,mughees_hyperspectral_2017,ashitha_classification_2014}.
Ashitha \etal \cite{ashitha_classification_2014} classify 1D wavelet scattering features with an SVM after smoothing each channel of the HSI with 2D Gaussian filters.
Sandwiching a spectral NN between two 2D Gaussian blur layers with trainable variance greatly improves performance and remains one of the most competitive HSI methods \cite{ma_hyperspectral_2015}.
Lee \etal followed up on this work with a deep spectral NN with residual connections following a single 3D filter layer \cite{lee_going_2017}.
Acquarelli \etal made changes instead to the training process and included a spatial term in the regularizer of a purely spectral 1D convolutional neural network (CNN) \cite{acquarelli_spectral-spatial_2018}.

\item
Spatial methods that include some spectral pre-processing have also been applied to HSI data
\cite{chen_deep_2016,deng_hyperspectral_2018,liang_hyperspectral_2016,franchi_deep_2016,yang_hyperspectral_2018,our_spie}.
Classical 2D CNNs and Recurrent CNNs (RCNN) have been used on each channel of HSI input independently \cite{yang_hyperspectral_2018}.
Also popular has been using PCA or other dimensionality reduction methods to reduce the number of channels in the the HSI before using a 2D CNN \cite{chen_deep_2016,liang_hyperspectral_2016,dffn}, or 2D wavelet scattering \cite{franchi_deep_2016}.
After using PCA to reduce the dimension, {\it Attribute Profiles} can be used to extract spatial features on each principle component and expand the dimension by a small amount \cite{He_review}. These features capture morphological properties like area and shape per principle component, and the concatenation of many such 2D feature images can then be fed into a 2D CNN \cite{eap}.
Our previous work performed competitively using 1D Fourier scattering preprocessing followed by 2D wavelet scattering \cite{our_spie}.
Recently Deng \etal used a new CapsNet NN architecture \cite{capsnet} with 2D filters on each channel independently to achieve competitive results.

\item
Integrated spectral-spatial methods which combine information from spectral signatures and spatial neighborhoods simultaneously are also common.
Some papers consider sequences of 1D spectra \cite{guo_guided_2018}, for example in a variety of NNs known as Long Short Term Memory (LSTM), or convert the HSI cube to a matrix and use standard 2D methods \cite{haridas_comparative_2015}. However, by far the most popular methods involve building 3D filters \cite{chen_deep_2016,Bau,jia_gabor_2015,he_discriminative_2017,cao_integration_2017,shen_three-dimensional_2011,qian_hyperspectral_2013,tang2015hyperspectral,yang_hyperspectral_2018}.
3D convolutional layers in NNs, CNNs, and RCNNs, both shallow and deep have been evaluated \cite{chen_deep_2016,yang_hyperspectral_2018}.
But the lack of training data challenges models with many learnable parameters, and these networks struggled in comparison to methods with predetermined filters such as the
3D Gabor wavelets that Shen and Jia \etal used to extract features \cite{jia_gabor_2015,jia_gabor_2016,shen_three-dimensional_2011}, and classify with a variety of algorithms, for instance a sparse representation based classification (3D WT+SRC) \cite{shen_three-dimensional_2011}.
Bau \etal \cite{Bau} used the real part of 3D Gabor filters sampled densely in the time-frequency domain to get features used with a Mahalanobis distance classifier.
He \etal \cite{he_discriminative_2017} decomposed the same filter into 8 subfilters, using only 3 to construct a discriminative low-rank Gabor mother filter (DLRGF) used to extract features, a hand designed feature which proved to be very competitive with a least squares based classifier (3D DLRGF+LS).
Qian and Cao \etal \cite{qian_hyperspectral_2013,cao_integration_2017} used a Haar 3D wavelet filter bank (3D DWT-FB) and discrete wavelet transform (3D DWT) with various classifiers.
Tang \etal \cite{tang2015hyperspectral} proposed a 3D Gabor wavelet scattering approach to extract features (3D WST),
which decomposes the HSI across multiple wavelet scales and orientations and uses local averaging to keep class labels consistent in neighborhoods, and classified with a radial basis function SVM (3D WST+RBF-SVM).

\end{enumerate}

The paradigm that we have ordered our review by has been the degree to which each method integrates spectral-spatial information, which greatly affects classification performance. Another characteristic that can be used to distinguish HSI methods, which we would like to mention here, is the amount of interaction between the spectral-spatial features that each technique employs, as \cite{He_review} points out.

In the terminology of \cite{He_review}, the simplest {\it dependency system} is the case where features are extracted directly from the HSI data. This encompasses the majority of the methods we presented.
However, NNs with multiple layers naturally model a hierarchical interaction of features, with as many degrees of interaction as number of layers in the network: \cite{acquarelli_spectral-spatial_2018,chen_deep_2016,deng_hyperspectral_2018,liang_hyperspectral_2016,guo_guided_2018,hu_deep_2015,yang_hyperspectral_2018,zhong_learning_2017,gao_convolution_2018,lee_going_2017}. The same can be said for the layers of scattering networks: \cite{franchi_deep_2016,tang2015hyperspectral,our_spie,ashitha_classification_2014,haridas_comparative_2015}. This distinguishes these two classes of approaches from wavelet techniques \cite{cao_integration_2017,shen_three-dimensional_2011,qian_hyperspectral_2013,He} or from time-frequency methods \cite{He_review,Bau,jia_gabor_2015,he_discriminative_2017}, and yields more sophisticated features that provide improved classification results, as we demonstrate in \cref{sec:experiments}.

\subsection{Fourier Scattering Transform}

\label{sec:scattering}

\begin{figure}[!t]
\centering
\includegraphics[width=0.5\textwidth]{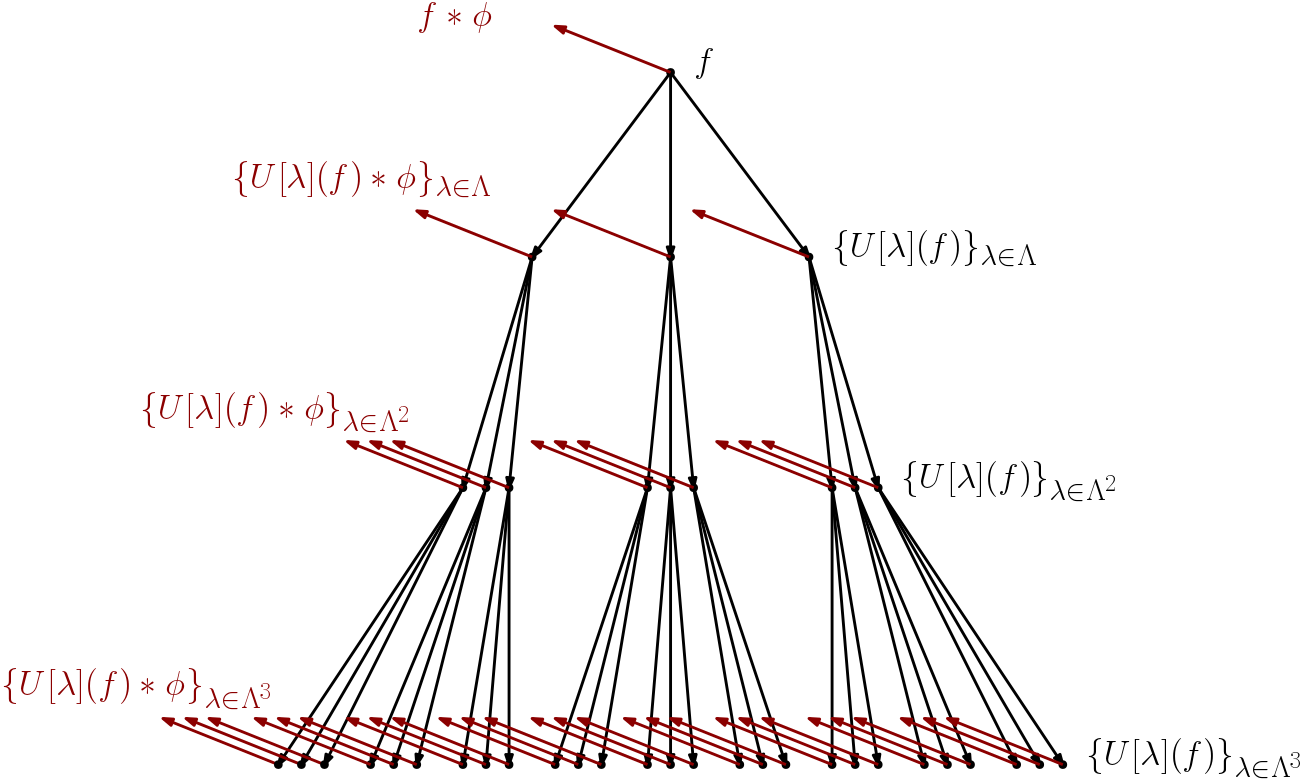}
\caption{The network structure of the scattering transform for square integrable signals in a Hilbert space. The functions $U[\lambda](f)$ are generated iteratively and they are represented by the black dots. The scattering coefficients of $f$, represented by the red dots, are found by convolving each $U[\lambda](f)$ with $\phi$.}
\label{fig:scattering-net}
\end{figure}

We begin this section by formally defining the mathematical concept of the scattering transformation. Fix a sequence $\Phi=\{\phi,\phi_\lambda\}_{\lambda\in\Lambda}$ of square integrable functions on $\R^d$, where $\Lambda$ is the index set of the sequence. Given an input function $f$ defined on $\R^d$, we iteratively convolve it with this sequence and take the modulus in the following way. For each index $\lambda\in\Lambda$, let
\[
U[\lambda](f)=|f*\phi_\lambda|,
\]
where $*$ is the convolution of functions on $\R^d$. We can extend this rule to multi-indices. For each $\lambda=(\lambda_1,\dots,\lambda_k)\in\Lambda^k$, let
\[
U[\lambda](f)
=U[\lambda_k]\cdots U[\lambda_1](f).
\]
The {\it scattering transform} $\S_\Phi$ associated with $\Phi$ is formally defined as the sequence of functions
\[
\S_\Phi(f)
=\{f*\phi,\ U[\lambda](f)*\phi\}_{\lambda\in\Lambda^k,k\geq 1}.
\]
See Figure \ref{fig:scattering-net} for a visualization of the scattering transform as a convolutional network.

The mathematical properties of the scattering transform and the features that it generates greatly depend on the underlying sequence of functions $\Phi$. Mallat \cite{mallat2012group} and his collaborators \cite{bruna2013invariant} primarily considered the wavelet (time-scale) case, where $\phi$ is the father wavelet and $\{\phi_\lambda\}_{\lambda\in \Lambda}$ are dilations of the mother wavelet function. The resulting transform is called the {\it wavelet scattering transform} (WST) and it provides a powerful multi-scale representation \cite{bruna2013invariant}. In contrast, two authors of this paper studied the time-frequency analogue \cite{czaja2017analysis,czaja2017rotationally}, where $\phi$ is a band-limited function and $\{\phi_\lambda\}$ are frequency modulations of $\phi$. The resulting transformation is called the {\it Fourier scattering transform} (FST) and it provides a novel hierarchical time-frequency representation of the data.

Although wavelet-based techniques have recently dominated the spectrum of signal processing applications in the field of HSI analysis, due to their overall impact on image processing, see e.g., \cite{jia_gabor_2015,cao_integration_2017,shen_three-dimensional_2011,hsu_evaluating_2016}, the time-frequency methods form a natural foundation for spectral data exploration. They were the basis for the early Fourier transform imaging spectroscopy methods \cite{Bennett,Lewis}, as well as for recent attempts to analyze hyperspectral imagery \cite{he_discriminative_2017}.


We note here that wavelet and Fourier scattering transforms provide entirely different representations: the WST computes localization and scale characteristics, whereas the FST provides frequency distribution information. Both transformations satisfy several similar representation-theoretic properties: they are energy preserving, non-expansive, and contract sufficiently small translations and diffeomorphisms, see the results in \cite{czaja2017analysis,mallat2012group} for precise statements. However, it is the FST that has a provable exponentially fast convergence of finite approximations, which is crucial in implementations. These properties explain why FST is an effective feature extractor. 
Indeed, as a mathematical construct, the scattering transform has an infinite number of layers and each node has infinitely many children. Thus, for practical applications we can only compute a finite subset of its coefficients. Thanks to the aforementioned property, the FST can be truncated to a finite approximation without loosing its properties. Indeed, theoretical results guarantee that the total energy contained in the $k$-th order FST coefficients is at most $\epsilon^{k-1}$ of the original energy of $f$ for some small $\epsilon\in(0,1)$, see \cite{czaja2017analysis}. 

We close this section by observing that, in a recent work, He \etal\cite{he_discriminative_2017} introduced the concept of \emph{discriminative low-rank Gabor filtering} (DLRGF) for spectral-spatial classification. The DLRGF is built upon a foundation formed by a 3D harmonic modulated with a 3D Gaussian - a concept which in mathematics is known sometimes as a \emph{Gabor frame}. The mathematical representation system $U[\lambda](f)*\phi$ generated by the FST, using the iterative convolution process described above, is in fact a generalization of a Gabor frame and is known as a \emph{uniform covering frame}. As such, our method provides the same theoretical guarantees for the analysis of HSI data as that of \cite{he_discriminative_2017}. But the main difference between these two methods is the manner in which they are computed. DLRGF builds the representation using a priori filters, while our method constructs the efficient representation through an appropriately designed iterative procedure. The advantage of the latter is that it can be made significantly faster, even by orders of magnitude. The implementation process which enables this is described in the next section.

\section{Proposed Discrete Fourier Scattering Network for HSI Classification}
\label{sec:methodology}

\begin{figure*}[!t]
\centering

	\resizebox{\textwidth}{!}{\input{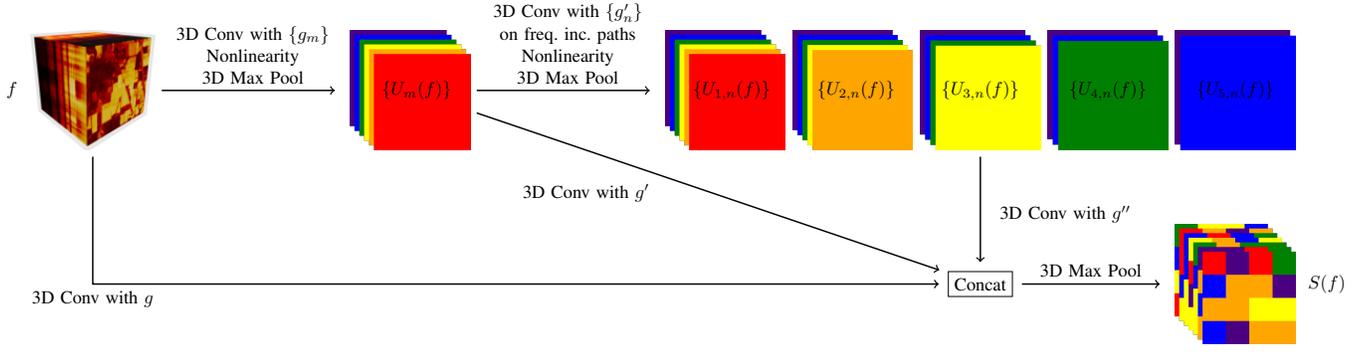}}
	\caption{Our proposed 3D FST network for HSI data. The padded input $f$ is the HSI cube is convolved with the collection of filters $\{g_m \}$ and the modulus nonlinearity is applied followed by a downsampling to create the first order intermediate 3D FST coefficients. These are in turn convolved along frequency increasing paths with $\{ g_n'\}$, followed by a modulus nonlinearity and downsampling to create the second order intermediate 3D FST coefficients. Then the input and the intermediate coefficients are locally averaged with $g,\ g'$ and $g''$, concatenated, and downsampled a final time to create our 3D FST representation $S(f)$. Because of the downsampling throughout the feature size is never significantly increased compared to the size of the input. This is in contrast to the network in \Cref{fig:scattering-net} where the number of scattering coefficient functions $\S_\Phi(f)$ grows exponentially with depth.}
\label{fig:scat_net_3d}
\end{figure*}

The FST is a generic transformation and a mathematical model that is suitable for many applications and purposes. To differentiate between the generic FST and the particular algorithm which we introduce in this paper for HSI classification, we shall call the latter the {\it three-dimensional Fourier scattering transform} (3D FST) algorithm. We now give its detailed description. 

In the context of HSI data, we consider its data dimension to be equal to $d=3$ and we represent an HSI datacube as a function $f$ defined on a 3-orthotope (i.e., rectangular box) subset of $\Z^3$. That is, $f(x,y,b)$ is the value of the image at spatial location $(x,y)$ and spectral band $b$. Fix a function $g$ on $\Z^d$, which is typically called the {\it window function}. Following standard convention, we select the window $g$ to be compactly supported in a $3$-dimensional rectangle with side lengths $M=(M_1,M_2,M_3)$. Let $\Lambda_M$ be the collection of $m\in\Z^3$ such that $0\leq m_j\leq M_j-1$ for each $j$. We define the functions $\{g_m\}_{m\in\Lambda_M}$ by the formula, 
\begin{equation}
\label{eq:g}
g_m(x,y,b)=\exp\Big(2\pi i\Big(\frac{xm_1}{M_1}+\frac{ym_2}{M_2}+\frac{bm_3}{M_3}\Big)\Big) g(x,y,b). 
\end{equation}
Here, $x$ and $y$ are the spatial coordinates and $b$ is the spectral coordinate.


There is the usual trade-off with time-frequency representations: Larger values of $M$ provide worse spatial localization but better frequency resolution, whereas a smaller $M$ yields the opposite effect. For this reason, it is reasonable to use different functions in each layer of the network to maximize the performance of the transform. Let $M'=(M'_1,M'_2,M'_3)$ and $M''=(M''_1,M''_2,M''_3)$ be multi-integers and let $g'$ and $g''$ denote functions supported in rectangles of size $M'$ and $M''$ respectively. We define $\{g'_m\}_{m\in\Lambda_{M'}}$ and $\{g''_m\}_{m\in\Lambda_{M''}}$ analogous to the definition of $g_m$ given in equation \eqref{eq:g}, except with $M'$ and $M''$ replacing $M$ and $g'$ and $g''$ replacing $g$, respectively. 

Time-frequency representations are inherently redundant. We can downsample the features in such a way that we do not lose important information, e.g., see \cite{grochenig2013foundations}, and this type of result is closely related to the classical Shannon sampling theorem. In our case, we do not downsample in the spatial dimensions. We fix positive integers $P$, $P'$, $P''$ which shall be the downsampling factors in the each of the three layers. 

The zero order 3D FST coefficient, $S_0(f)$, is defined as
\[
S_0(f)(x,y,b)
=(f*g)(x,y,Pb).
\]
Here, $*$ denotes the convolution operator on $\Z^3$. This is simply a local averaging of the input by the window function $g$ and downsampled by $P$. The first order intermediate 3D FST coefficients are
\begin{align*}
U_m(f)(x,y,b)
&=|(f*g_m)(x,y,Pb)|.
\end{align*}
The collection $\{U_m(f)\}_{m\in\Lambda_M}$ can be interpreted as the modulus of the {\it windowed Fourier transform} of $f$ (also called the {\it short-time Fourier transform} in signal processing or the {\it spectrogram} in audio processing).  

The windowed Fourier transform is not stable to small perturbations of the input function. The basic reason is that if $f$ consists of a single high frequency component, then there exists a $\tilde f$ such that $\tilde f$ is a small diffeomorphism of $f$ and its frequency support is disjoint from that of $f$; consequently, $f$ and $\tilde f$ are very different in both the $L^2$ metric, see \cite{mallat2012group,anden2014deep} for a rigorous analysis. To avoid this behavior in 3D FST we proposes to locally average $U_m(f)$ with $g'$. The first order 3D FST coefficients are then:
\[
S_m(f)(x,y,b)
=(U_m(f)*g')(x,y,P'b).
\]
Hence, the first order 3D FST coefficients carry information about a spatially-averaged short-time Fourier transform of $f$. 

While naive local averaging improves stability to small deformations, it also removes a significant amount of high-frequency information because $g$ is a low-pass filter. The lost components are aggregated in the functions $U_m(f)*g_n$. However, these functions suffer from the same instability properties as $U_m(f)$. The second order intermediate 3D FST coefficients are thus,
\begin{align*}
U_{m,n}(f)(x,y,b)
&=|(U_m(f)*g_n')(x,y,P'b)|.
\end{align*}
These intermediate coefficients are also unstable to small diffeomorphisms, so we perform a local averaging. The second order 3D FST coefficients are now defined to be:
\[
S_{m,n}(f)(x,y,b)
=(U_{m,n}(f)*g'')(x,y,P''b).
\]
We also note that theoretical results in \cite{czaja2017analysis} guarantee that $S_{m,n}(f)$ is small when $n\geq m$, so we can improve the computational efficiency of the algorithm by only computing the coefficients for which $n\neq m$.

The zero and first order 3D FST coefficients can be interpreted as spatially-smoothed versions of classical spectral-spatial representations. It is not as obvious what the second order coefficients represent. At first glance, the second order 3D FST coefficients appear similar to the Mel-frequency cepstral coefficients (MFCCs), but there is an important distinction. The MFCCs are calculated by fixing the spatial coordinate and then further decomposing the spectrogram along the frequency axis in log scale. MFCCs play an important role in audio analysis because more global characteristics, which are not captured by the spectrogram, contain important information. 

In contrast to the MFCCs, the second order 3D FST coefficients are calculated by fixing the spectral variable and then further decomposing along the spatial coordinate. That is, $U_{m,n}(f)$ describes whether the $m$-th  frequency of $f$ over intervals of length $M$ (a local property captured by the first order coefficients) varies at frequency $n$ over intervals of length $MM'$ (a more global property). 

In summary, given a hyperspectral image $f$, the features generated by the 3D FST at location $(x,y)$ are the collection of vectors 
\begin{align*}
\text{Zero order:} &\quad S_0(f)(x,y,\cdot) \\
\text{First order:} &\quad \{S_m(f)(x,y,\cdot)\}_{m\in\Lambda_M} \\
\text{Second order:} &\quad \{S_{m,n}(f)(x,y,\cdot)\}_{m\in\Lambda_M,n\in\Lambda_{M'}}. 
\end{align*}
These vectors are concatenated to form a feature vector for each pixel $(x,y)$ of the hyperspectral image $f$.

In \Cref{fig:scat_net_3d} is a schematic of our proposed 3D FST method as we implemented it with standard tensorflow \cite{tf_short} primitives on the GPU: 3D convolutions with nonlinearities and 3D max pooling operations.
This spatial domain implementation, as opposed to a frequency domain implementation, has the advantage of benefiting from the highly optimized tensorflow programming interface, which can distribute the necessary computations over ubiquitous and very powerful modern GPUs. The scattering layer that we programmed can also thus be seamlessly blended into any stage of a deep network since it allows backpropagation, we leave such integrations to future work.
\Cref{fig:scat_net_3d} also illustrates how our method can be performed on an input of any size since it maps any HSI cube to an HSI-feature cube with equal spatial dimension.
We release our implementation publicly \cite{pyfst} and discuss its runtime performance in \Cref{sec:perf}.

\begin{table*}[t]
  \centering
  \caption{Attributes of the datasets used.}
  \label{tab:data}
  \begin{tabular}{c|c|c|c|c|c|c|c}
Name     & Satellite  & No. Bands & Bandwidth   & Meters per Pixel & Dimensions H$\times$W & No. Labeled Pixels & No. Classes \\ \hline
PaviaU   & ROSIS     & 103       & 430-860 nm  & 1.3 m            & 610x340        & 42776              & 9           \\
IP       & AVIRIS    & 224       & 400-2500 nm & 3.7 m            & 145x145        & 10249              & 16          \\
KSC   & AVIRIS     & 224       & 400-2500 nm & 18 m            & 512x614        & 5211               & 13          \\
Botswana     & NASA EO-1 & 242       & 400-2500 nm & 30 m             & 1476x256       & 3248               & 14         

  \end{tabular}
\end{table*}

\begin{figure*}[!t]
\centering
\begin{subfigure}[b]{.25\textwidth}
	\centering
	\includegraphics[width=.95\linewidth]{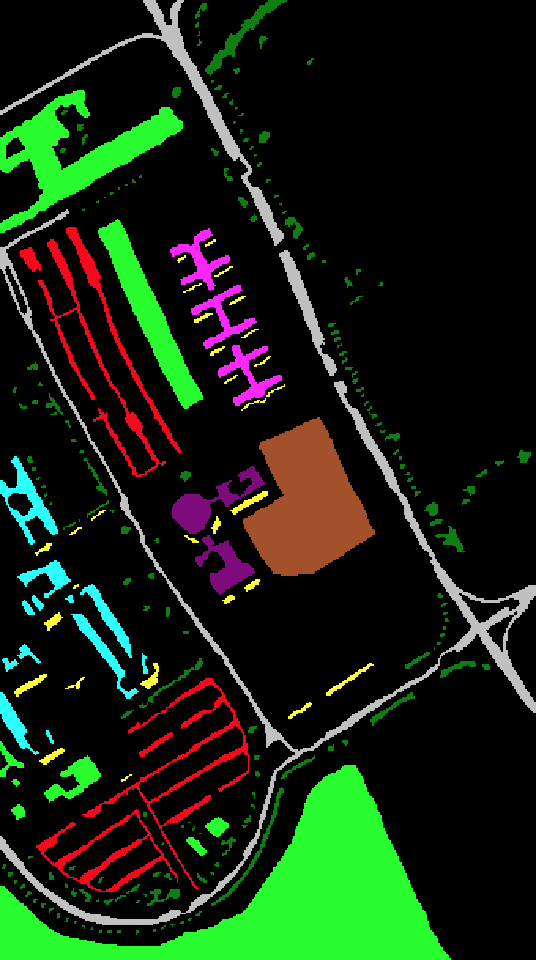}
	\caption{}
	\label{fig:PUground_truth}
\end{subfigure}%
\begin{subfigure}[b]{.25\textwidth}
	\centering
	\includegraphics[width=.95\linewidth]{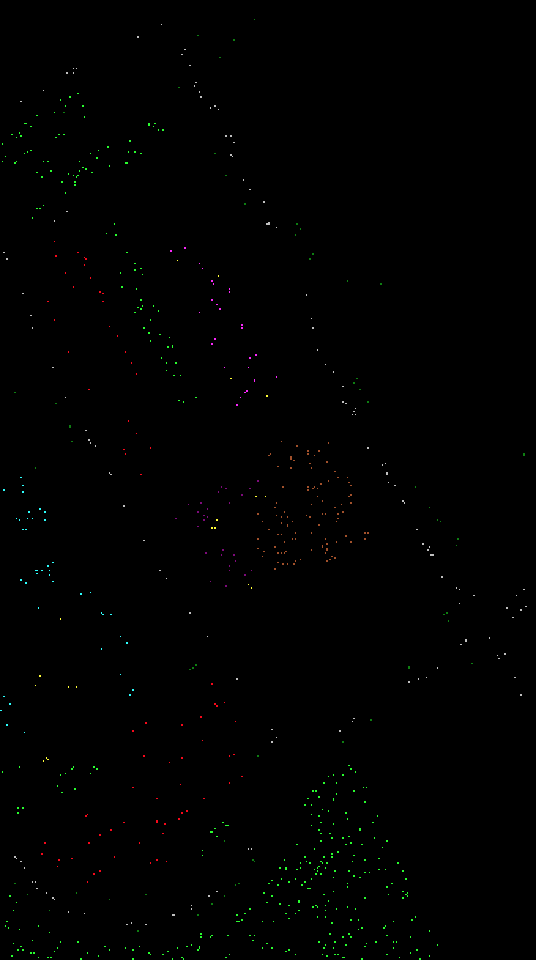}
	\caption{}
	\label{fig:PUdistributed}
\end{subfigure}%
\begin{subfigure}[b]{.25\textwidth}
	\centering
	\includegraphics[width=.95\linewidth]{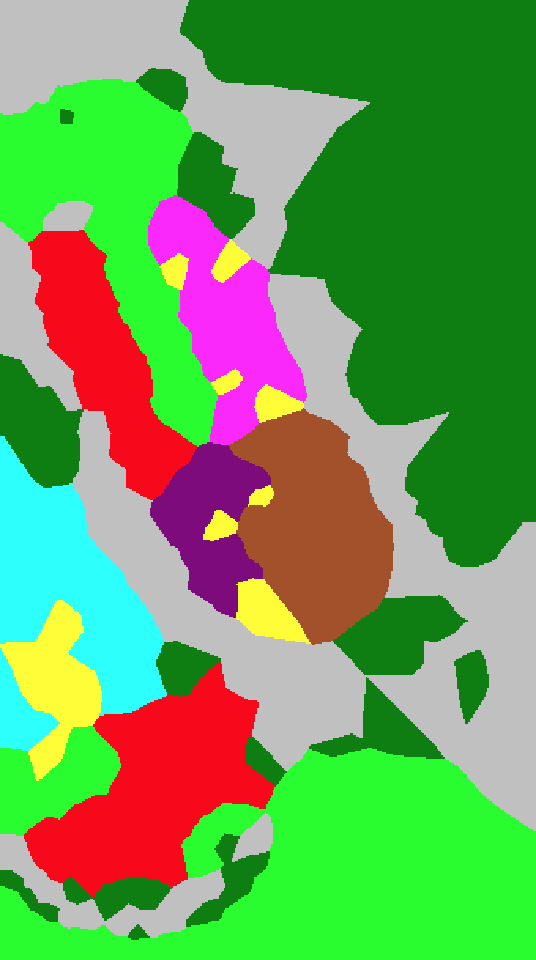}
	\caption{}
	\label{fig:PUknn}
\end{subfigure}%
\begin{subfigure}[b]{.25\textwidth}
	\centering
	\includegraphics[width=.95\linewidth]{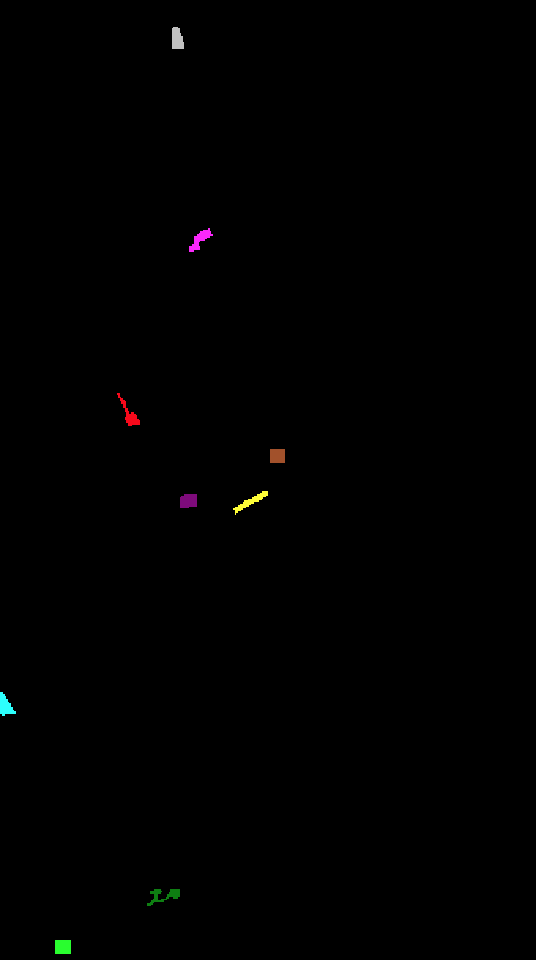}
	\caption{}
	\label{fig:PUsss}
\end{subfigure}
\hfill
\caption{An example using Pavia University to illustrate strictly site specific sampling for creating training masks. \protect\subref{fig:PUground_truth} Ground Truth Labels. \protect\subref{fig:PUdistributed} 2\% of labels randomly selected. \protect\subref{fig:PUknn} 1-KNN interpolation of the labels in \protect\subref{fig:PUdistributed} which is correct for 93\% of all labels. \protect\subref{fig:PUsss} 2\% of the labels selected in a strictly site specific manner. Note how there is only 1 site per class label with all its pixels in a connected set. The 1-KNN interpolation of the labels in \protect\subref{fig:PUsss} would be correct for 22\% of all the labels.}
\label{fig:PaviaU_full_class}
\end{figure*}

\section{Experimental Results and Discussion}
\label{sec:experiments}

\subsection{Data Sets}

We test the performance of these feature extractors on the following hyper-spectral databases:

\begin{itemize}
    \item
    {\it Indian Pines (IP)} acquired over the Indian Pines test site in Northwestern Indiana in 1992 by the Airborne Visible / Infrared Imaging Spectrometer (AVIRIS) sensor \cite{indianpinesdata}. 49\% of all pixels are labelled.
    \item
    {\it Pavia University (PaviaU)} acquired during a 2001 flight campaign over Pavia, northern Italy, using the reflective optics system imaging spectrometer (ROSIS) sensor \cite{paviaUdata}. 21\% of all pixels are labelled.
    \item
    {\it Kennedy Space Center (KSC)} acquired over the wetlands on the west shore of the Kennedy Space Center (KSC) and the Indian River using the AVIRIS sensor \cite{kscdata}. 1.7\% of all pixels are labelled.
    \item
    {\it Botswana} acquired over the Okavango Delta, Botswana in 2001, by the Hyperion sensor on the NASA EO-1 satellite \cite{botswanadata}. 0.86\% of all pixels are labelled.
\end{itemize}

Table \ref{tab:data} shows additional information on all datasets.
All datasets can be downloaded from the webpage \cite{hyperspectral2018}, and we also make them available in our released code \cite{pyfst}.

\begin{figure*}[!t]
\centering
\begin{subfigure}[b]{.25\textwidth}
	\centering
	\includegraphics[width=.95\linewidth]{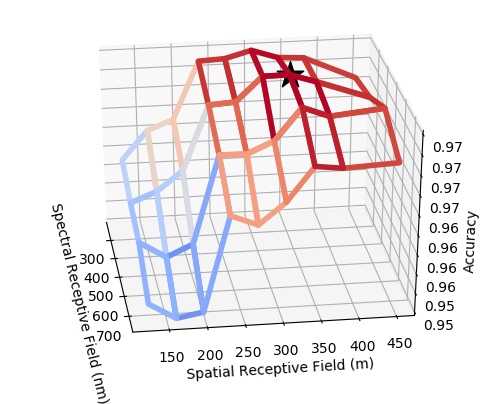}
	\caption{PaviaU}
\end{subfigure}%
\begin{subfigure}[b]{.25\textwidth}
	\centering
	\includegraphics[width=.95\linewidth]{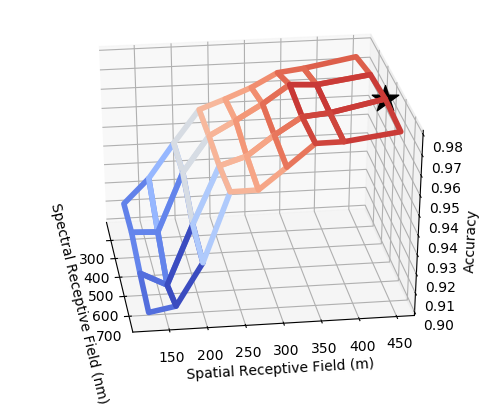}
	\caption{Indian Pines}
\end{subfigure}%
\begin{subfigure}[b]{.25\textwidth}
	\centering
	\includegraphics[width=.95\linewidth]{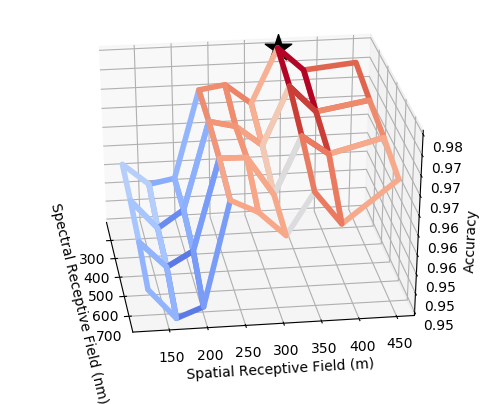}
	\caption{KSC}
\end{subfigure}%
\begin{subfigure}[b]{.25\textwidth}
	\centering
	\includegraphics[width=.95\linewidth]{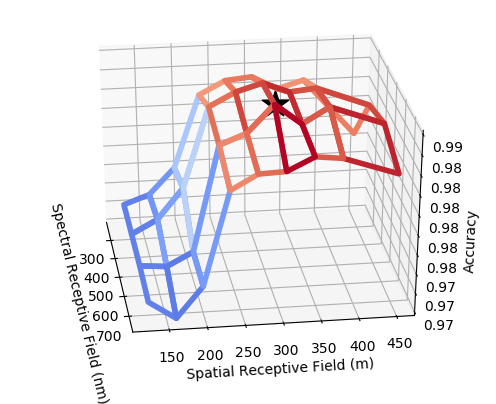}
	\caption{Botswana}
\end{subfigure}
\hfill
\begin{subfigure}[b]{.25\textwidth}
	\centering
	\includegraphics[width=.95\linewidth]{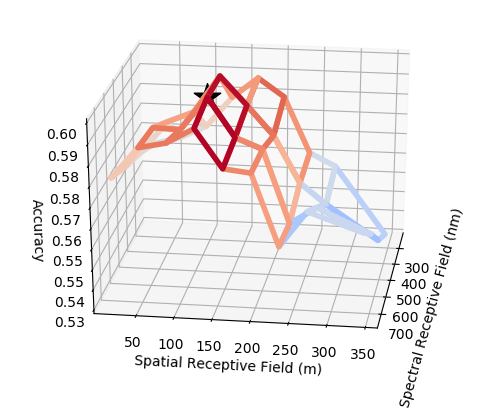}
	\caption{PaviaU (SSS)}
\end{subfigure}%
\begin{subfigure}[b]{.25\textwidth}
	\centering
	\includegraphics[width=.95\linewidth]{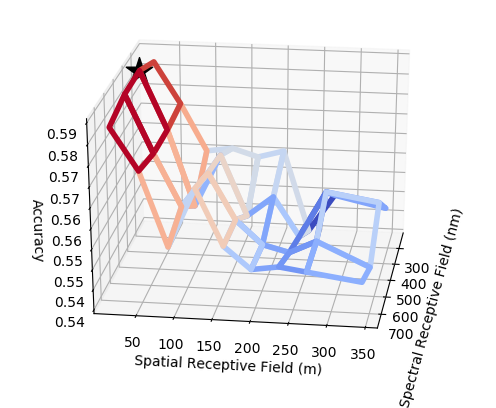}
	\caption{Indian Pines (SSS)}
\end{subfigure}%
\begin{subfigure}[b]{.25\textwidth}
	\centering
	\includegraphics[width=.95\linewidth]{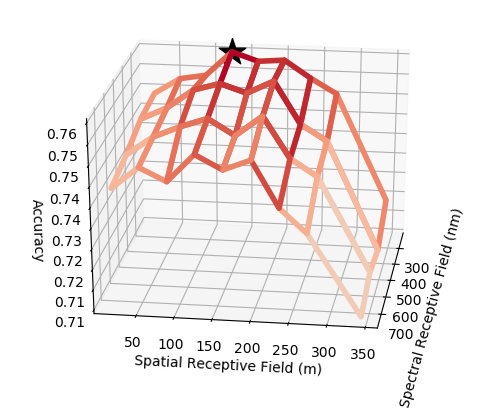}
	\caption{KSC (SSS)}
\end{subfigure}%
\begin{subfigure}[b]{.25\textwidth}
	\centering
	\includegraphics[width=.95\linewidth]{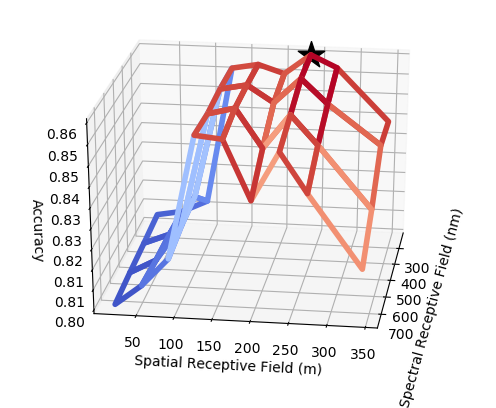}
	\caption{Botswana (SSS)}
\end{subfigure}
\hfill
\caption{Gridsearches over the hyperparameters $M,M',M''$ of the 3D FST for each dataset we analyze. The vertical axis is the average classification accuracy over 10 trials. Each dataset generates a unique shape because of the various degree of spatial or spectral noise that should be smoothed. Larger receptive fields perform more smoothing. Each of the datasets in the second row is constructed in a single site specific manner as elaborated in the main text. Each of the datasets in the first row is constructed with the more typical random distributed sampling. A black star marks the best result per gridsearch which determine the hyperparameters we use with our 3D FST method.}
\label{fig:gridsearches}
\end{figure*}


\subsection{Methods Compared}

To better evaluate the performance of our proposed method we implement three state of the art feature HSI classification methods, and score all methods in the same exact training and testing conditions. We release our tensorflow implementation of these methods with the rest of our code \cite{pyfst}.

\begin{itemize}
    \item
    {\it PCA + Deep Learning (DFFN)} We re-implement a Deep Feature Fusion Network exactly to the specification of \cite{dffn}, which is publicly available in a caffe implementation \cite{dffn_code}. This method projects the HSI data to a few PCA components and passes windows to a deep network consisting of three towers with 4 or 5 residual convolution blocks. This is a very deep network that in the case of PaviaU contains 34 2D convolutional layers. It has a large receptive field of 23 or 25. With our implementation we are able to replicate the same performance as in the original paper for the Indian Pines and PaviaU datasets \cite{dffn_code}.
    \item
    {\it Extended Attribute Profiles + Deep Learning (EAP)} We implement a Deep Learning With Attribute Profiles method to the specification of the EAP-Area method in \cite{eap}. This method projects the HSI data to 4 PCA components and then creates APs of length 4 for each component. With a receptive field of $9$, the $9\times9\times 36$ cube is passed to a neural network containing 3 2D convolutional layers and 2 Fully Connected layers before a softmax classifier. In our implementation we exceed the performance in the original paper for the PaviaU dataset \cite{eap}.
    \item
    {\it 3D Wavelet Scattering Transform (WST)} We implement a 3D Wavelet scattering approach to the specification of \cite{tang2015hyperspectral}. It is a 2 layer network with $7\times7\times7$ wavelet filters (receptive field of 19) with 9 orientations and 3 scales per layer.
    We use the same strategy of downsampling that we employ in our 3D FST in our implementation to enhance performance. We classify the features with a linear SVM and are able to match the performance in \cite{tang2015hyperspectral} for our datasets.
\end{itemize}

The DFFN and EAP methods are Deep Learning methods for which we perform from-scratch training for every single training dataset. In this paper the results we report are from hundreds of training runs for DFFN and EAP.
To make sure we train the best neural network possible for each trial, we extract a validation set from each trial's test set, and we train until validation loss is non-decreasing for 20 consecutive epochs. We evaluate every 2 epochs and save the model with the highest validation accuracy. For overall accuracy we report the accuracy on the original test set.
We use the same batch sizes as in the original papers \cite{dffn,eap}, and adjust the learning rate (in the range 0.1 to 1e-5) to ensure convergence on our training sets of a variety of sizes. We use the Adam optimizer for training all of our networks \cite{adam}. Further details are available in our released source code \cite{pyfst}. We do not use data augmentation for any method.
For additional comparison we also include additional popular methods:

\begin{itemize}
    \item
    {\it 3D Gabor Filters} Three dimensional Gabor filters are a popular feature extractor for HSI classification \cite{Bau,shen_three-dimensional_2011,jia_gabor_2016}. In fact, a truncation at the first layer of our proposed 3D Fourier Scattering feature extractor ($U_m(f)$) is equivalent to Gabor filtering. Thus we compare to such a truncated version of the FST on each dataset, using the same parameters for the Gabor filters as used in the the first layer of our scattering filters. This allows us to directly see what additional information the following layers of the FST extract to aid classification. After extracting the Gabor filters, we use a linear SVM for classification.
    \item
    {\it Raw features} We denote this in the text as the {\it Raw} method. It is simply the raw spectrum of a pixel used as a feature vector for linear SVM classification.
\end{itemize}

Every time that we use an SVM for classification, it is linear with the regularization parameter fixed to $C=1000$. We chose this parameter by cross-validating for $C=10^{-10}, 10^{-9}, ..., 10^{10}$ on Raw SVM, where we found $C=1000$ leads to the best Raw+SVM overall classification accuracy.

\subsection{Hyperparameters of 3D FST}

\begin{figure}[!t]
\centering
\begin{subfigure}[b]{.24\textwidth}
	\centering
	\includegraphics[width=.95\linewidth]{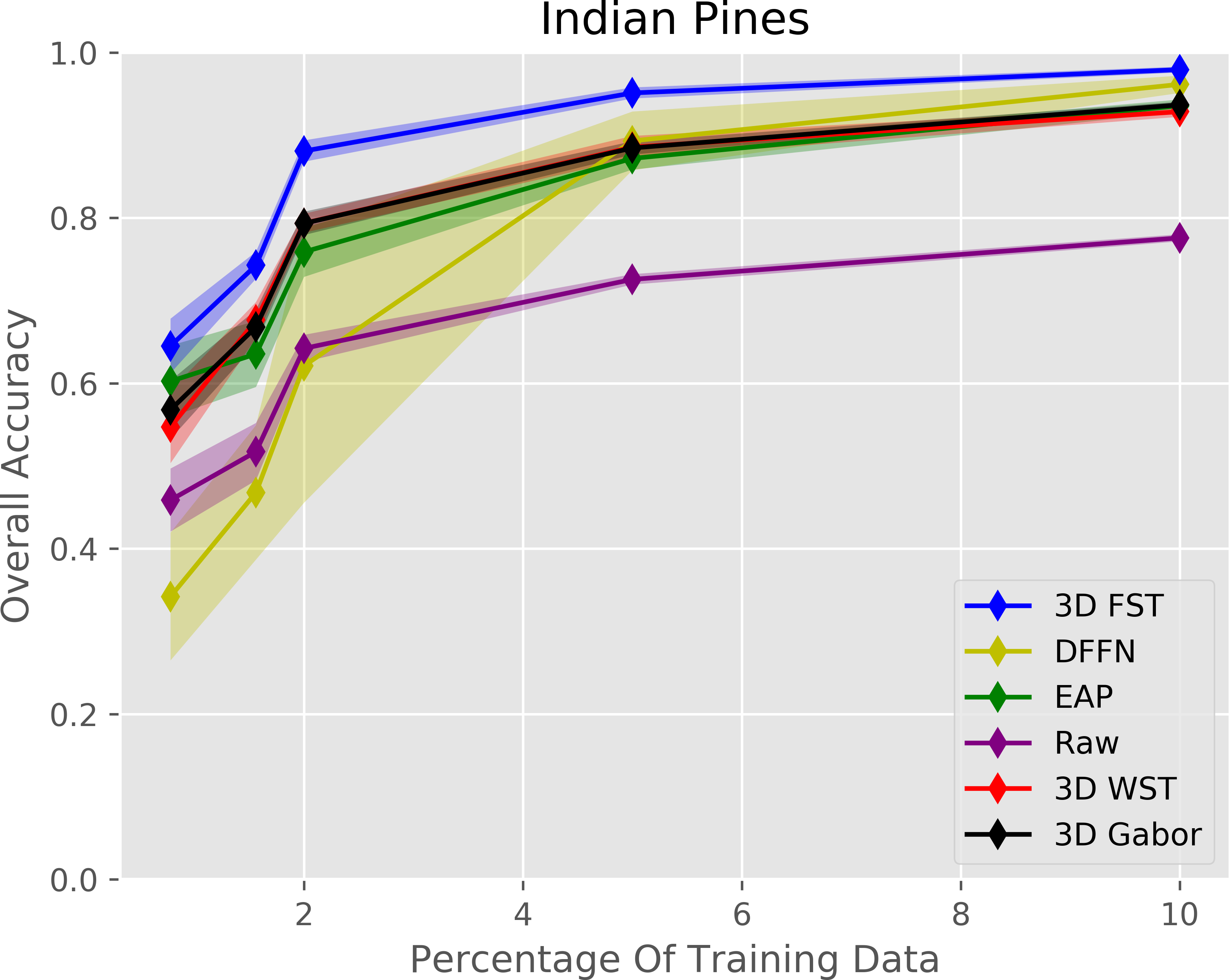}
\end{subfigure}%
\begin{subfigure}[b]{.24\textwidth}
	\centering
	\includegraphics[width=.95\linewidth]{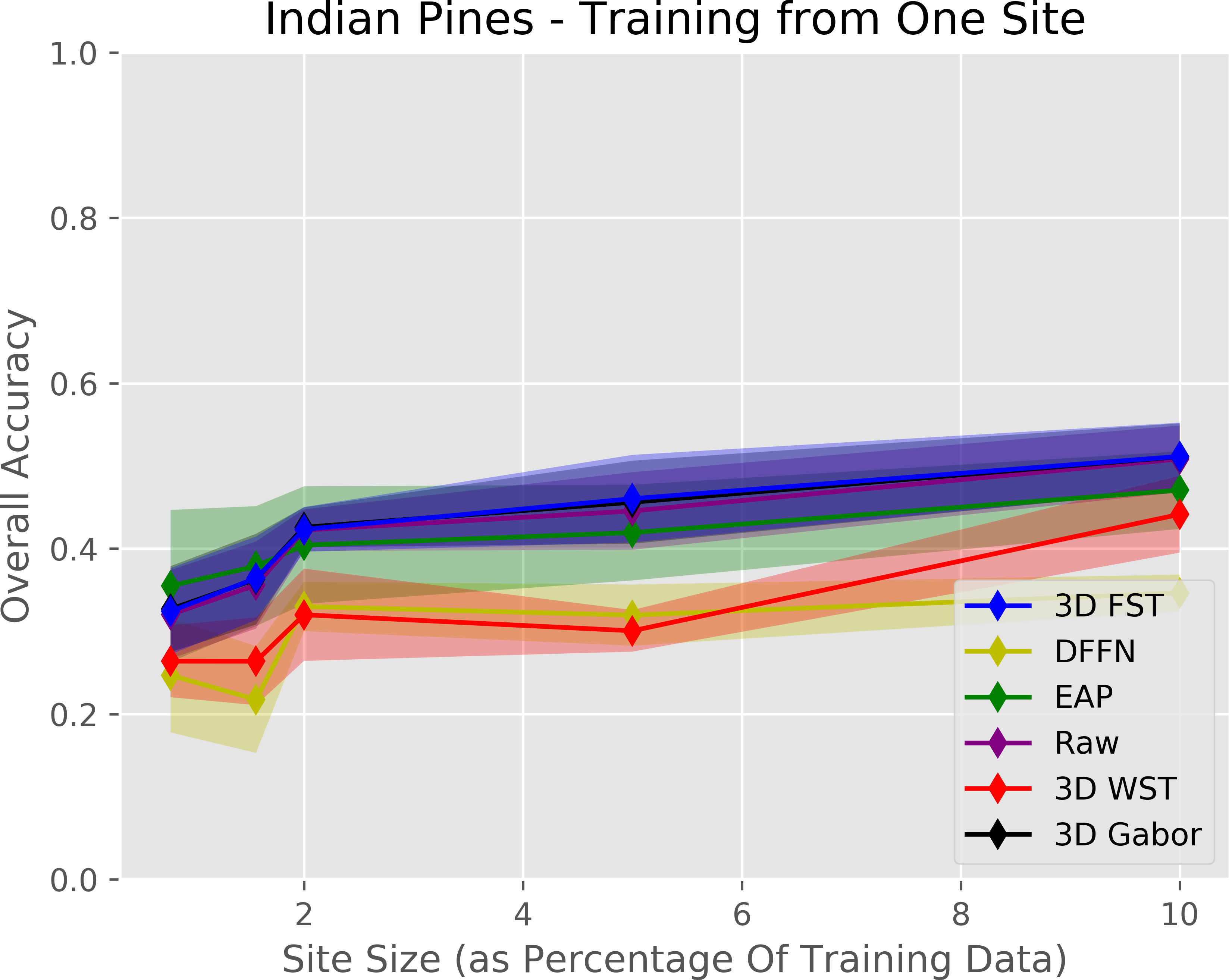}
\end{subfigure}%
\hfill
\caption{Performance of the 6 implemented methods on randomly sampled training sets from Indian Pines. Each point is the average of 10 trials, the shading is the standard deviation over the 10 trials. Each method is tested on the same 10 training sets per training set size. 3D FST performs optimally in all training scenarios on randomly distributed data. SSS training is noisy and low performance for this dataset, FST provides a slight benefit over Gabor features.}
\label{fig:IPperf}
\end{figure}

\begin{figure}[!t]
\centering
\begin{subfigure}[b]{.24\textwidth}
	\centering
	\includegraphics[width=.95\linewidth]{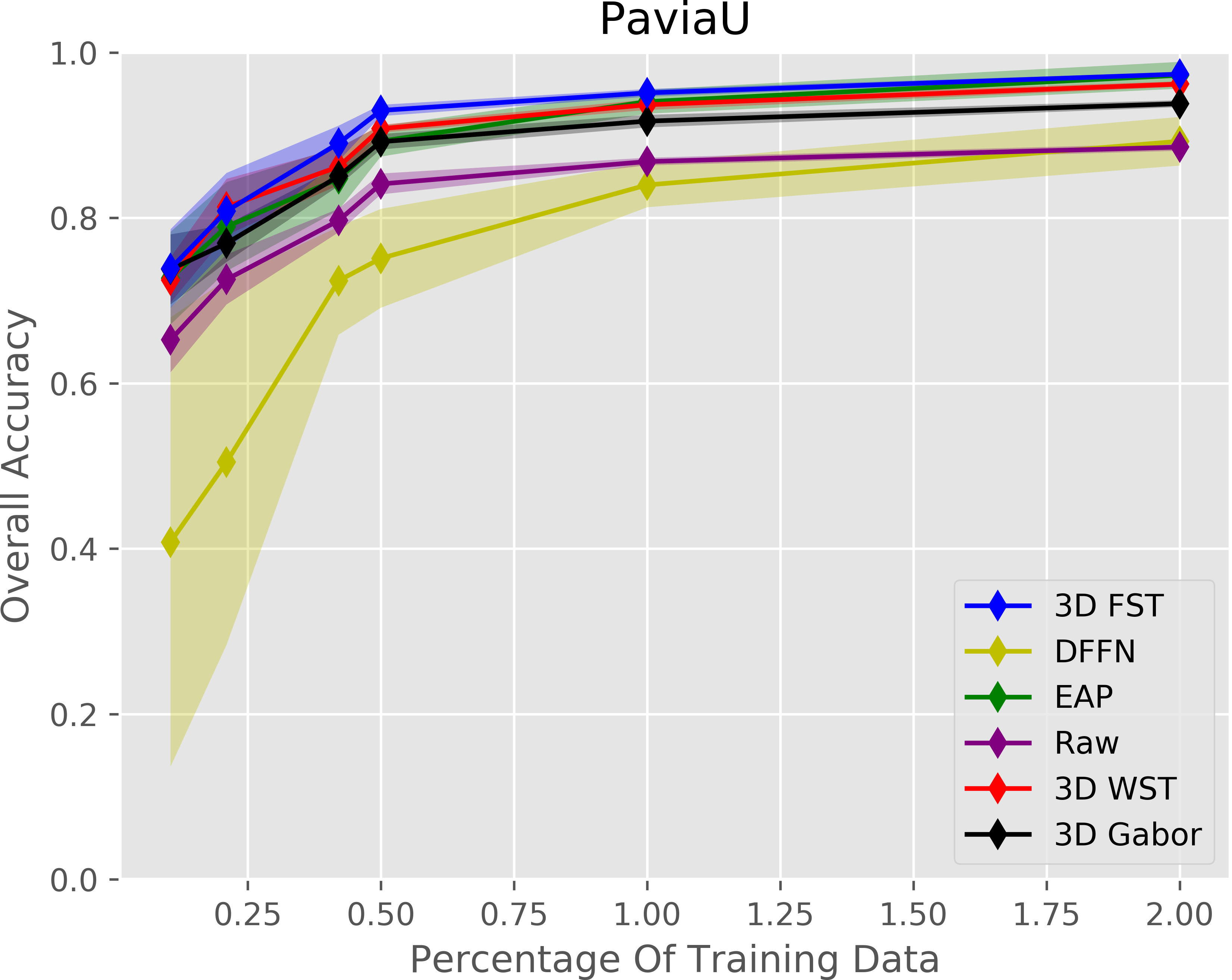}
\end{subfigure}%
\begin{subfigure}[b]{.24\textwidth}
	\centering
	\includegraphics[width=.95\linewidth]{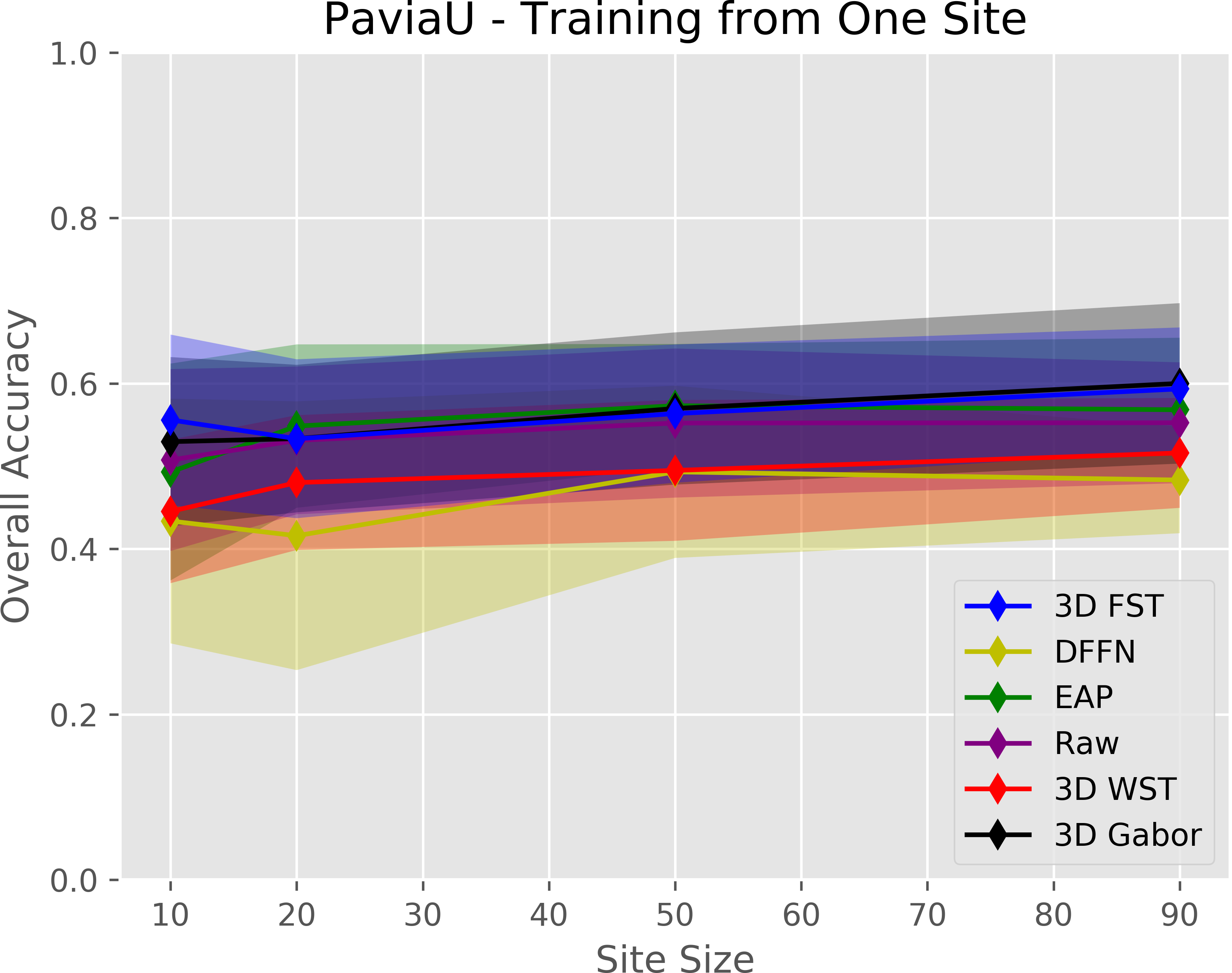}
\end{subfigure}%
\hfill
\caption{Performance of the 6 implemented methods on randomly sampled training sets from PaviaU. Only DFFN, which is the deepest method by far, does not clearly surpass raw features in this setting of very limited training data. SSS training very noisy, and the less deep networks tie, and Wavelet features underperform Fourier features.}
\label{fig:Paviaperf}
\end{figure}

\begin{figure*}[ht!]
\begin{minipage}[b][10cm][t]{.8\textwidth}
\centering
\begin{subfigure}[b]{.2\textwidth}
	\centering
	\includegraphics[width=.98\linewidth]{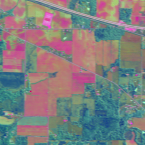}
	\caption{False Color}
\end{subfigure}~
\begin{subfigure}[b]{.2\textwidth}
	\centering
	\includegraphics[width=.98\linewidth]{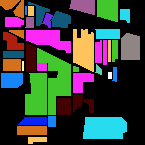}
	\caption{Ground Truth}
\end{subfigure}~
\begin{subfigure}[b]{.2\textwidth}
	\centering
	\includegraphics[width=.98\linewidth]{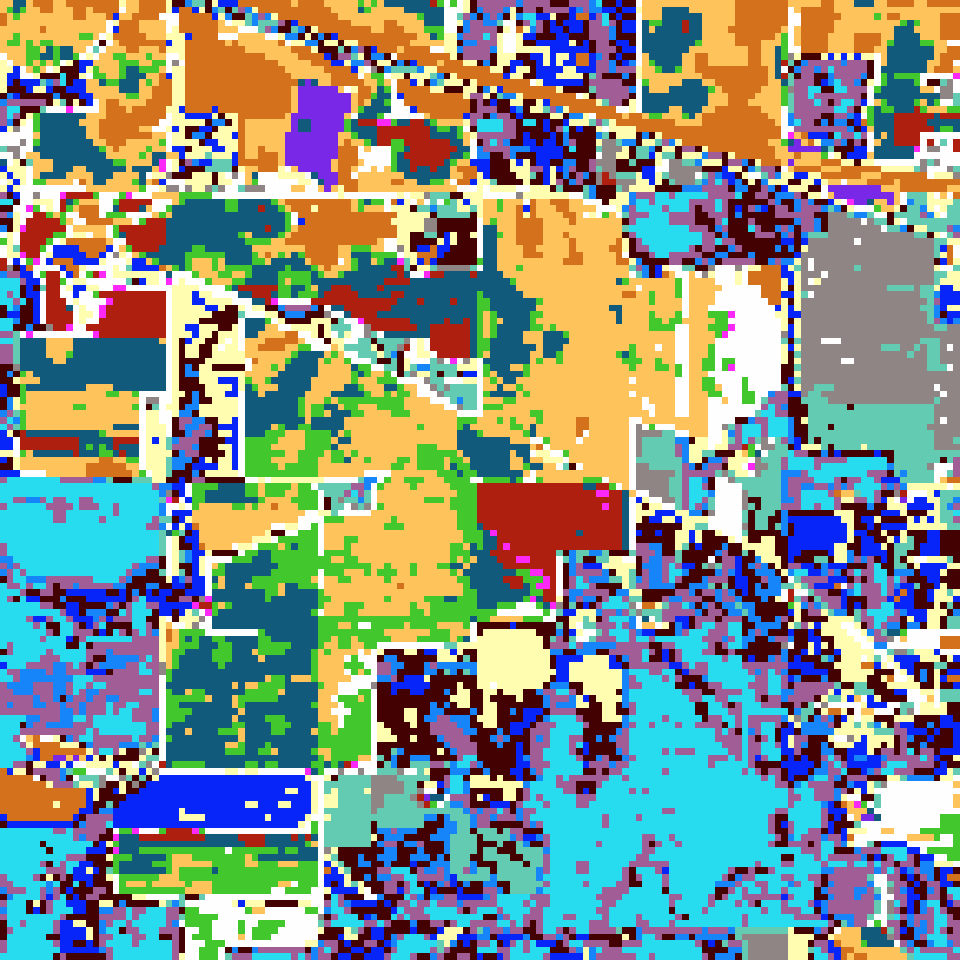}
	\caption{Raw}
\end{subfigure}~
\begin{subfigure}[b]{.2\textwidth}
	\centering
	\includegraphics[width=.98\linewidth]{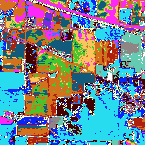}
	\caption{Raw SSS}
\end{subfigure}
\hfill
\begin{subfigure}[b]{.17\textwidth}
	\centering
	\includegraphics[width=.98\linewidth]{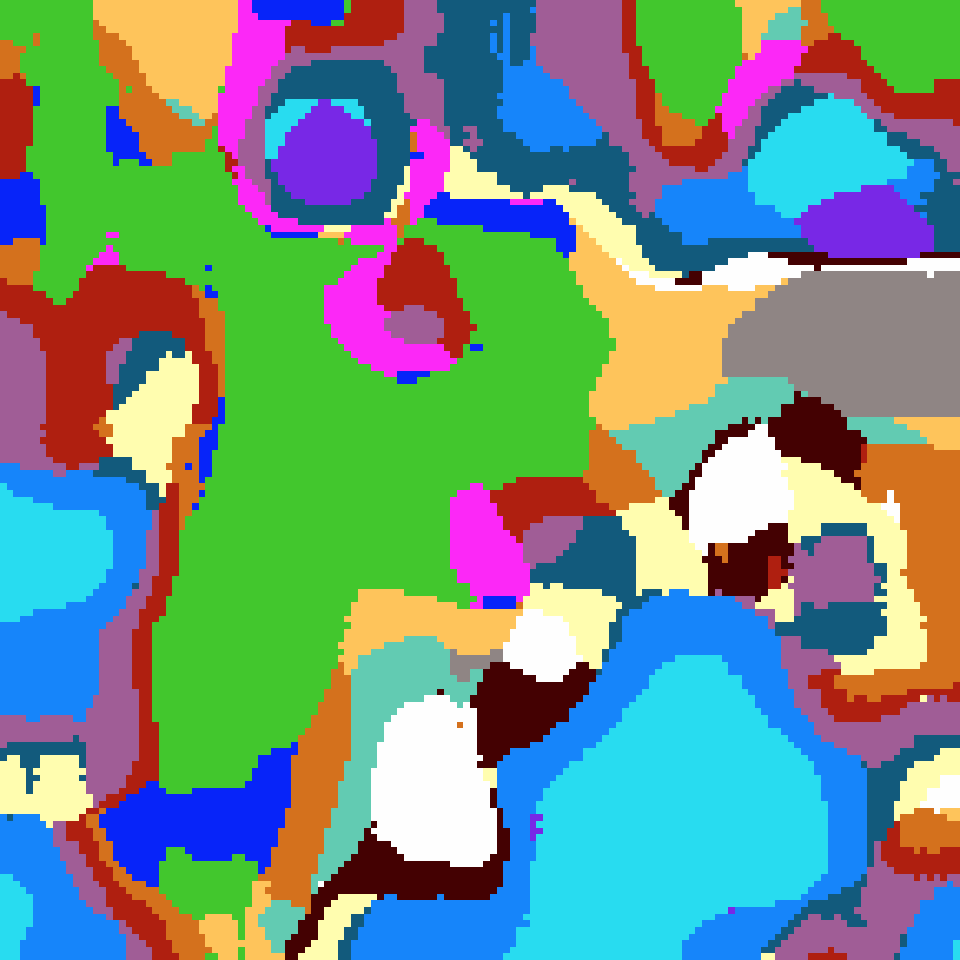}
	\caption{DFFN}
\end{subfigure}~
\begin{subfigure}[b]{.17\textwidth}
	\centering
	\includegraphics[width=.98\linewidth]{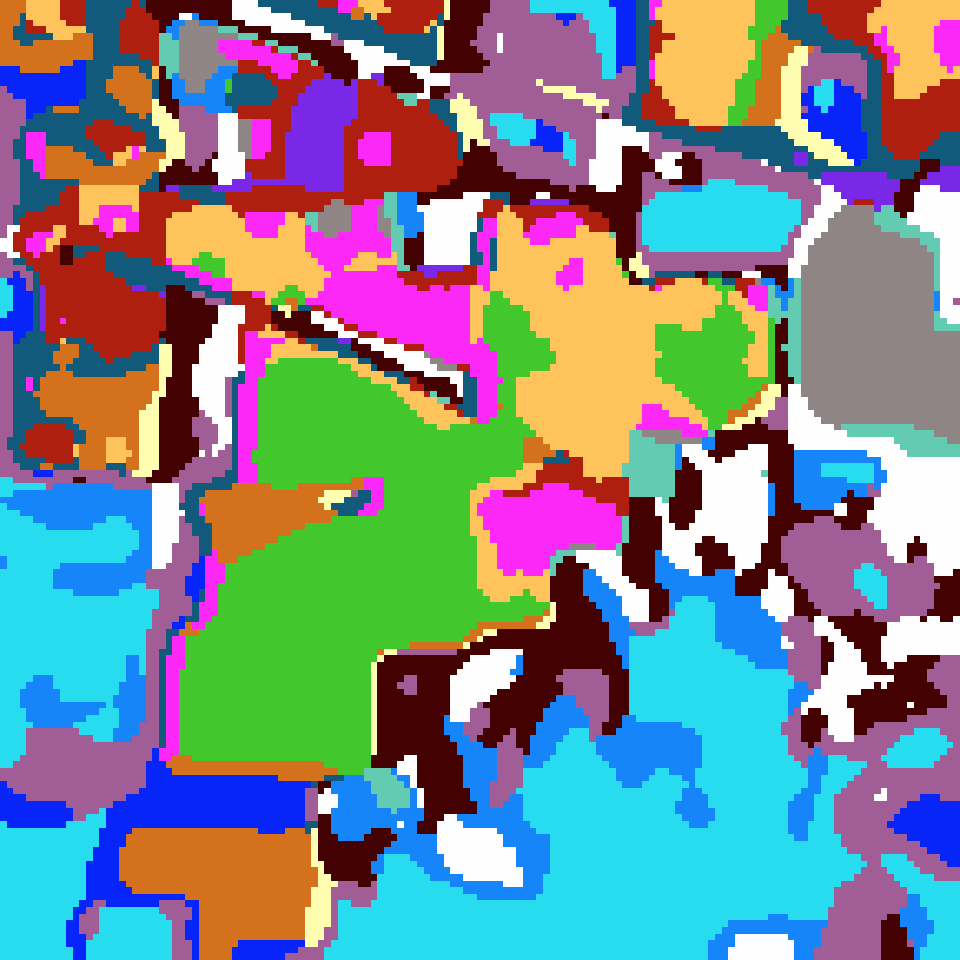}
	\caption{EAP}
\end{subfigure}~
\begin{subfigure}[b]{.17\textwidth}
	\centering
	\includegraphics[width=.98\linewidth]{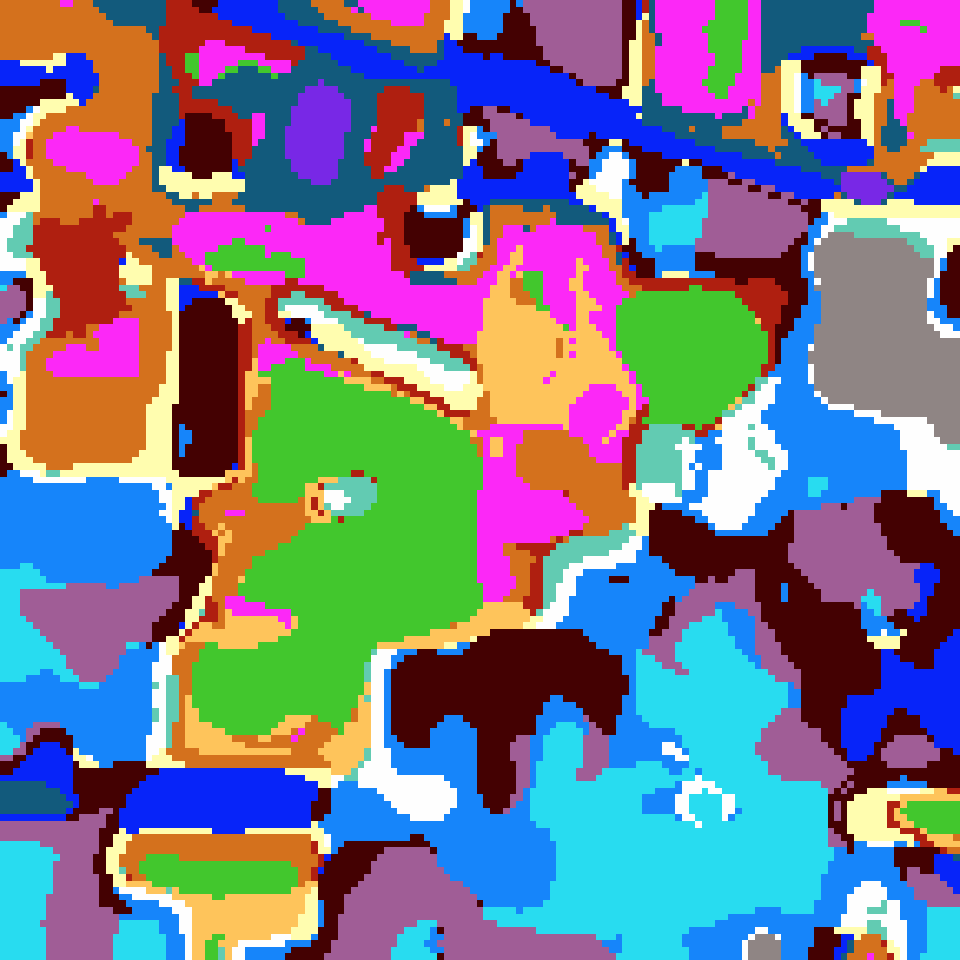}
	\caption{3D WST}
\end{subfigure}~
\begin{subfigure}[b]{.17\textwidth}
	\centering
	\includegraphics[width=.98\linewidth]{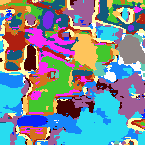}
	\caption{3D Gabor}
\end{subfigure}~
\begin{subfigure}[b]{.17\textwidth}
	\centering
	\includegraphics[width=.98\linewidth]{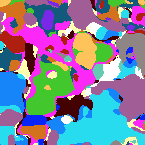}
	\caption{3D FST}
\end{subfigure}
\hfill
\begin{subfigure}[b]{.17\textwidth}
	\centering
	\includegraphics[width=.98\linewidth]{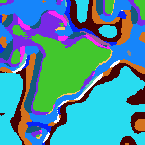}
	\caption{DFFN SSS}
\end{subfigure}~
\begin{subfigure}[b]{.17\textwidth}
	\centering
	\includegraphics[width=.98\linewidth]{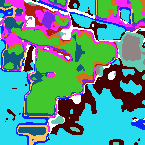}
	\caption{EAP SSS}
\end{subfigure}~
\begin{subfigure}[b]{.17\textwidth}
	\centering
	\includegraphics[width=.98\linewidth]{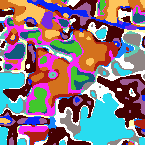}
	\caption{3D WST SSS}
\end{subfigure}~
\begin{subfigure}[b]{.17\textwidth}
	\centering
	\includegraphics[width=.98\linewidth]{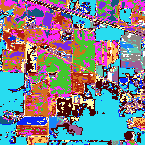}
	\caption{3D Gabor SSS}
\end{subfigure}~
\begin{subfigure}[b]{.17\textwidth}
	\centering
	\includegraphics[width=.98\linewidth]{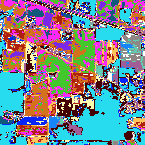}
	\caption{3D FST SSS}
\end{subfigure}
\hfill
\end{minipage}%
\begin{minipage}[b][10cm][t]{.15\textwidth}
\vspace*{\fill}
\centering
\begin{subfigure}{\linewidth}
	\centering
	\includegraphics[width=.85\linewidth]{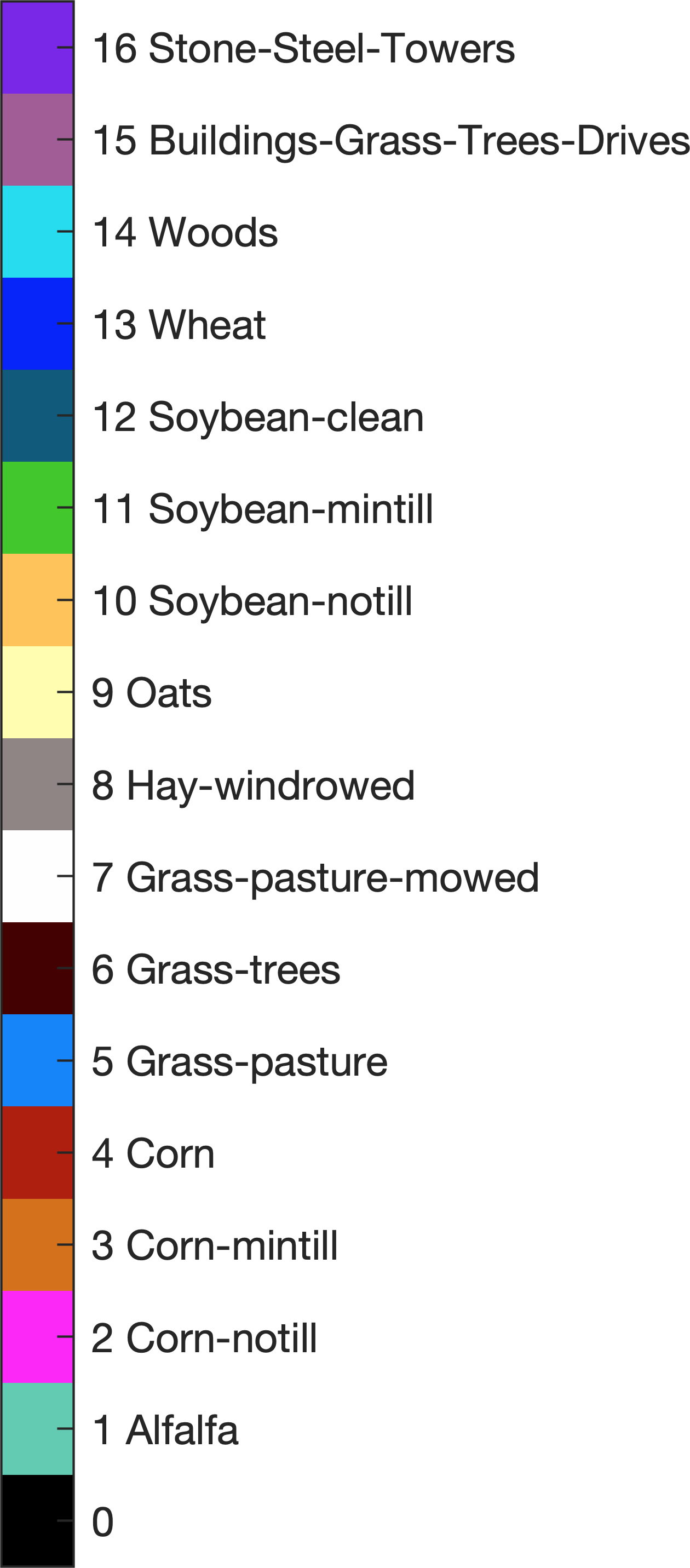}
	\caption{}
\end{subfigure}
\vspace*{\fill}
\end{minipage}
\caption{Full classification for Indian Pines with 5 samples per class (0.8\% of training data). The best performing trial by classification accuracy per model is plotted. The classification performance for these methods is in \Cref{fig:IPperf}.}
\label{fig:IPall}
\end{figure*}

\begin{figure*}[ht!]
\centering
\begin{subfigure}[b]{.18\textwidth}
	\centering
	\includegraphics[width=.95\linewidth]{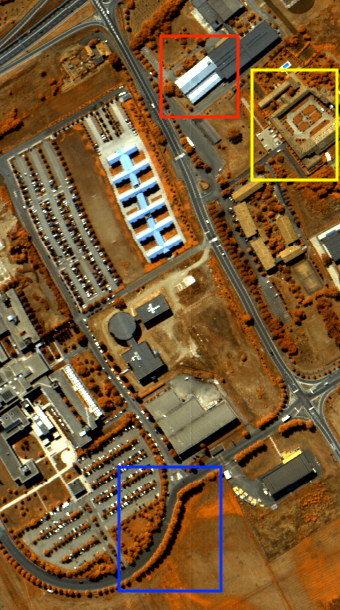}
	\caption{False Color}
	\label{fig:PUcolor}
\end{subfigure}~
\begin{subfigure}[b]{.18\textwidth}
	\centering
	\includegraphics[width=.95\linewidth]{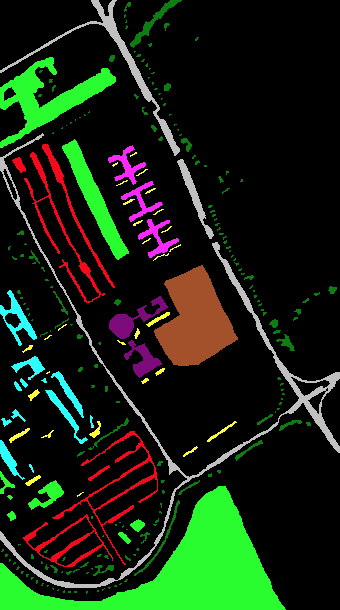}
	\caption{Ground Truth}
	\label{fig:Pugt}
\end{subfigure}~
\begin{subfigure}[b]{.18\textwidth}
	\centering
	\includegraphics[width=.95\linewidth]{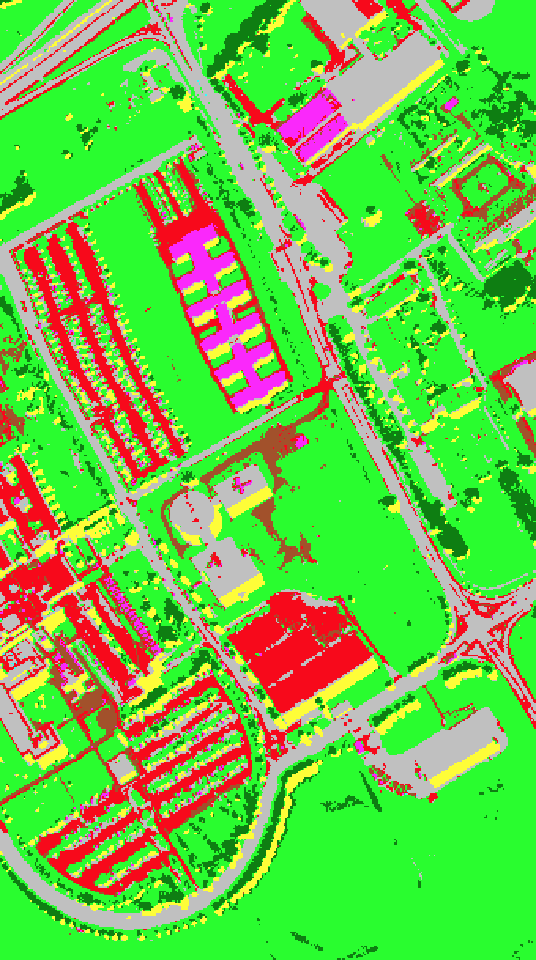}
	\caption{Raw}
	\label{fig:Puraw}
\end{subfigure}~
\begin{subfigure}[b]{.18\textwidth}
	\centering
	\includegraphics[width=.95\linewidth]{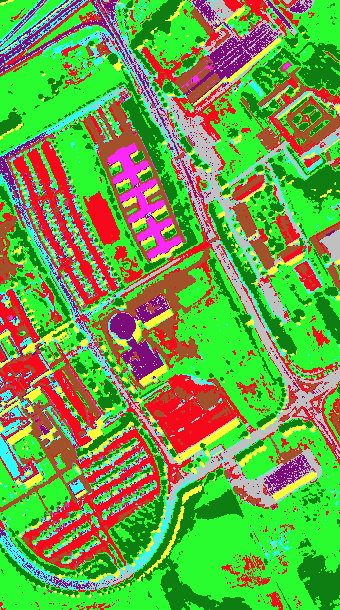}
	\caption{Raw SSS}
	\label{fig:Puraw}
\end{subfigure}~
\begin{subfigure}[b]{.18\textwidth}
	\centering
	\includegraphics[width=.55\linewidth]{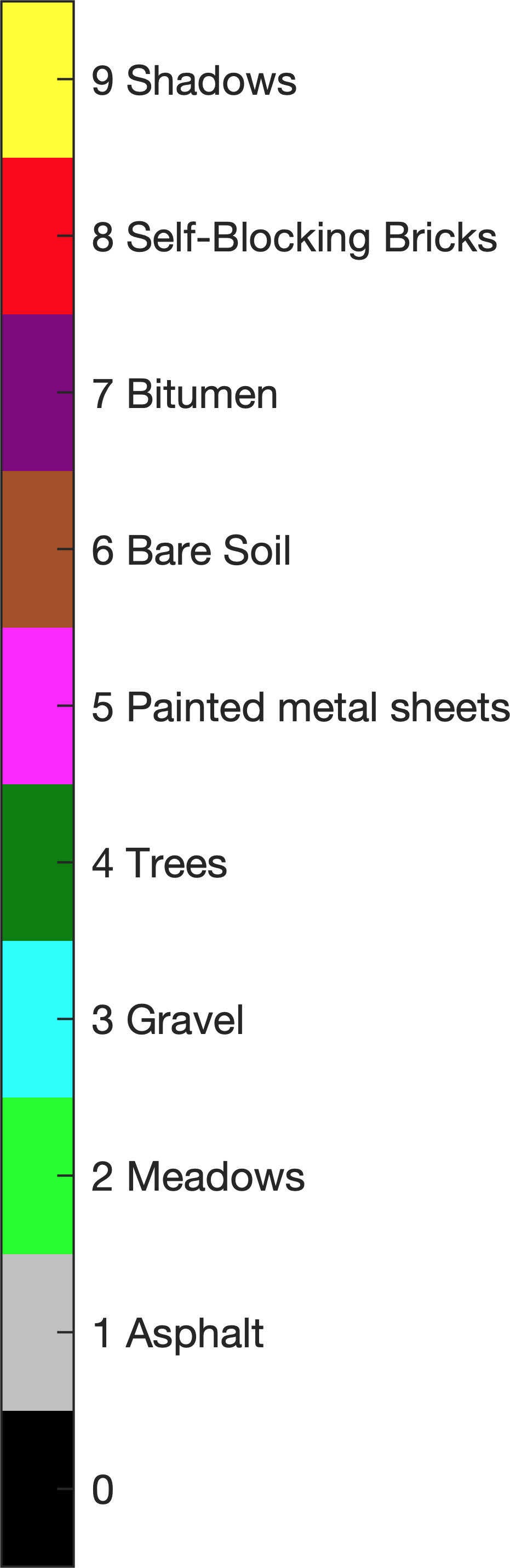}
	\caption{Labels}
\end{subfigure}
\hfill
\begin{subfigure}[b]{.18\textwidth}
	\centering
	\includegraphics[width=.95\linewidth]{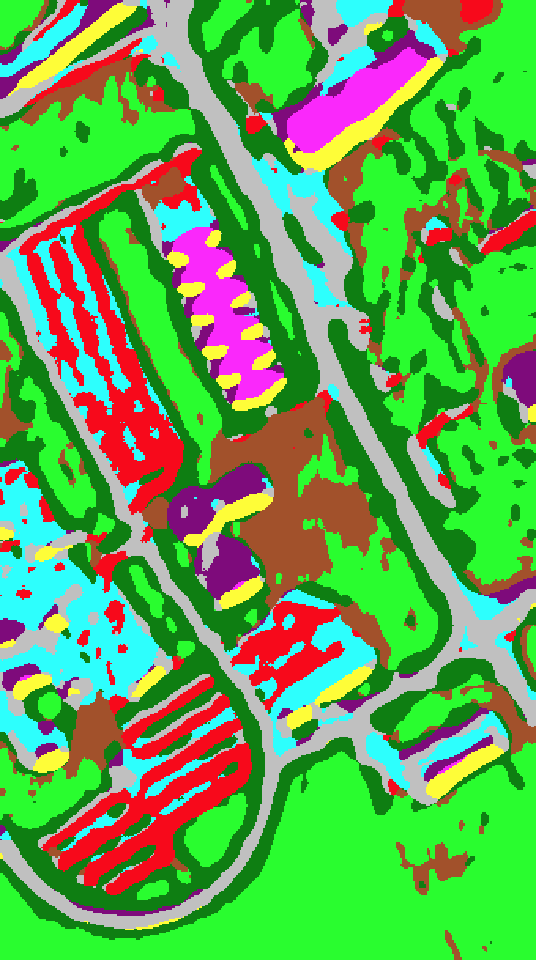}
	\caption{DFFN}
	\label{fig:PUdffnrandom}
\end{subfigure}~
\begin{subfigure}[b]{.18\textwidth}
	\centering
	\includegraphics[width=.95\linewidth]{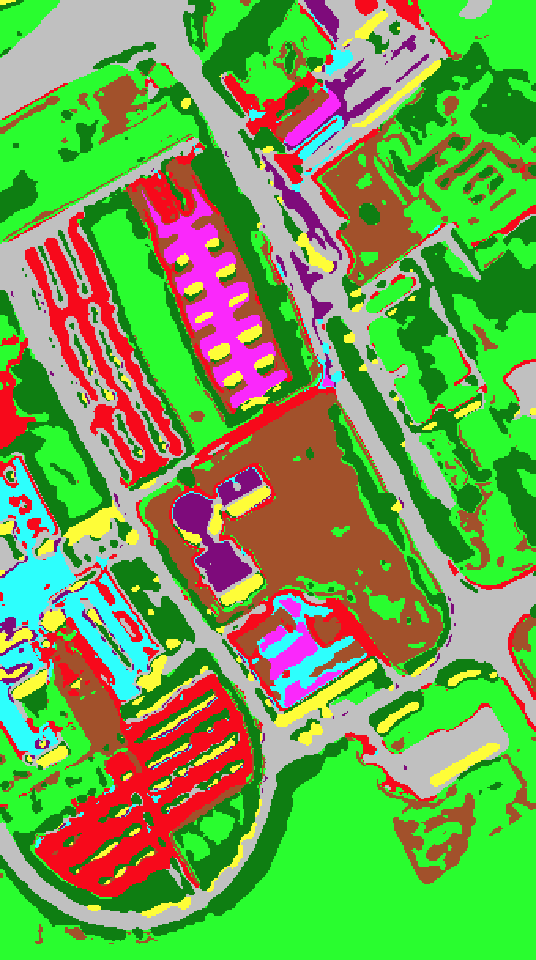}
	\caption{EAP}
	\label{fig:PUeaprandom}
\end{subfigure}~
\begin{subfigure}[b]{.18\textwidth}
	\centering
	\includegraphics[width=.95\linewidth]{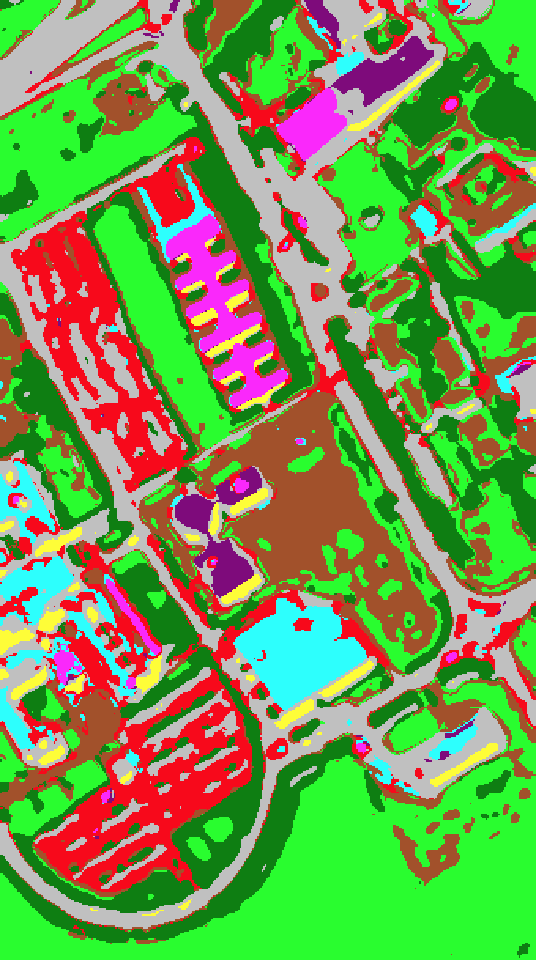}
	\caption{3D WST}
	\label{fig:PUfstrandom}
\end{subfigure}~
\begin{subfigure}[b]{.18\textwidth}
	\centering
	\includegraphics[width=.95\linewidth]{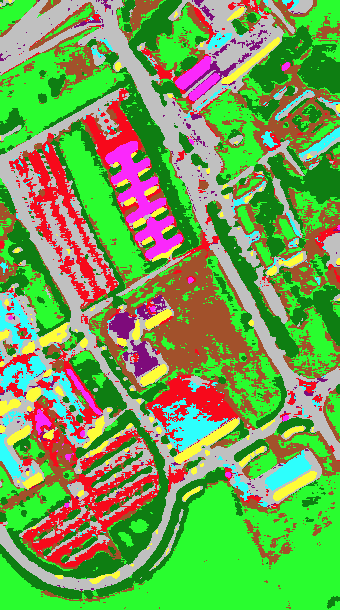}
	\caption{3D Gabor}
	\label{fig:PUfstrandom}
\end{subfigure}~
\begin{subfigure}[b]{.18\textwidth}
	\centering
	\includegraphics[width=.95\linewidth]{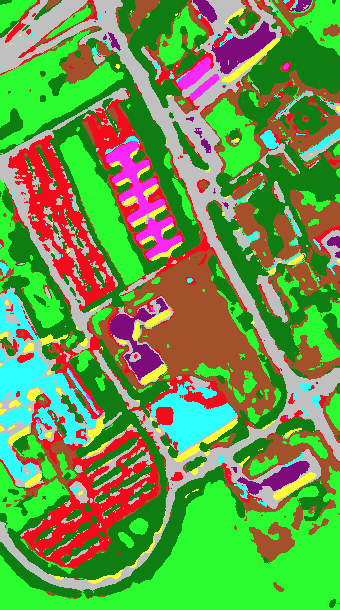}
	\caption{3D FST}
	\label{fig:PUfstrandom}
\end{subfigure}
\hfill
\begin{subfigure}[b]{.18\textwidth}
	\centering
	\includegraphics[width=.95\linewidth]{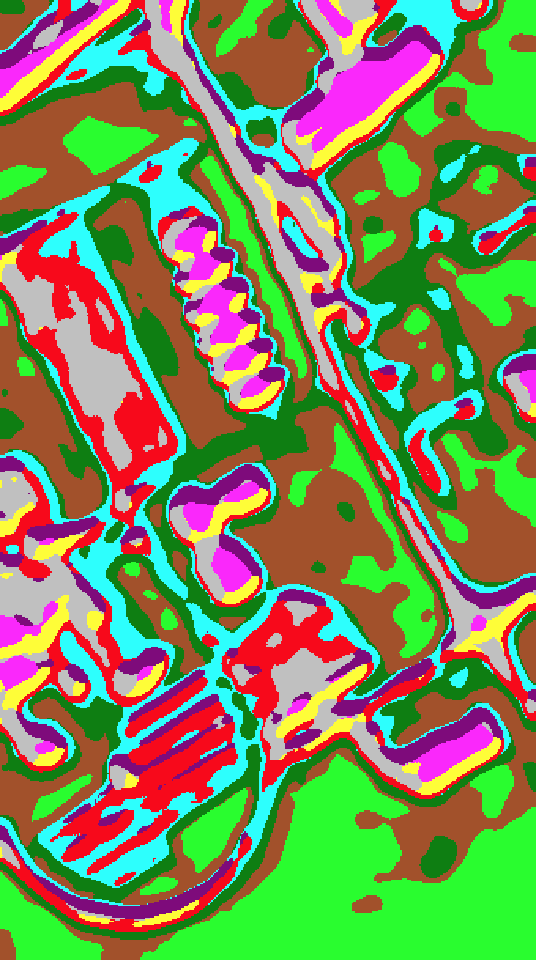}
	\caption{DFFN SSS}
	\label{fig:PUdffnsss}
\end{subfigure}~
\begin{subfigure}[b]{.18\textwidth}
	\centering
	\includegraphics[width=.95\linewidth]{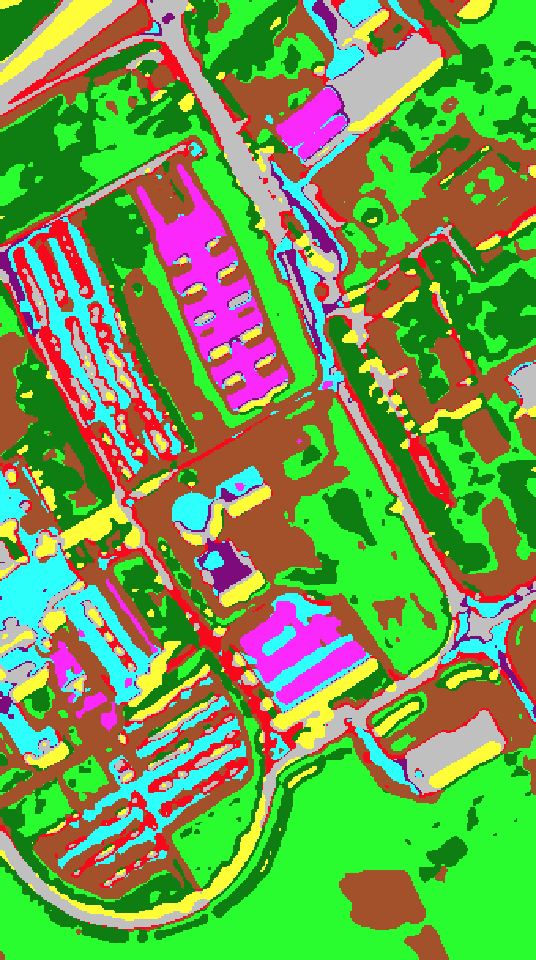}
	\caption{EAP SSS}
	\label{fig:PUeapsss}
\end{subfigure}~
\begin{subfigure}[b]{.18\textwidth}
	\centering
	\includegraphics[width=.95\linewidth]{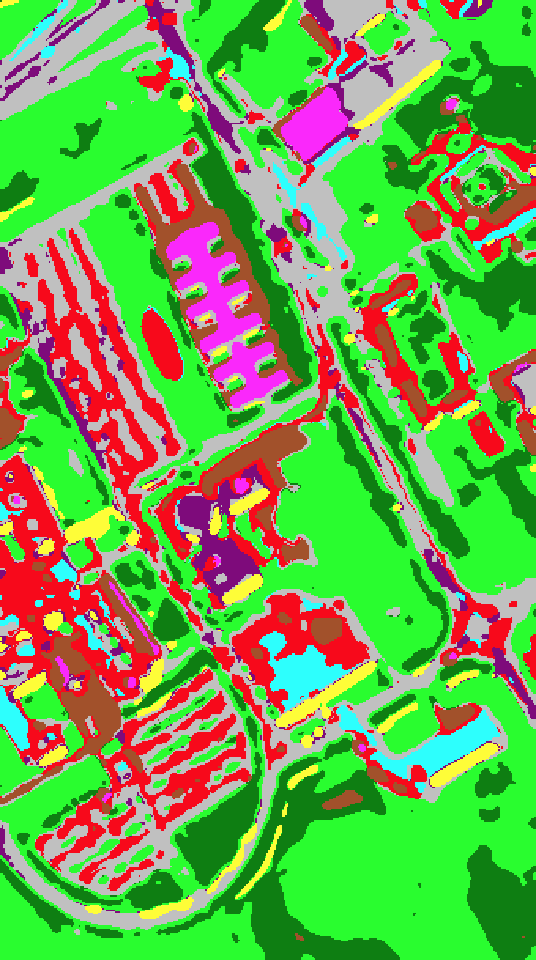}
	\caption{3D WST SSS}
	\label{fig:PUfstsss}
\end{subfigure}~
\begin{subfigure}[b]{.18\textwidth}
	\centering
	\includegraphics[width=.95\linewidth]{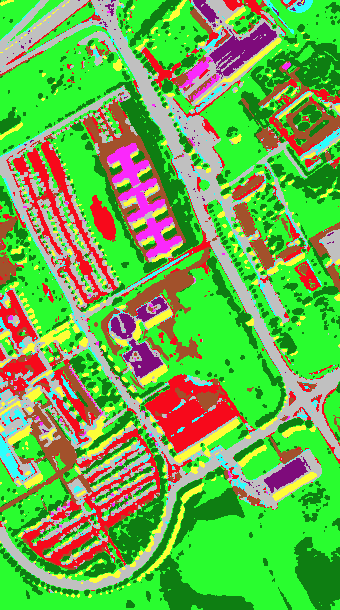}
	\caption{3D Gabor SSS}
	\label{fig:PUfstsss}
\end{subfigure}~
\begin{subfigure}[b]{.18\textwidth}
	\centering
	\includegraphics[width=.95\linewidth]{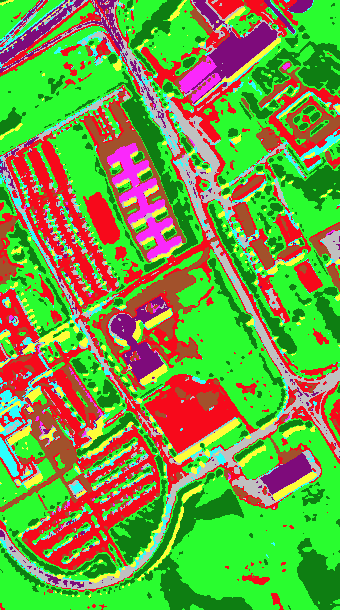}
	\caption{3D FST SSS}
	\label{fig:PUfstsss}
\end{subfigure}
\hfill

\caption{Full classification on PaviaU. The middle row is trained with 2\% of training data with samples randomly selected. The classification performance for these methods is in \Cref{fig:Paviaperf}. The bottom row is trained with 90 samples per class (1.8\% of training data) Strictly Site Specific. The best performing trial by classification accuracy per model is plotted. Though the training set sizes are roughly the same in size, distributed versus site specific training shows a big difference in full classification maps, and an even bigger performance in classification accuracy: with distributed sampling the methods perform with over 90\% OA versus with SSS the methods perform with less than 60\% OA (see \Cref{fig:Paviaperf}). The courtyard in the yellow rectangle, the asphalt gap between the painted metal sheets of the red rectangle, and the shadows in the blue rectangle are well preserved in the 3D FST images.}
\label{fig:PUall}
\end{figure*}

In this section we perform a robustness analysis on the hyperparameter values of 3D FST and discuss the performance impact of various choices of spatial and spectral window sizes.
The promise of deep learning methods like DFFN or EAP is that they learn their parameters to adjust to the dataset. With 3D FST a choice of several parameters serve the same crucial function before the features extracted with it are fed to a learning classifier (a linear SVM in this paper).
When using the Fourier scattering transform on HSI data, there are several parameter choices that impact the effectiveness of our method. Recall that $g,g',g''$ are the window functions used in each layer of the 3D FST, where $M,M',M''$ are the size of their supports, and $P,P',P''$ are the downsampling parameters.
The most impactful parameters to choose are $M,M',M''$. A large spatial support of the FST windows will lead to larger spatial features being extracted, as well as more spatial blurring.
The size of the FST window supports in the spectral domain will also lead to a greater range of frequency features extracted in the spectrum and also more averaging in the spectrum.
These hyperparameters can be selected to best exploit the different physical conditions that a dataset was collected according to, like meters/pixel and sampling rate in the spectral domain.
We would like to choose $M,M',M''$ once per dataset, and a natural approach is to choose the $M,M',M''$ that yield the best classification score on a validation set.
However, we also have to be careful when we create our validation set to ensure that $M,M',M''$ will not be influenced by artificial biases created by picking this validation set.

Uncontrolled random sampling to create a training set of HSI pixels has been previously observed to create a positive bias to methods that use larger spatial neighborhoods \cite{liang_sampling,acquarelli_spectral-spatial_2018}.
For methods that extract features from neighboring pixels to perform a classification on a single test pixel, if the neighborhood includes a pixel that was present in the training set, then the training and testing features will overlap in the spatial domain. A larger neighborhood used to extract features will lead to a greater overlap between training and testing neighborhoods and thus a greater similarity between training and testing features. Thus sampling a HSI image randomly and choosing the spatial window size to use for feature extraction based on classification performance can lead to an artificial preference for large windows.
This exact effect is displayed in \Cref{fig:gridsearches}. In the first row of \Cref{fig:gridsearches} we see a trend that larger spatial windows lead to higher overall accuracy on the test set when the training set is randomly distributed across the image.

\begin{figure}[!t]
\centering
\begin{subfigure}[b]{.24\textwidth}
	\centering
	\includegraphics[width=.95\linewidth]{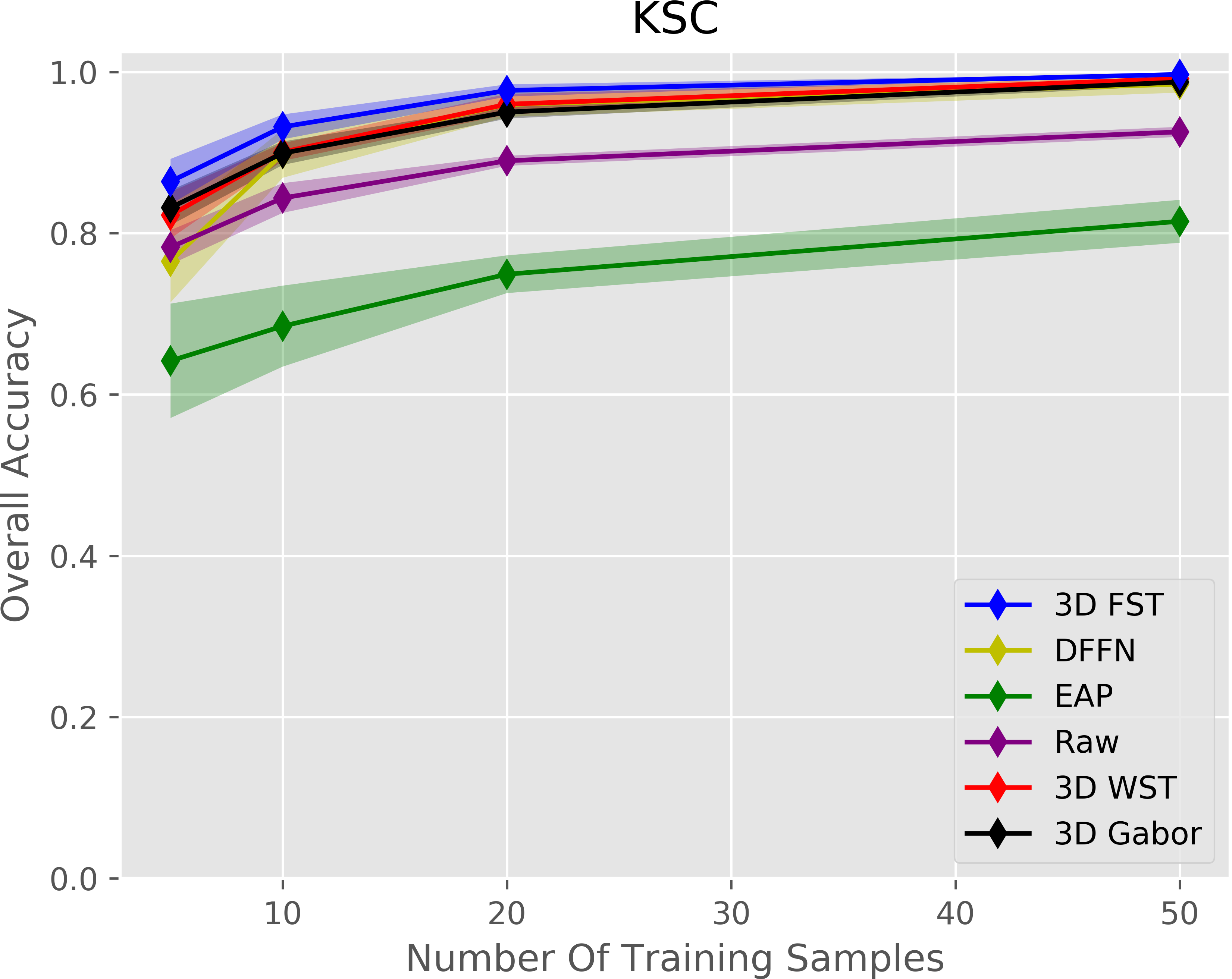}
\end{subfigure}%
\begin{subfigure}[b]{.24\textwidth}
	\centering
	\includegraphics[width=.95\linewidth]{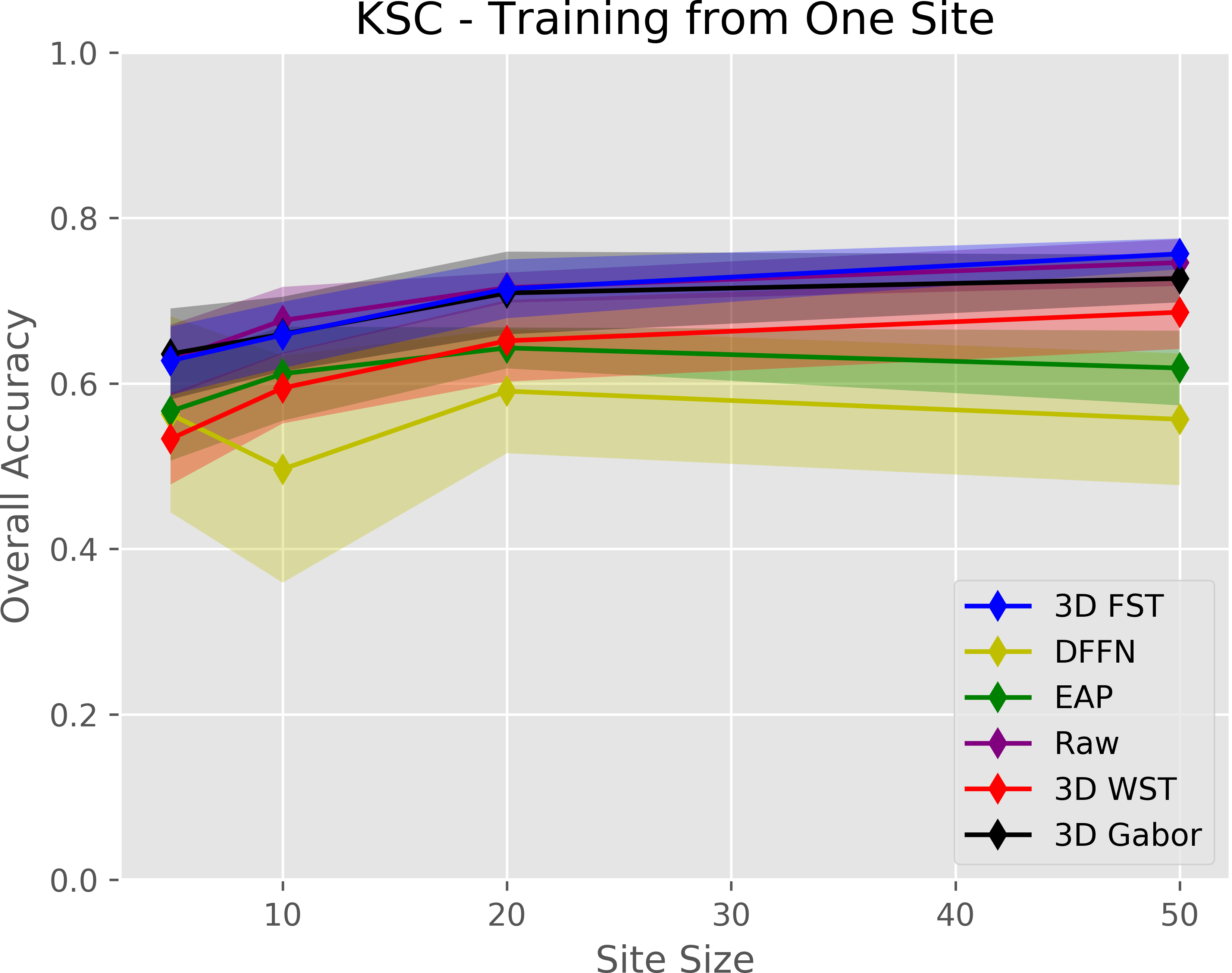}
\end{subfigure}%
\hfill
\caption{Performance of the 6 implemented methods on randomly sampled training sets from KSC. Performance converges quickly at just 50 training samples per class, but FST keeps an edge. For SSS datasets the Fourier features of 3D FST and 3D Gabor methods lead to the best performance.}
\label{fig:KSCperf}
\end{figure}

\begin{figure}[!t]
\centering
\begin{subfigure}[b]{.24\textwidth}
	\centering
	\includegraphics[width=.95\linewidth]{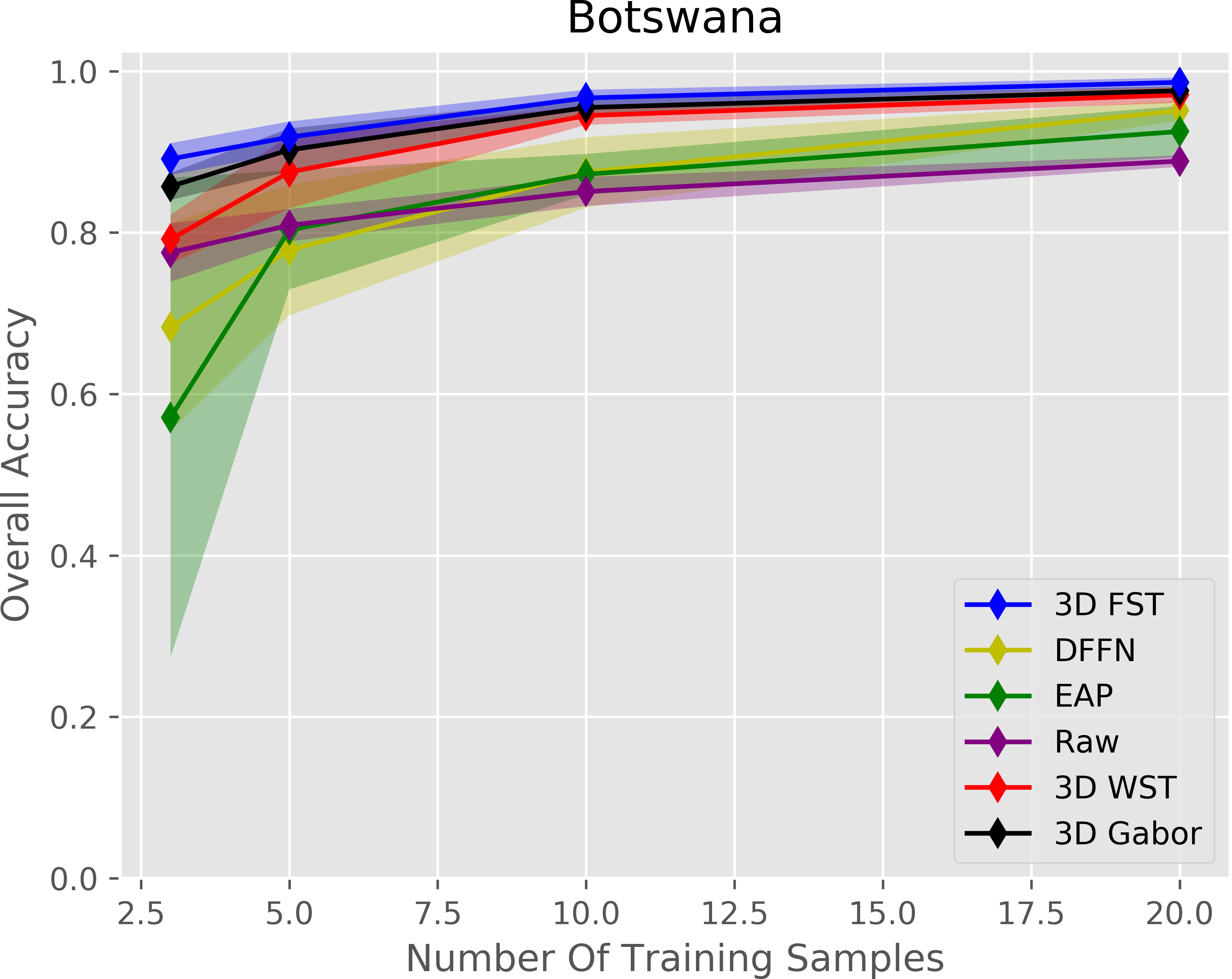}
\end{subfigure}%
\begin{subfigure}[b]{.24\textwidth}
	\centering
	\includegraphics[width=.95\linewidth]{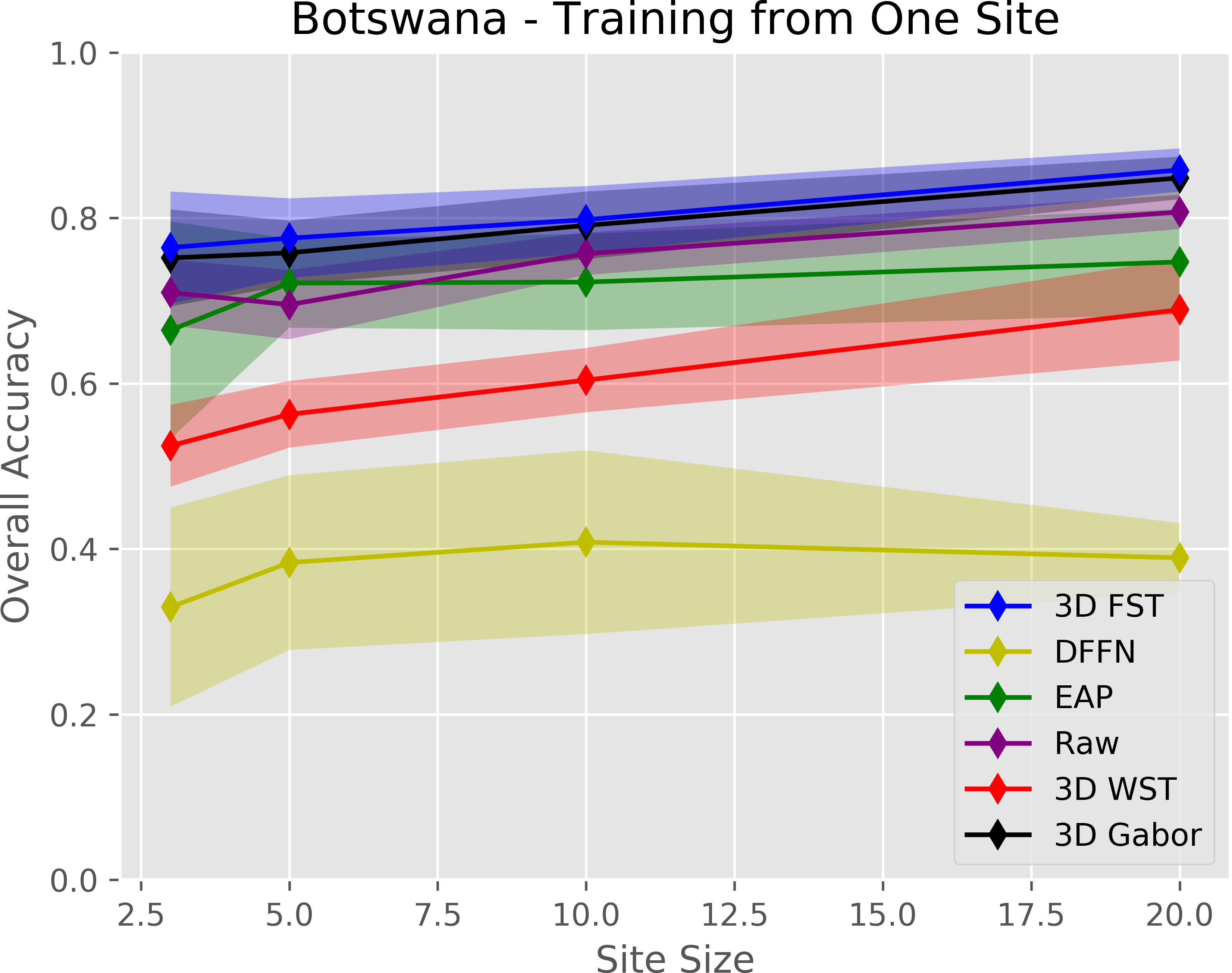}
\end{subfigure}%
\hfill
\caption{Performance of the 6 implemented methods on randomly sampled training sets from Botswana. Performance is very high on the very small datasets tested. The SSS sampling strategy has the biggest spread of any dataset, but the results are consistent in that 3D FST, 3D Gabor, and Raw features are all competitive with each other.}
\label{fig:Botsperf}
\end{figure}

To limit this preference for larger spatial neighborhoods we can compare to a sampling strategy similar to \cite{liang_sampling} for picking the training set.
In contrast to sampling randomly from the labels to create a training set like in \Cref{fig:PUdistributed}, we may sample in a local manner according to connected components as discussed in \cite{liang_sampling}. To create a training set, we can instead pick a single pixel per class, and add pixels to our training set only when they are directly adjacent to already selected pixels of the same class. As long as we do not request too many pixels to be in our training set, this method leads to a single site of training pixels per class, and we refer to such a training set as {\it strictly site specific} (SSS). The largest such training set we create for this paper (90 samples per class) is shown in \Cref{fig:PUsss} (90 samples per class is about 1.9\% of training data for PaviaU). Note how when sampling 2\% of pixels randomly in \Cref{fig:PUdistributed} a 1-KNN classifier on the pixel coordinates will perform at 93\% accuracy as shown in \Cref{fig:PUknn} while for a SSS dataset of equal size the same type of classifier will perform at 22\% accuracy. This shows that neighborhood overlap in feature generation has the potential to inflate classification accuracy by a large degree for the randomly sampled training set, but not for the strictly site specific training set.

The grid searches over hyperparameters $M,M',M''$ that we perform are shown in \Cref{fig:gridsearches}, the first row contains grid search results on randomly distributed training data, and the second row contains grid search results on SSS datasets which limit unnecessary spatial bias. For the SSS datasets, for PaviaU we build training sets with 9 sites of size 90 (1.9\% of training data), for Indian Pines we use 16 sites sized such that 10\% of the data is used for training, for KSC 13 sites of size 20 (5\% of training data), and for Botswana 14 sites of size 20 (9\% of training data). The gridsearches for randomly distributed data in the first row of \Cref{fig:gridsearches} use the same amount of data as the SSS gridsearches. For each point in the grid we perform 10 trials each on a different training set, and plot the mean accuracy. 3,200 experiments are summarized in \Cref{fig:gridsearches}.
Each selection of $M,M',M''$ has 6 degrees of freedom (we use square spatial windows), and each individual window we test has a support of 3,5,7, or 9 for randomly distributed data, and 1,3,5,7 for SSS data.
We plot the spatial receptive field (proportional to $M_1 + M_1' + M_1''$) and the spectral receptive field (proportional to $M_3 + M_3' + M_3''$) against the mean accuracy.
We find that the general shape of the loss surface does not change as we vary the size of the training sets but is smoother for larger training sets.
The best hyperparameters $M,M',M''$ are:

\begin{itemize}
    \item 
	{\it PaviaU}. $7\times7\times5,7\times7\times5,7\times7\times5$ which has a receptive field of 24 m by 179 nm (19 spatial samples by 43 spectral samples).
	For a strictly site specific setting for this dataset we choose the hyperparameters
	$3\times3\times7,3\times3\times7,3\times3\times7$ which has a receptive field of 9 m by 254 nm (7 spatial samples by 61 spectral samples).
	\item 
	{\it Indian Pines}. $9\times9\times7,9\times9\times7,9\times9\times7$ which has a receptive field of 92 m by 571 nm (25 spatial samples by 61 spectral samples).
	For a strictly site specific setting for this dataset we choose the hyperparameters
	$1\times1\times5,1\times1\times5,1\times1\times5$ which has a receptive field of 3 m by 403 nm (1 spatial samples by 43 spectral samples).
	\item
	{\it KSC}. $7\times7\times3,7\times7\times3,7\times7\times3$ which has a receptive field of 342 m by 234 nm (19 spatial samples by 25 spectral samples).
	For a strictly site specific setting for this dataset we choose the hyperparameters
	$7\times7\times3,1\times1\times3,1\times1\times3$ which has a receptive field of 126 m by 234 nm (7 spatial samples by 25 spectral samples).
	\item 
	{\it Botswana}. $9\times9\times7,5\times5\times7,5\times5\times7$ which has a receptive field of 510 m by 529 nm (17 spatial samples by 61 spectral samples).
	For a strictly site specific setting for this dataset we choose the hyperparameters
	$5\times5\times3,5\times5\times3,5\times5\times3$ which has a receptive field of 390 m by 216 nm (13 spatial samples by 25 spectral samples).
	
\end{itemize}

Some general patterns we observe are that a larger spectral receptive field is preferred for randomly distributed training, and that the spatial window should only decrease in size in the network.
In the results of the grid search, we observe the intuitive trade off in making the spatial receptive larger: that noise is reduced, but as a result of blurring the ability to resolve details is lost, hence the optimal value is a spatial window that is neither too large nor too small.
In the SSS gridsearches we see that the largest spatial receptive field does not perform the best, hence we conclude that picking parameters with such a search yields a feature extractor that avoids unnecessary spatial bias.
This is in line with our expectations since in a randomly sampled training dataset like in \Cref{fig:PUdistributed}, for every test pixel, as we increase the size of the spatial window in which to extract features, the number of training point features that we mix with the test point features grows. But for the SSS training dataset like in \Cref{fig:PUsss} this is not true for most test points which are not near the single class sites.

The most dramatic difference in hyperparameters for the two sampling strategies is for Indian Pines, where the largest possible spatial window performs best with randomly distributed data, and the smallest possible spatial window performs best with single site data, yet number of spectra samples used does not dramatically differ. This is likely to do with the density of Indian Pines: 49\% of pixels have labels (compared with 21\%, 2\%, and $>1$\% for PaviaU, KSC, and Botswana) and because of the human-planned agricultural setting all the pixels belonging to a class are often adjacent. Thus with a randomly distributed training set, growing the window of features around a test pixel often incorporates features from training pixels nearby. At the same time for strictly site specific training sets, growing a window around a test pixel that is not near training pixels can incorporate features from the wrong class since the fields of different crops are nearby.

The receptive field sizes that our grid search picks are generally middle of the range compared to the literature. The EAP method we compare to has a smaller spatial receptive field of 9 while the DFFN has a larger receptive field of up to 25, and the 3D WST has a similar receptive field of 19.

Looking at \Cref{fig:gridsearches} performance varies much less along the spectral dimension than the spatial, and a middle-ground value for the spectral window is often picked over an extreme value. For PaviaU, which has the fewest spectral bands of all the datasets and most diverse material classes, switching from larger to smaller windows for randomly distributed data versus SSS data causes a preference for a larger spectral window, perhaps to compensate for the decreased spatial features. For Indian Pines where 14 of 16 classes are types of vegetation, we see that the best performing spatial windows have similar performance in the spectral dimension. For KSC and Botswana a smaller spectral window is preferred in general, perhaps also because of the material similarity of most classes.
We also note that for all the best hyperparameters that $M_3'=M_3''=M_3'''$, that the spectral receptive field should be equally distributed throughout the layers of scattering.


Regarding the other hyperparameters, we keep them constant throughout our experiments.
Higher downsampling improves the speed of our method but decreases the performance, so we fix $P,P',P''$ to be as low as possible without making our method prohibitively slow (also note that the higher the downsampling the higher the receptive field of the network).
Setting $P=M-2$, $P'=M'-2$, and $P''=M''-2$ achieves this balance. For simplicity, for $g=g'=g''$ we use the rectangular window. The parameters listed in this section are used for 3D FST experiments throughout the rest of the paper, though we do note that if we only wanted to improve classification accuracy for randomly sampled distributed training sets, the larger the spatial window the better the accuracy, but only for the reasons of spatial bias noted in this section. The best representation of HSI data is achieved with the $M,M',M''$ triplets listed above.

\subsection{Analysis}

\begin{figure*}[ht!]
\begin{minipage}[b][9.5cm][t]{.9\textwidth}
\centering
\begin{subfigure}[b]{.19\textwidth}
	\centering
	\includegraphics[width=.99\linewidth]{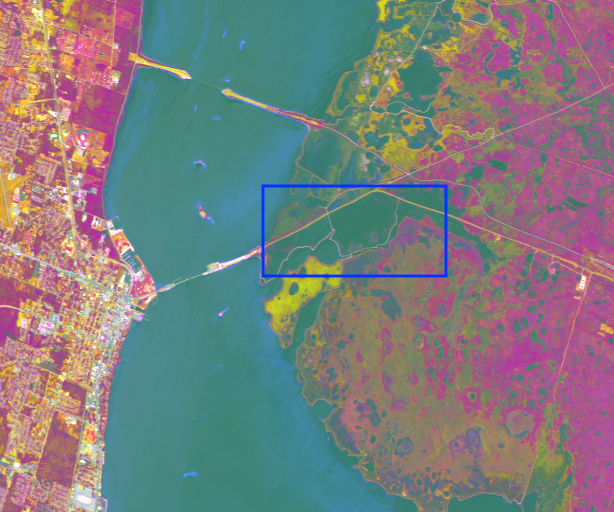}
	\caption{False Color}
	\label{fig:KSCcolor}
\end{subfigure}~
\begin{subfigure}[b]{.19\textwidth}
	\centering
	\includegraphics[width=.99\linewidth]{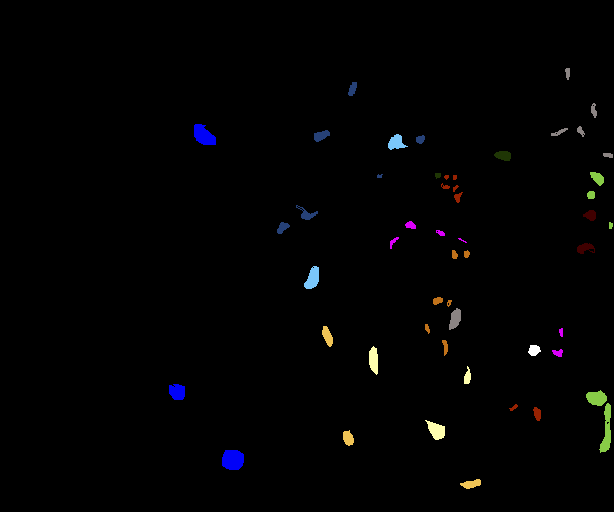}
	\caption{Ground Truth}
	\label{fig:KSCgt}
\end{subfigure}~
\begin{subfigure}[b]{.19\textwidth}
	\centering
	\includegraphics[width=.99\linewidth]{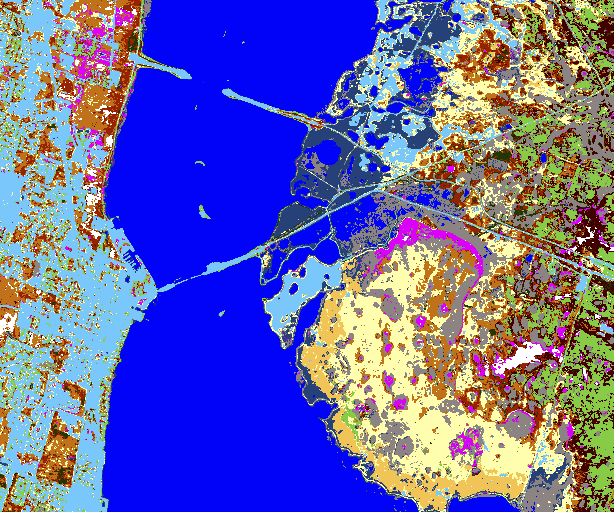}
	\caption{Raw}
\end{subfigure}~
\begin{subfigure}[b]{.19\textwidth}
	\centering
	\includegraphics[width=.99\linewidth]{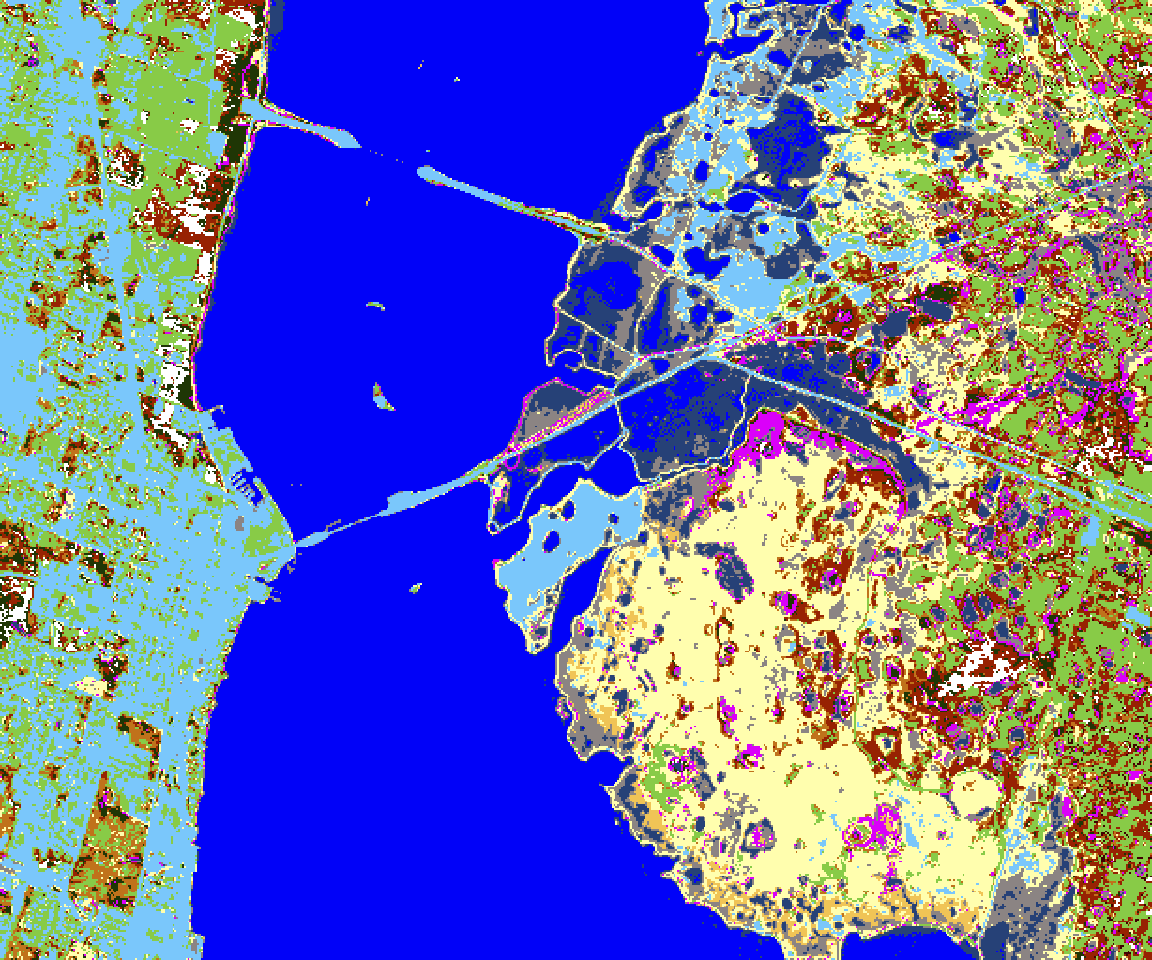}
	\caption{Raw SSS}
\end{subfigure}
\hfill
\begin{subfigure}[b]{.19\textwidth}
	\centering
	\includegraphics[width=.99\linewidth]{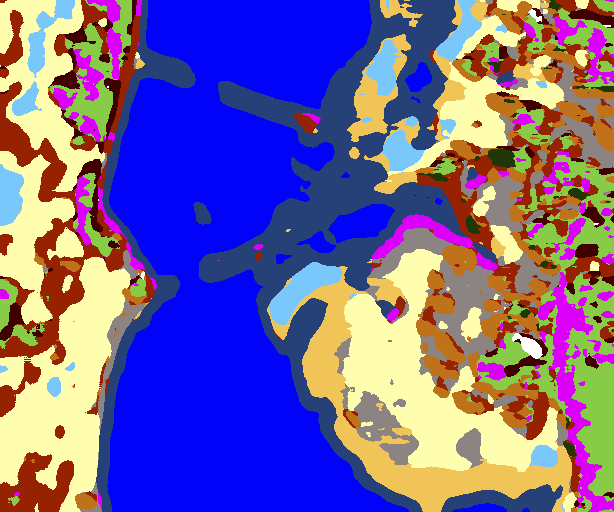}
	\caption{DFFN}
\end{subfigure}~
\begin{subfigure}[b]{.19\textwidth}
	\centering
	\includegraphics[width=.99\linewidth]{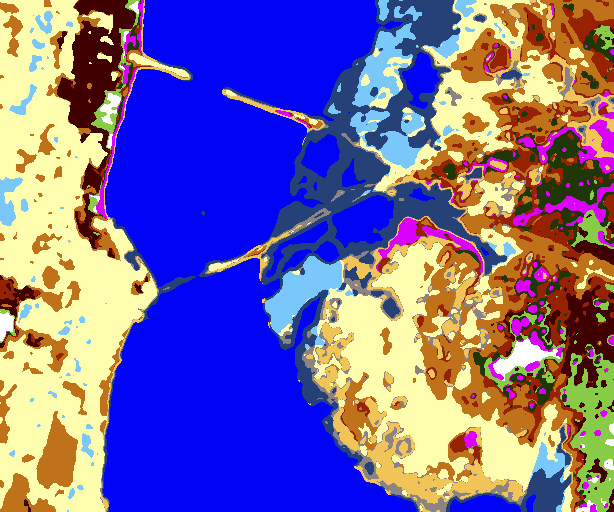}
	\caption{EAP}
\end{subfigure}~
\begin{subfigure}[b]{.19\textwidth}
	\centering
	\includegraphics[width=.99\linewidth]{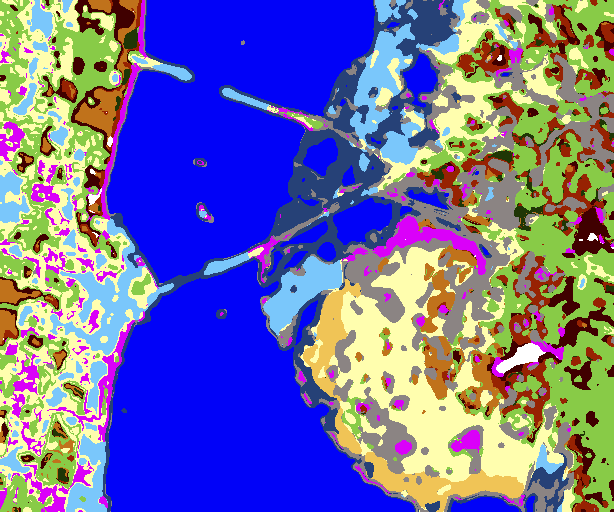}
	\caption{WST}
\end{subfigure}~
\begin{subfigure}[b]{.19\textwidth}
	\centering
	\includegraphics[width=.99\linewidth]{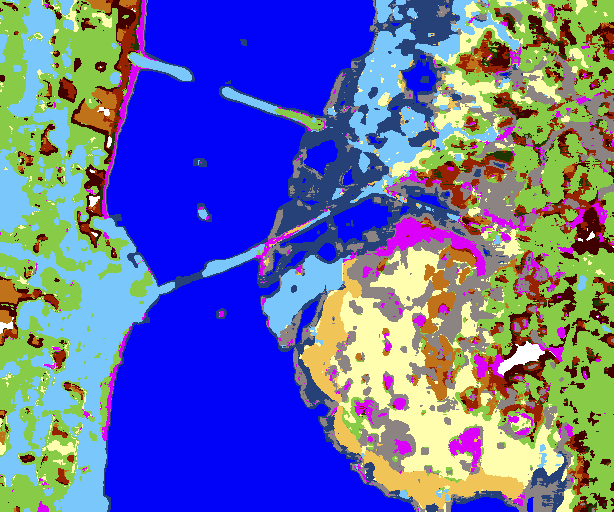}
	\caption{Gabor}
\end{subfigure}~
\begin{subfigure}[b]{.19\textwidth}
	\centering
	\includegraphics[width=.99\linewidth]{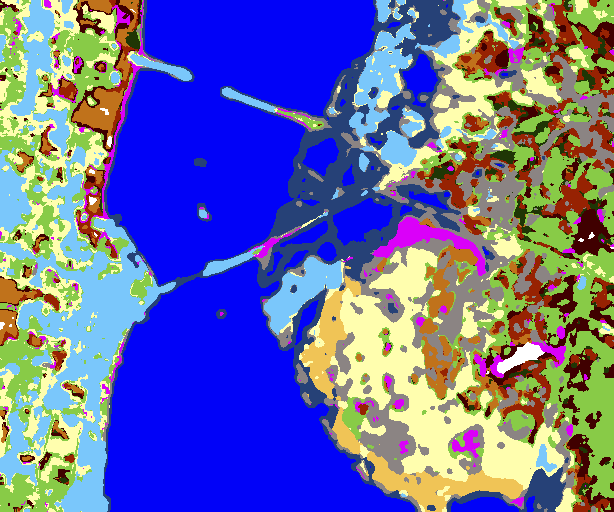}
	\caption{FST}
\end{subfigure}
\hfill
\begin{subfigure}[b]{.19\textwidth}
	\centering
	\includegraphics[width=.99\linewidth]{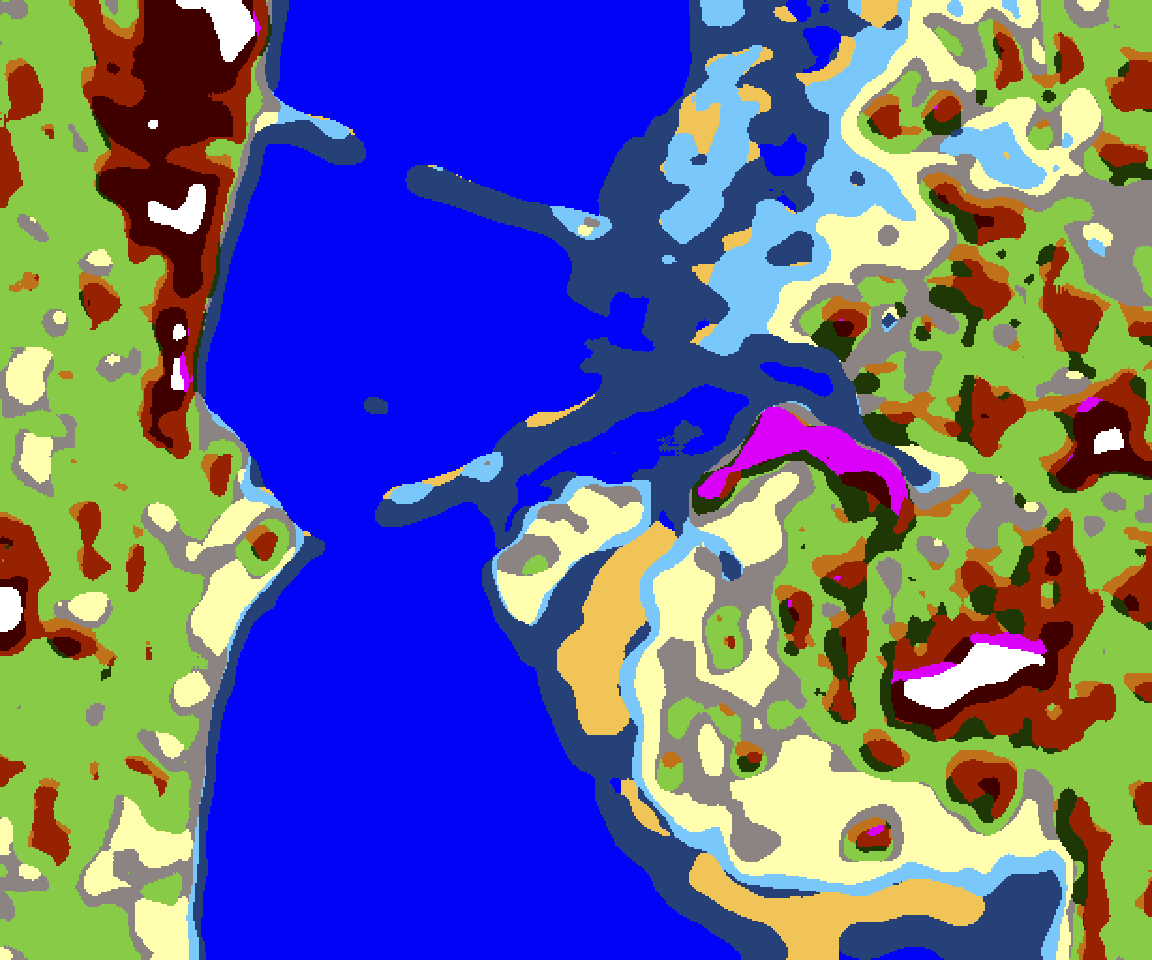}
	\caption{DFFN SSS}
\end{subfigure}~
\begin{subfigure}[b]{.19\textwidth}
	\centering
	\includegraphics[width=.99\linewidth]{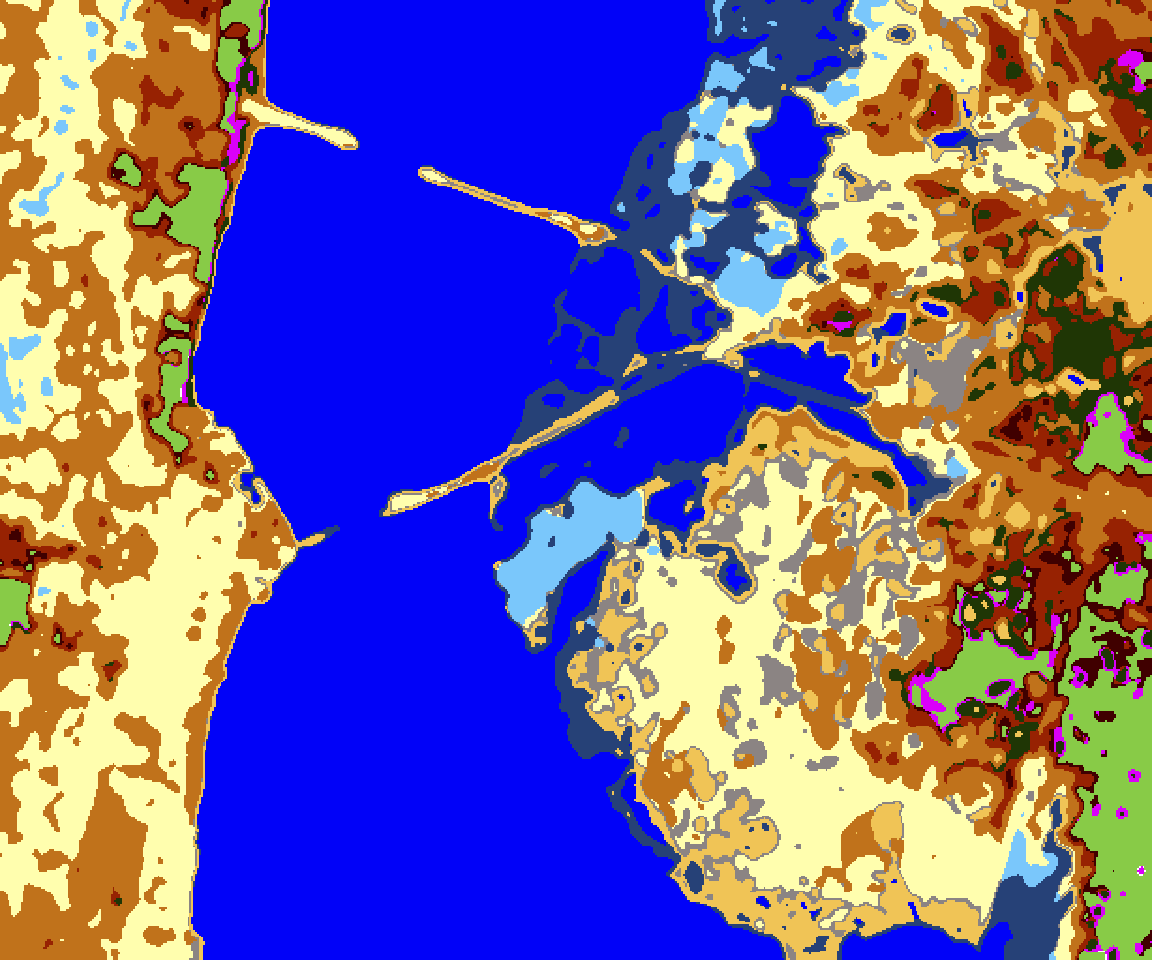}
	\caption{EAP SSS}
\end{subfigure}~
\begin{subfigure}[b]{.19\textwidth}
	\centering
	\includegraphics[width=.99\linewidth]{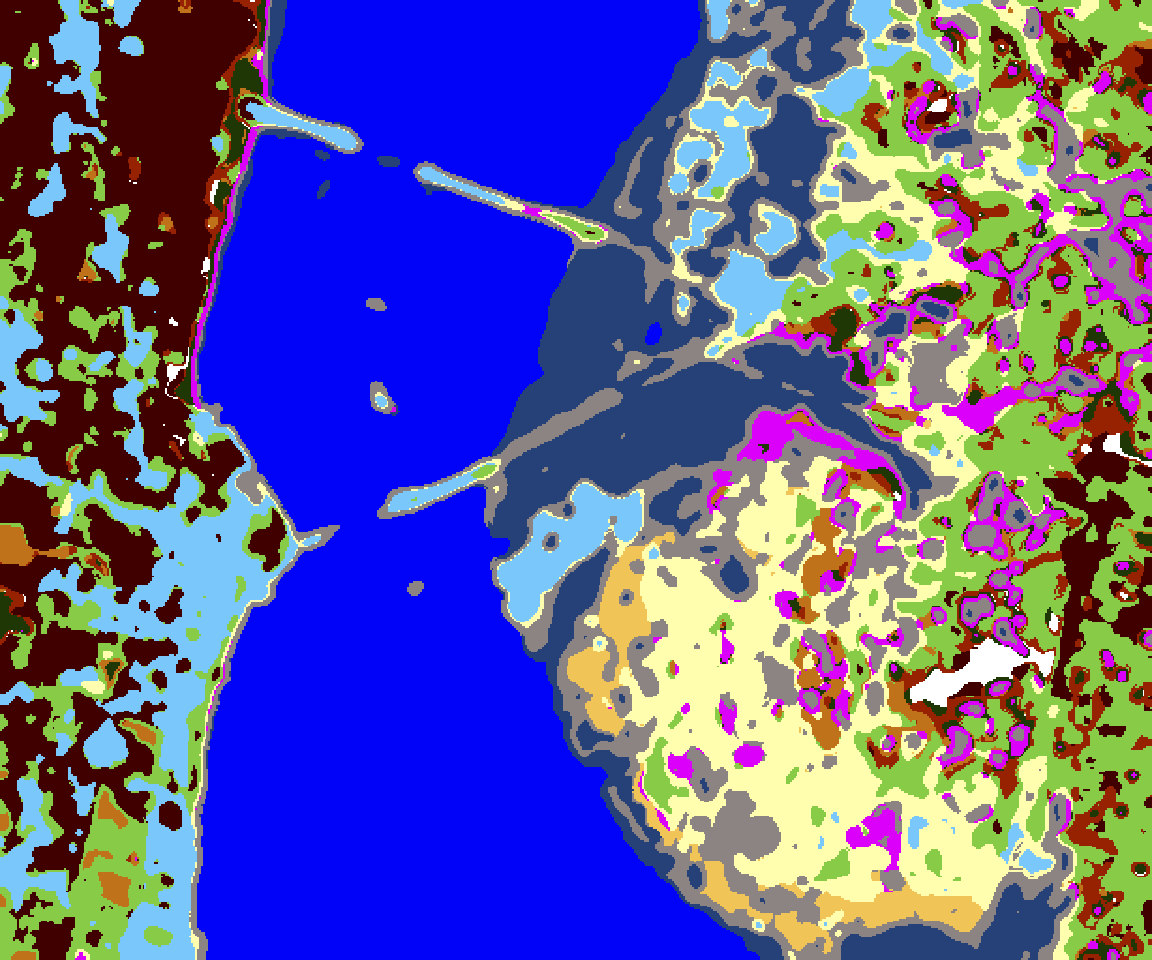}
	\caption{WST SSS}
\end{subfigure}~
\begin{subfigure}[b]{.19\textwidth}
	\centering
	\includegraphics[width=.99\linewidth]{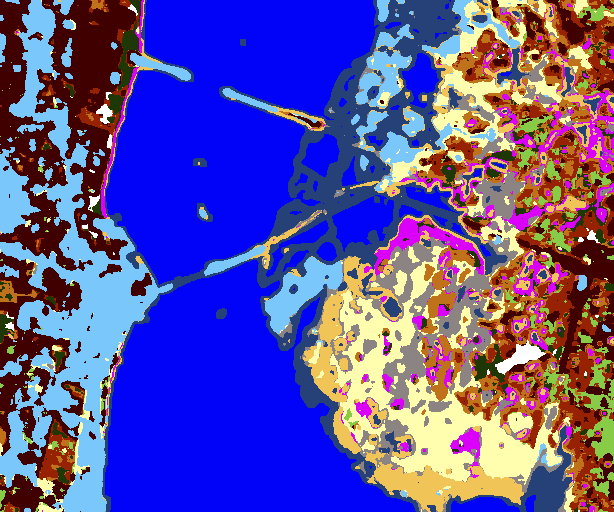}
	\caption{Gabor SSS}
\end{subfigure}~
\begin{subfigure}[b]{.19\textwidth}
	\centering
	\includegraphics[width=.99\linewidth]{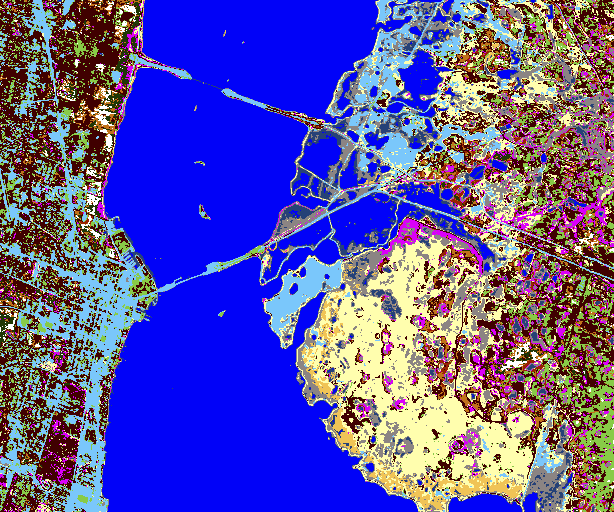}
	\caption{FST SSS}
\end{subfigure}
\hfill
\end{minipage}%
\begin{minipage}[b][9.5cm][t]{.1\textwidth}
\vspace*{\fill}
\begin{subfigure}{\linewidth}
	\centering
	\includegraphics[width=.99\linewidth]{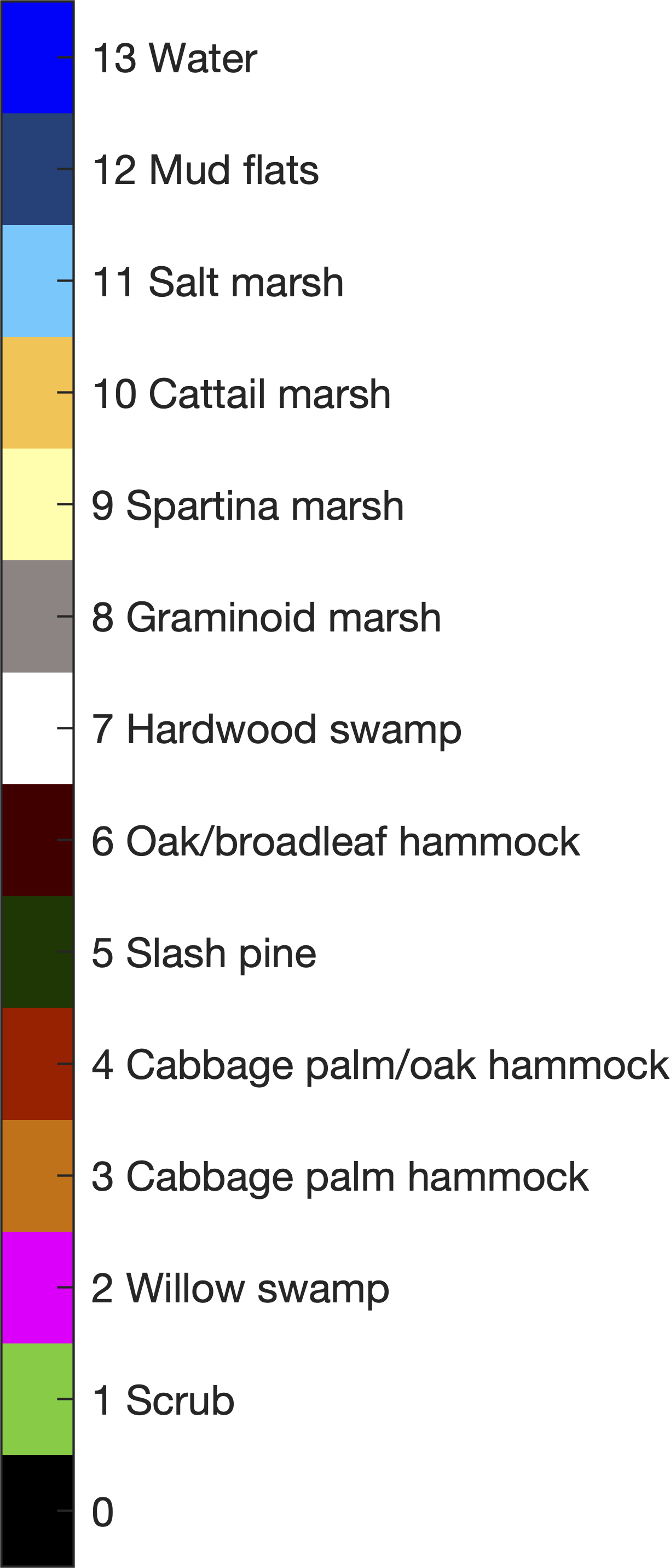}
	\caption{}
\end{subfigure}
\vspace*{\fill}
\end{minipage}

\caption{Full classification on KSC with 50 samples per class sampled randomly and in a Strictly Site Specific way. The best performing trial by classification accuracy per model is plotted.}
\end{figure*}

In this section we compare our methods with both the standard random sampling method and strictly site specific (SSS) training datasets for Indian Pines, PaviaU, KSC, and Botswana. PaviaU and Indian Pines were selected for their popularity, and have a very geometrically structured ground truth. In contrast KSC and Botswana have a more abstract ground truth that has a similar appearance to the SSS sampling strategy.

On Indian Pines our proposed method has the highest classification accuracies for all the training sets used between 5 samples per class to using of 10\% of the data, as seen in \Cref{fig:IPperf}.
In the full classification maps shown in \Cref{fig:IPall} for the highly limited training data scenario of 5 samples per class we can observe many interesting details.
For the randomly distributed datasets, DFFN and 3D FST have the same receptive field size, but DFFN blurs classes together much more, it only begins to perform well and above Raw features at 5\% of training data. When it comes to using the SSS dataset, the spatial blur seems to increase, showing that the large spatial receptive field cannot adapt to SSS data in Indian Pines. The extra layers of convolutions and nonlinearities that 3D FST adds to 3D Gabor pay off in overall accuracy, and leads to slight spatial blurring in the distributed datasets.
For SSS data 3D Gabor and 3D FST are not much different than Raw features, since the hyper-parameter search suggested a spatial window size of 1. EAP has a good balance between blurring and maintaining detail for this dataset and ties with many other methods. 3D WST has the roundest features of all the methods.

We note that for the SSS datasets all methods struggle, and the results are noisy. In some instances the performance can decrease as the site size grows (this happens with the lowest performing methods on SSS data, 3D WST and DFFN, several times). The standard deviation of all the methods for SSS data is also very high and prevents seeing a clear dominating method.
Our explanation for this is that the SSS setting is very challenging because of the great variability between different SSS datasets. Since each SSS dataset contains only a single site per class, there is little information on the overall shape of a class, or number of locations in the image a single class may be, and it is rare for training pixels to be different classes but near.
Though we believe an artificial bias for spatially smoothing feature extractors is diminished with SSS sampling, this dataset construction may leave little in the training data to be exploited other than the spectral information of each pixel, which would explain the consistent high performance of Raw single-pixel spectral features.


The classification accuracy for our proposed method on PaviaU for randomly sampled data is about the same as EAP and 3D WST at 2\% of training data, but performs the best at lower training data percentages and we see more details are preserved in the classification maps in \Cref{fig:PUall}.
We see a similar pattern in overall accuracy in \Cref{fig:IPperf} to Indian Pines, in that the deepest neural network method has the biggest gains as the amount of training data increases.
For PaviaU the filters for 3D FST are a slightly different shape than the $7\times7\times 7$ cube filters of 3D WST, and the two methods perform within a margin of error of each other. However in the full classification maps in \Cref{fig:PUall} we see that 3D WST has more spatial smoothing than 3D FST: details like the separation between the painted metal sheets in the red rectangle and the tree shadows in the blue rectangle are clearer in 3D FST.
Likely because this dataset has more thinly articulated features such as the self-blocking bricks parking lots our gridsearch chose a smaller receptive field for this dataset compared to Indian Pines.
We also notice in \Cref{fig:PUall} that 3D FST best preserves details that we see in \Cref{fig:PUcolor}, which are also clear in the SVM full classification maps, albeit with salt and pepper noise (though Raw misses the Bitumen and Gravel classes).
We also see an improvement over 3D Gabor with the additional layers of 3D FST. The shapes in the distributed datasets become less dithered in the distributed training sets.

\begin{figure*}[ht!]

\begin{minipage}[b][18.5cm][t]{.5\textwidth}
\begin{subfigure}{\linewidth}
	\centering
	\includegraphics[angle=90,width=.99\linewidth]{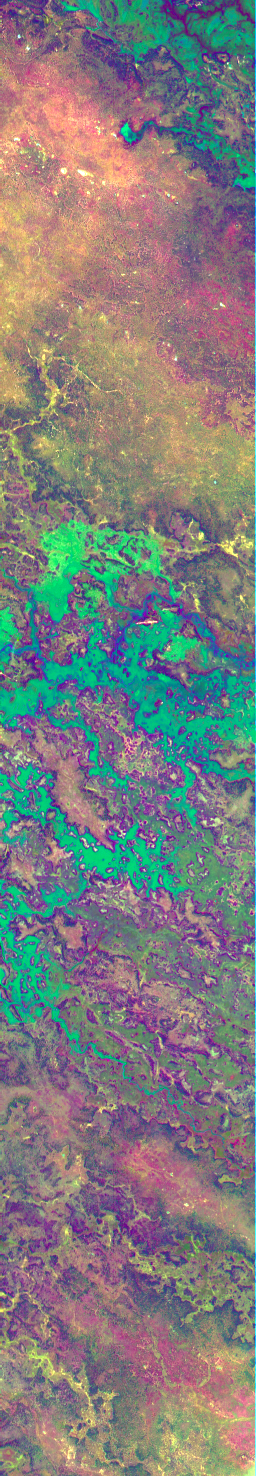}
	\caption{False Color}
\end{subfigure}
\begin{subfigure}{\linewidth}
	\centering
 	\includegraphics[angle=90,width=.99\linewidth]{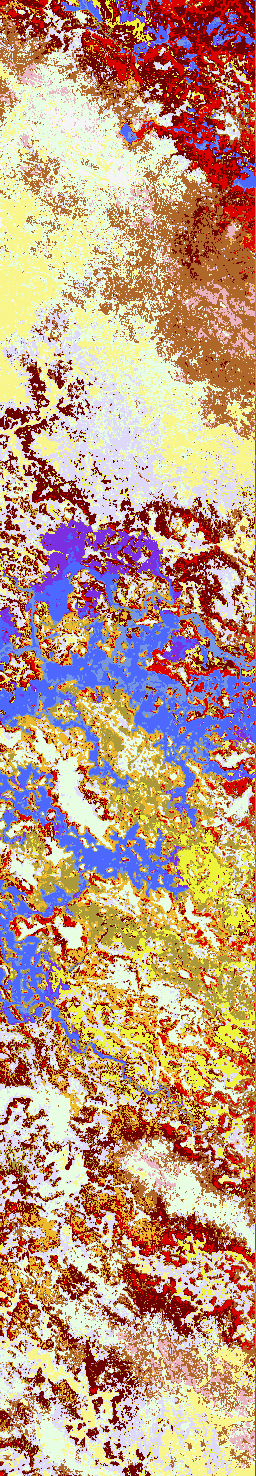}
	\caption{Raw}
\end{subfigure}
\begin{subfigure}{\linewidth}
	\centering
 	\includegraphics[angle=90,width=.99\linewidth]{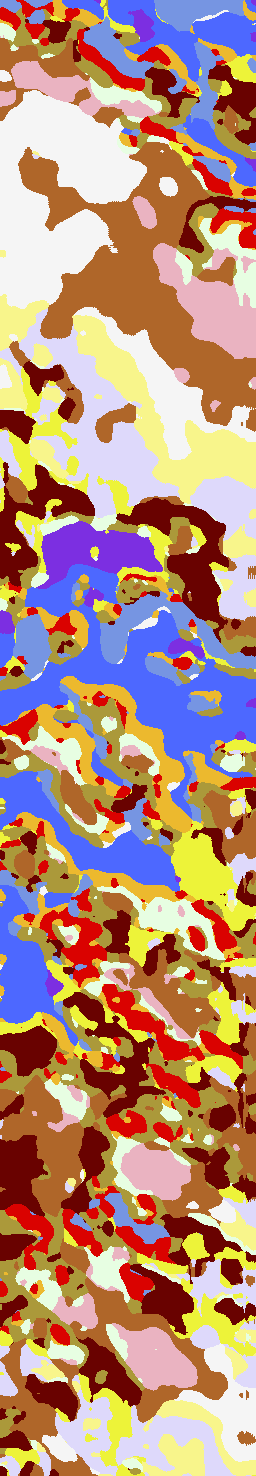}
	\caption{DFFN}
\end{subfigure}
\begin{subfigure}{\linewidth}
	\centering
	\includegraphics[angle=90,width=.99\linewidth]{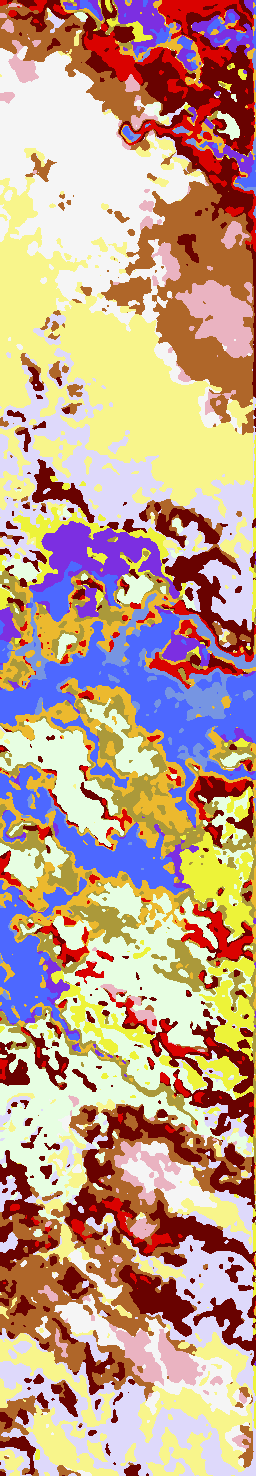}
	\caption{EAP}
\end{subfigure}
\begin{subfigure}{\linewidth}
	\centering
	\includegraphics[angle=90,width=.99\linewidth]{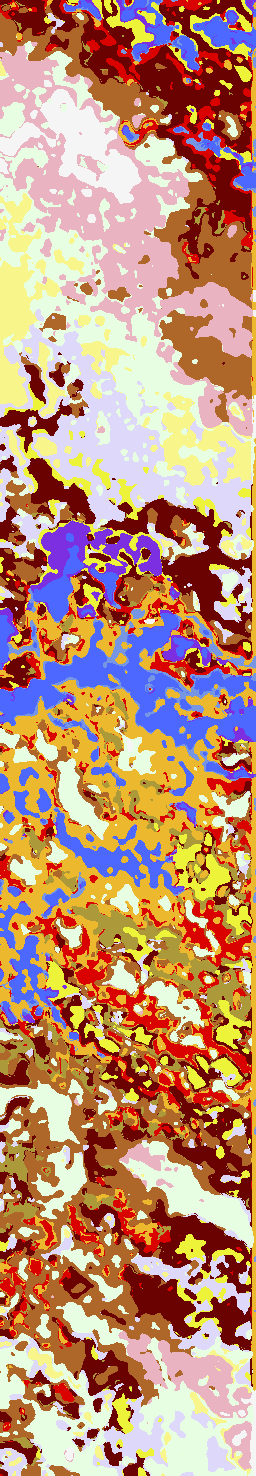}
	\caption{WST}
\end{subfigure}
\begin{subfigure}{\linewidth}
	\centering
	\includegraphics[angle=90,width=.99\linewidth]{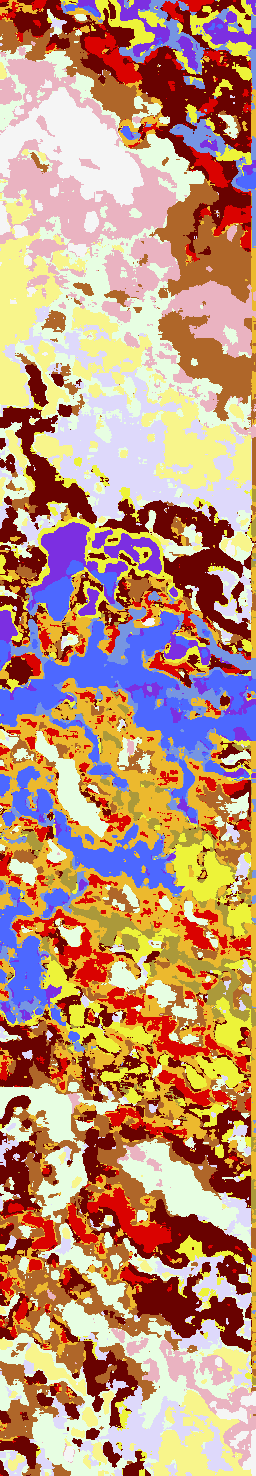}
	\caption{3D Gabor}
\end{subfigure}
\begin{subfigure}{\linewidth}
	\centering
	\includegraphics[angle=90,width=.99\linewidth]{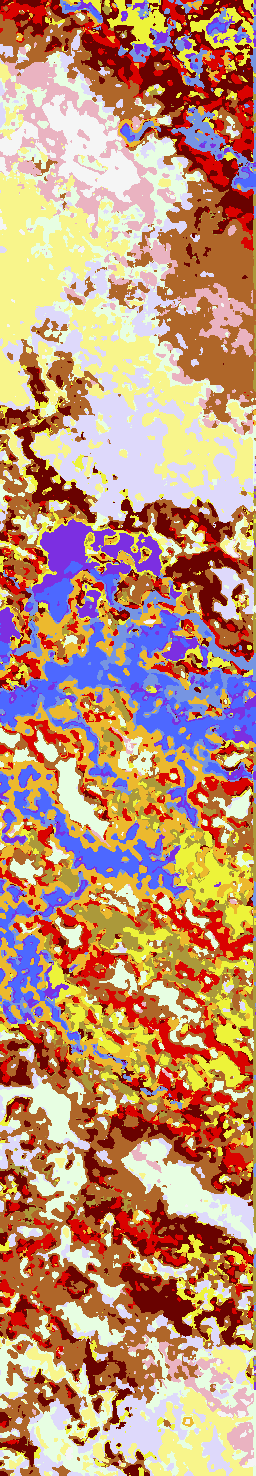}
	\caption{3D FST}
\end{subfigure}
\end{minipage}%
\begin{minipage}[b][18.5cm][t]{.5\textwidth}
\begin{subfigure}{\linewidth}
	\centering
	\includegraphics[angle=90,width=.99\linewidth]{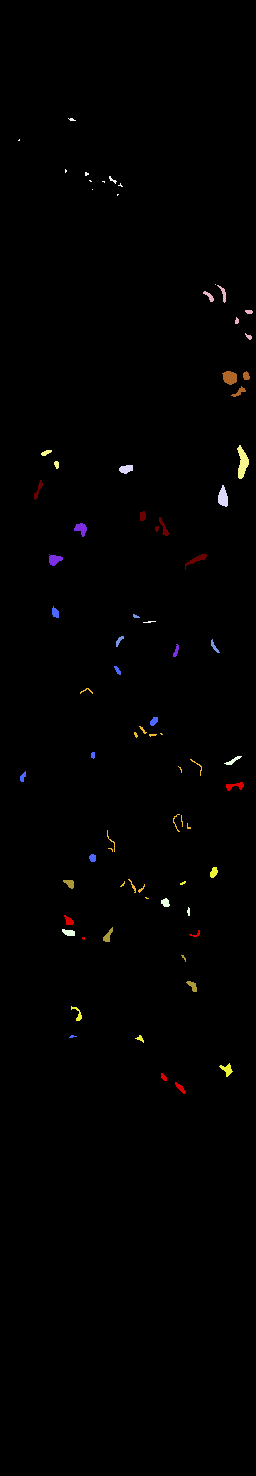}
	\caption{Ground Truth}
\end{subfigure}
\begin{subfigure}{\linewidth}
	\centering
 	\includegraphics[angle=90,width=.99\linewidth]{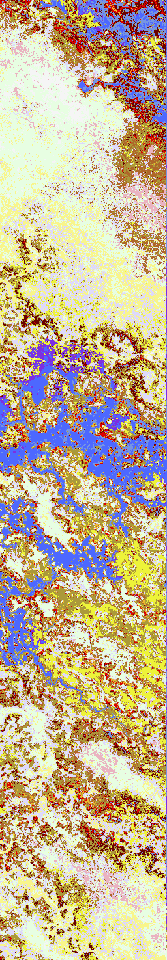}
	\caption{Raw SSS}
\end{subfigure}
\begin{subfigure}{\linewidth}
	\centering
 	\includegraphics[angle=90,width=.99\linewidth]{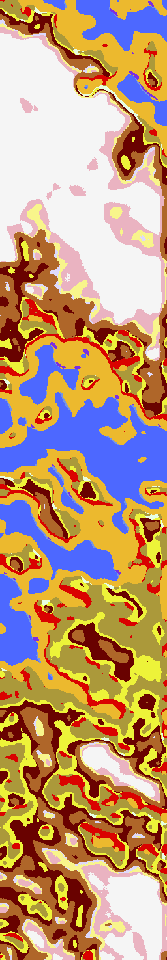}
	\caption{DFFN SSS}
	\label{fig:Botsground_truth}
\end{subfigure}
\begin{subfigure}{\linewidth}
	\centering
	\includegraphics[angle=90,width=.99\linewidth]{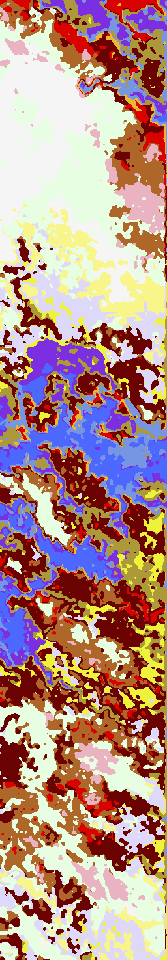}
	\caption{EAP SSS}
\end{subfigure}
\begin{subfigure}{\linewidth}
	\centering
	\includegraphics[angle=90,width=.99\linewidth]{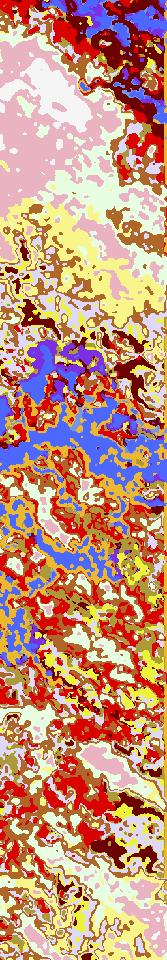}
	\caption{WST SSS}
\end{subfigure}
\begin{subfigure}{\linewidth}
	\centering
	\includegraphics[angle=90,width=.99\linewidth]{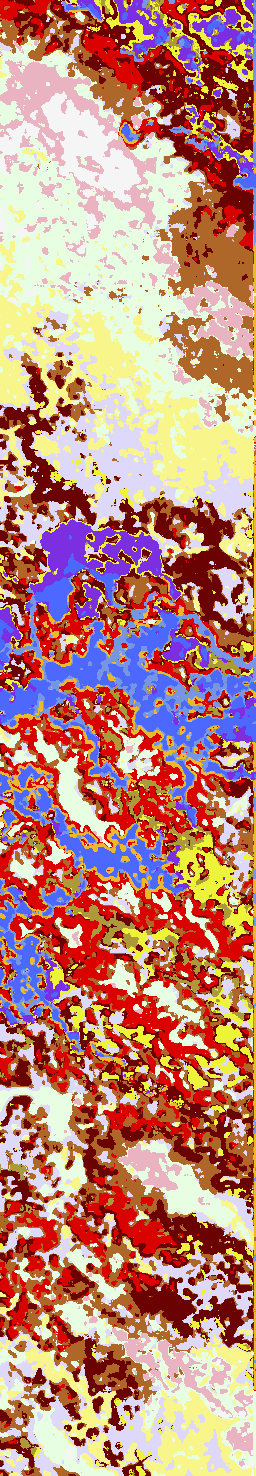}
	\caption{3D Gabor SSS}
\end{subfigure}
\begin{subfigure}{\linewidth}
	\centering
	\includegraphics[angle=90,width=.99\linewidth]{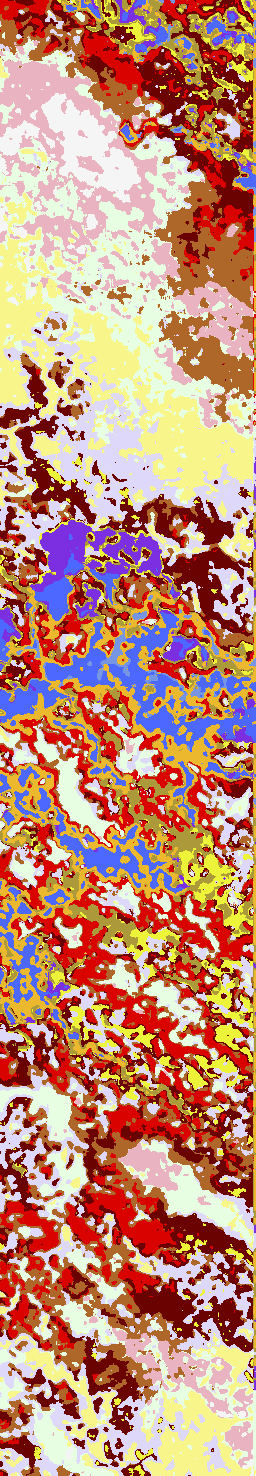}
	\caption{3D FST SSS}
\end{subfigure}
\begin{subfigure}{\linewidth}
	\centering
	\includegraphics[angle=90,width=.99\linewidth]{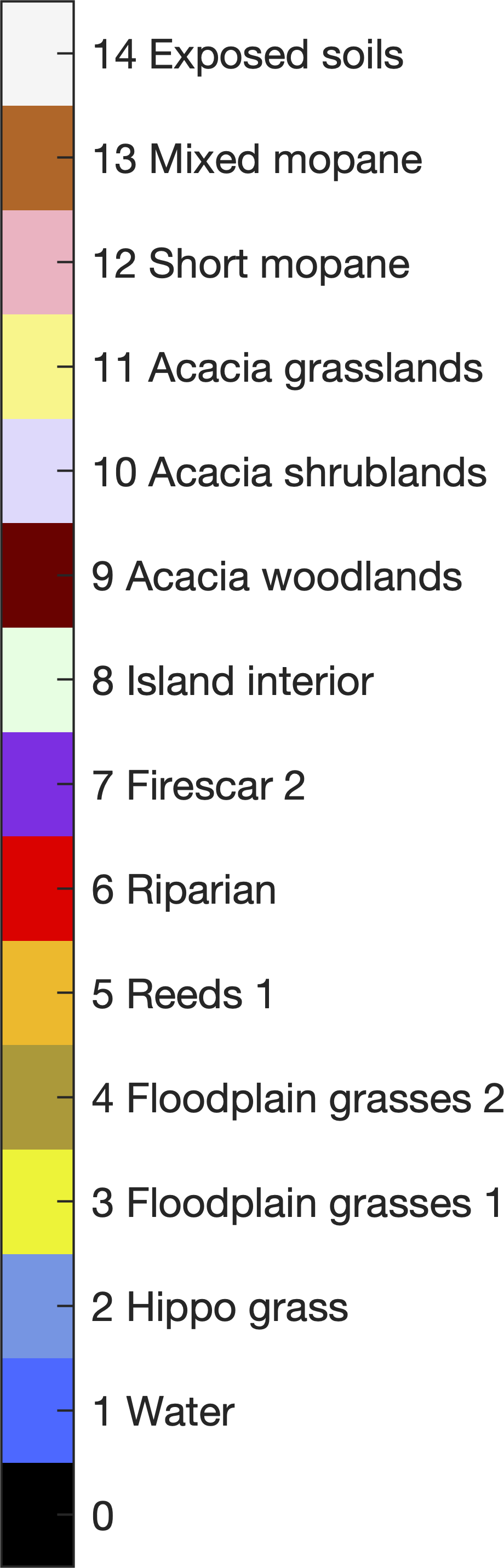}
	\caption{Labels}
\end{subfigure}
\end{minipage}

\caption{Full classification on Botswana with 20 samples per class. This dataset has labels for only 0.86\% of all pixels and is the hardest to interpret of all our datasets, but both these attributes have the potential to reveal spatial bias in classification methods. We see this is the case for DFFN and EAP, where compared to the false color image, delicate features near the central water like islands is blurred away.}
\label{fig:Botswana_full_class}
\end{figure*}

\begin{figure*}[ht!]
\centering
\begin{subfigure}[b]{.24\textwidth}
	\centering
	\includegraphics[width=.99\linewidth]{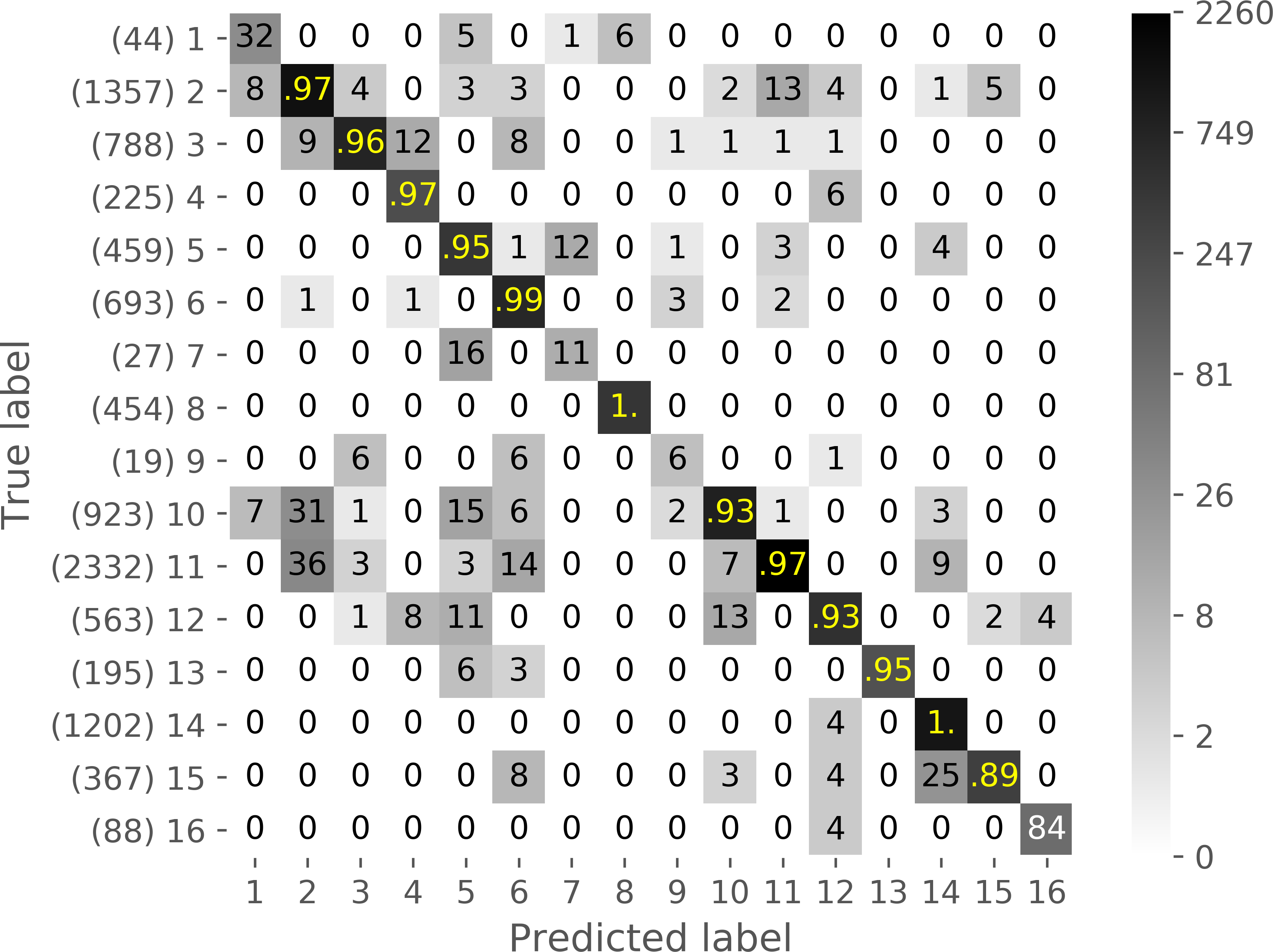}
	\caption{Indian Pines 3D FST}
\end{subfigure}~
\begin{subfigure}[b]{.24\textwidth}
	\centering
	\includegraphics[width=.99\linewidth]{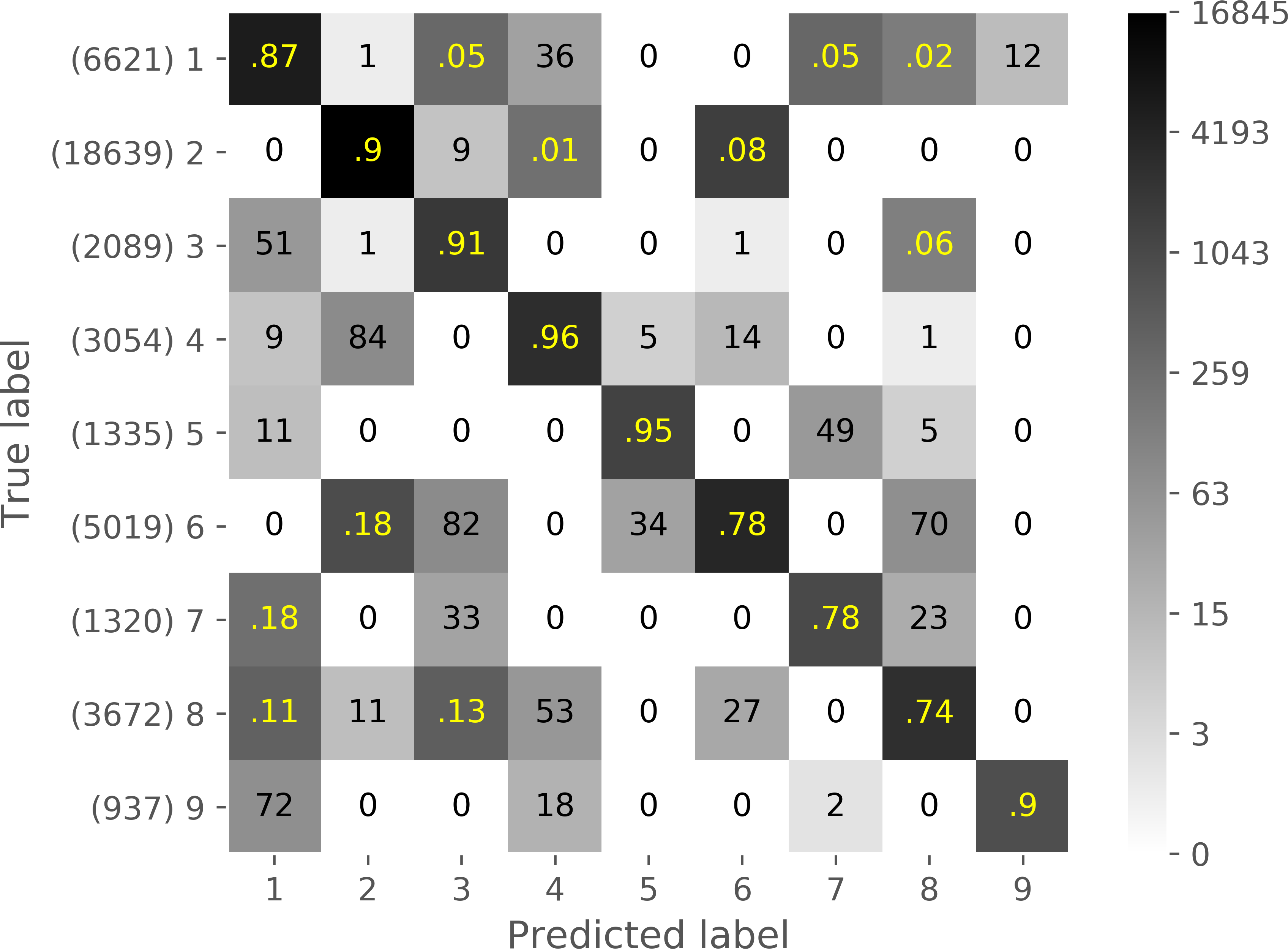}
	\caption{PaviaU 3D FST}
\end{subfigure}~
\begin{subfigure}[b]{.24\textwidth}
	\centering
	\includegraphics[width=.99\linewidth]{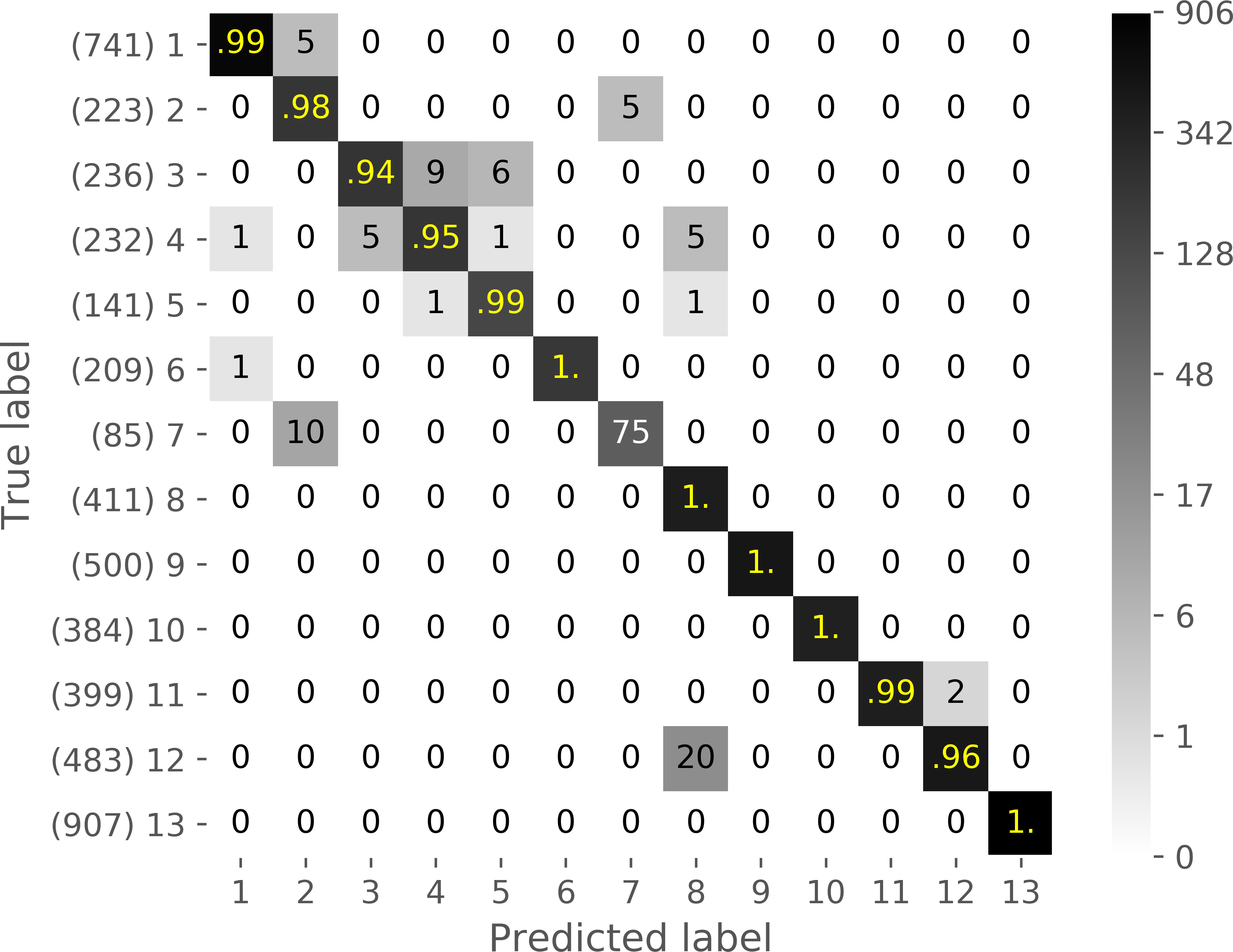}
	\caption{KSC 3D FST}
\end{subfigure}~
\begin{subfigure}[b]{.24\textwidth}
	\centering
	\includegraphics[width=.99\linewidth]{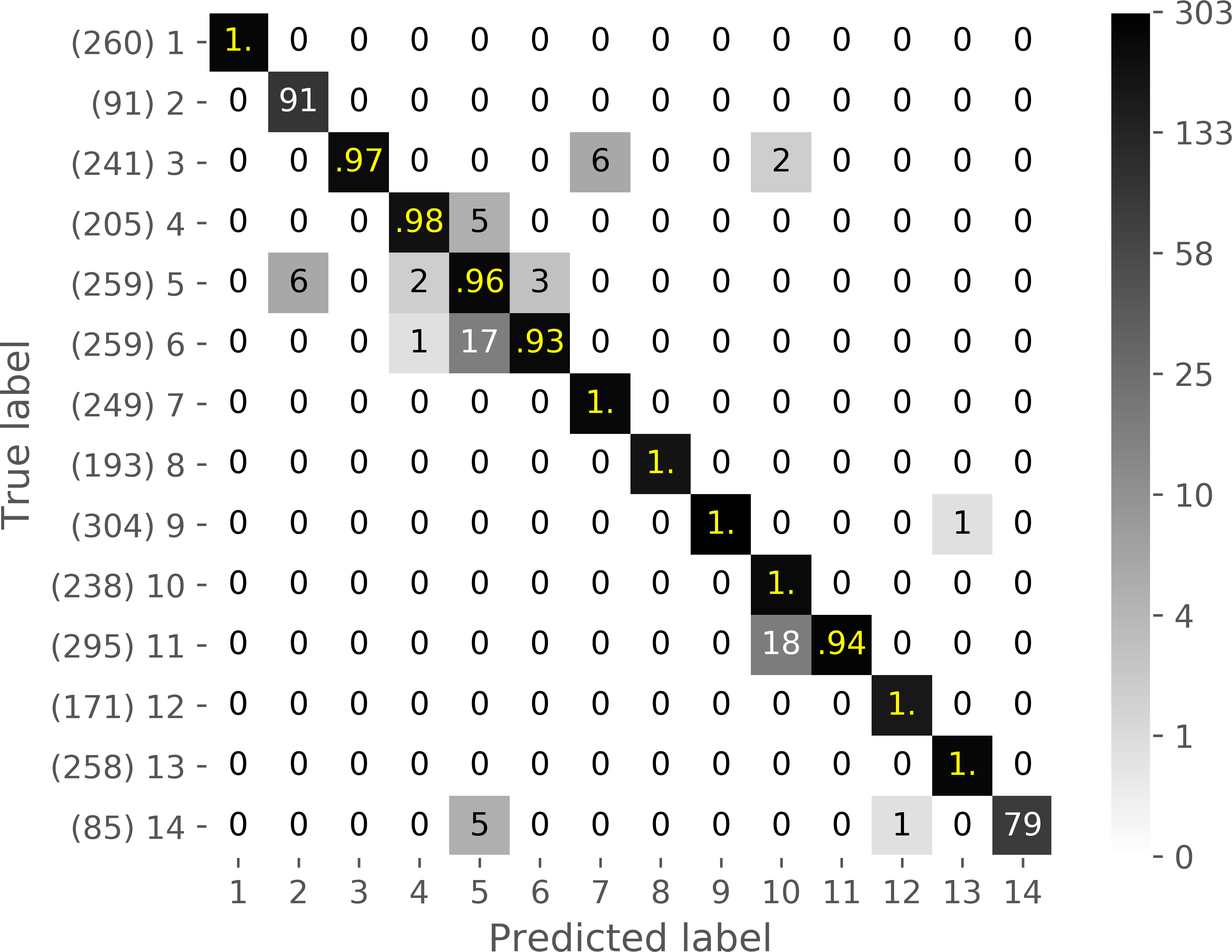}
	\caption{Botswana 3D FST}
\end{subfigure}
\hfill
\begin{subfigure}[b]{.24\textwidth}
	\centering
	\includegraphics[width=.99\linewidth]{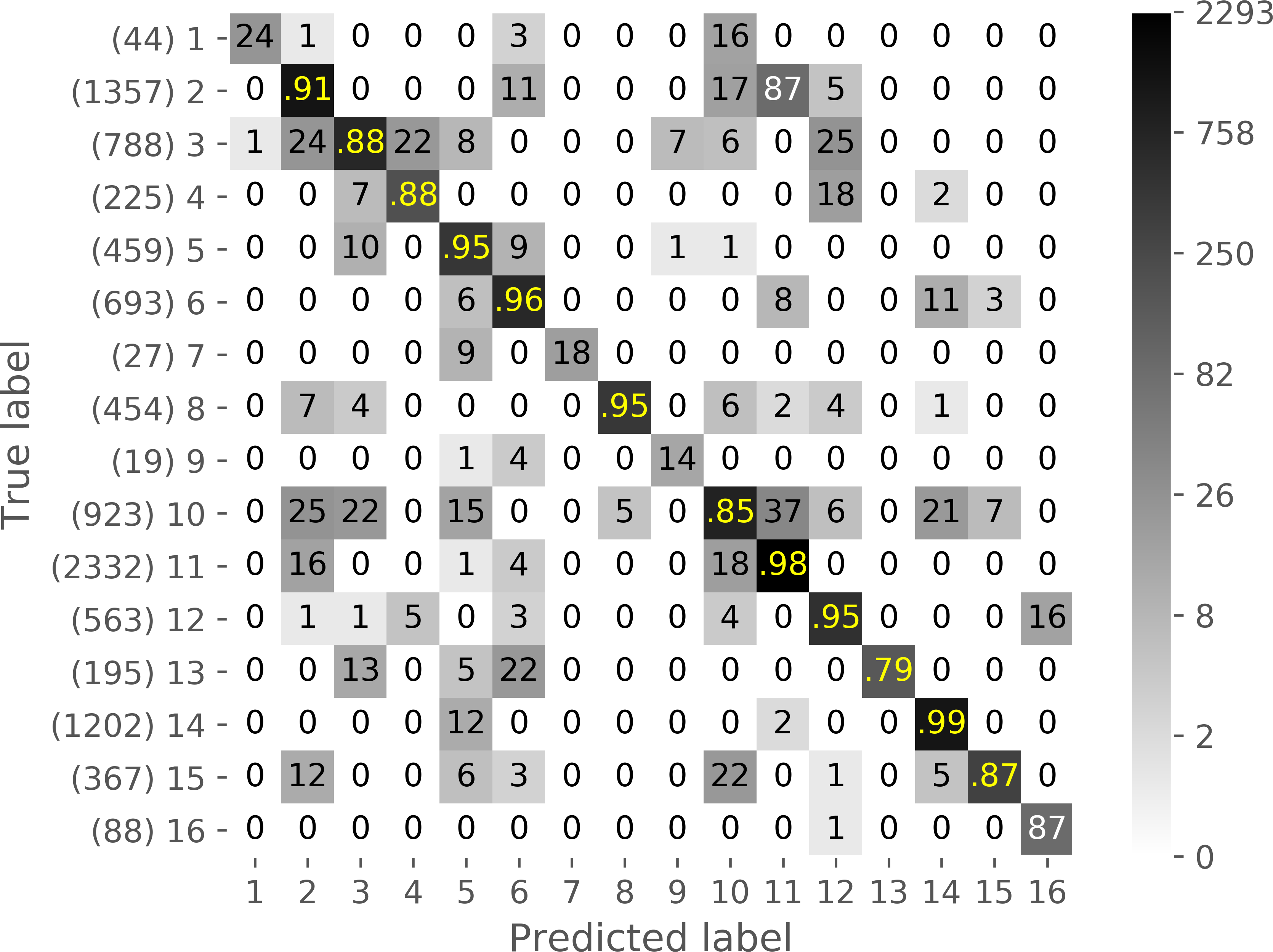}
	\caption{Indian Pines DFFN}
\end{subfigure}~
\begin{subfigure}[b]{.24\textwidth}
	\centering
	\includegraphics[width=.99\linewidth]{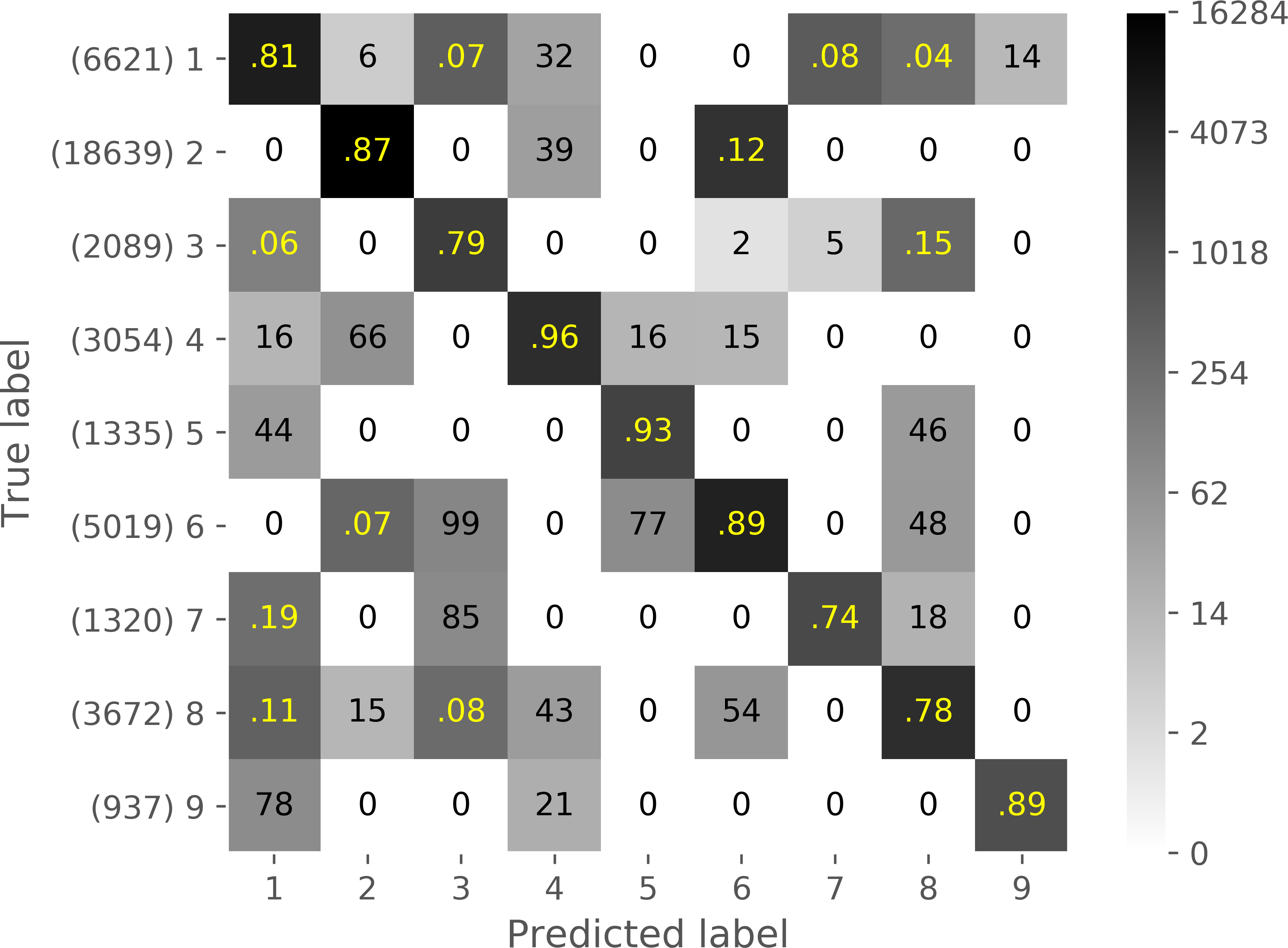}
	\caption{PaviaU WST}
\end{subfigure}~
\begin{subfigure}[b]{.24\textwidth}
	\centering
	\includegraphics[width=.99\linewidth]{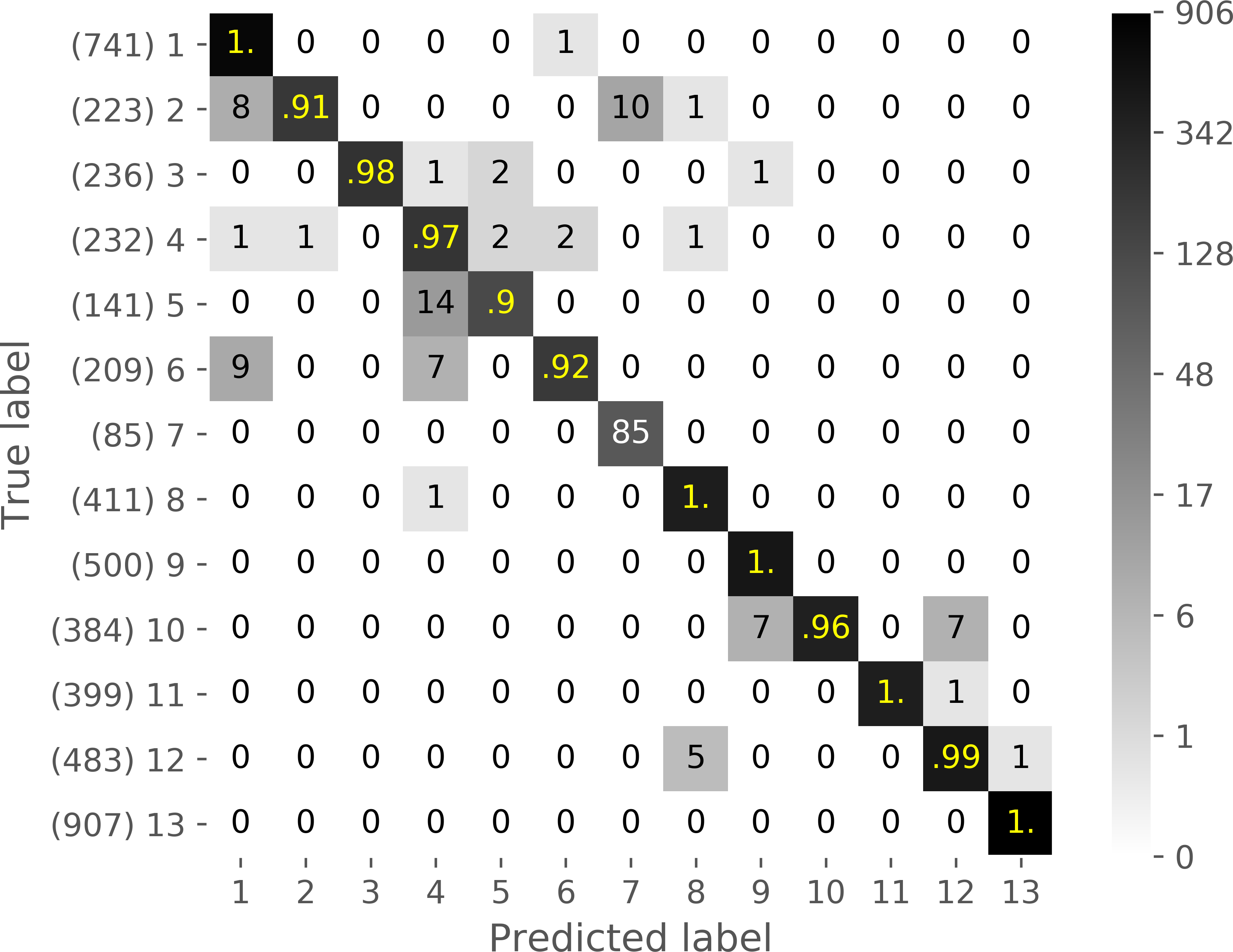}
	\caption{KSC WST}
\end{subfigure}~
\begin{subfigure}[b]{.24\textwidth}
	\centering
	\includegraphics[width=.99\linewidth]{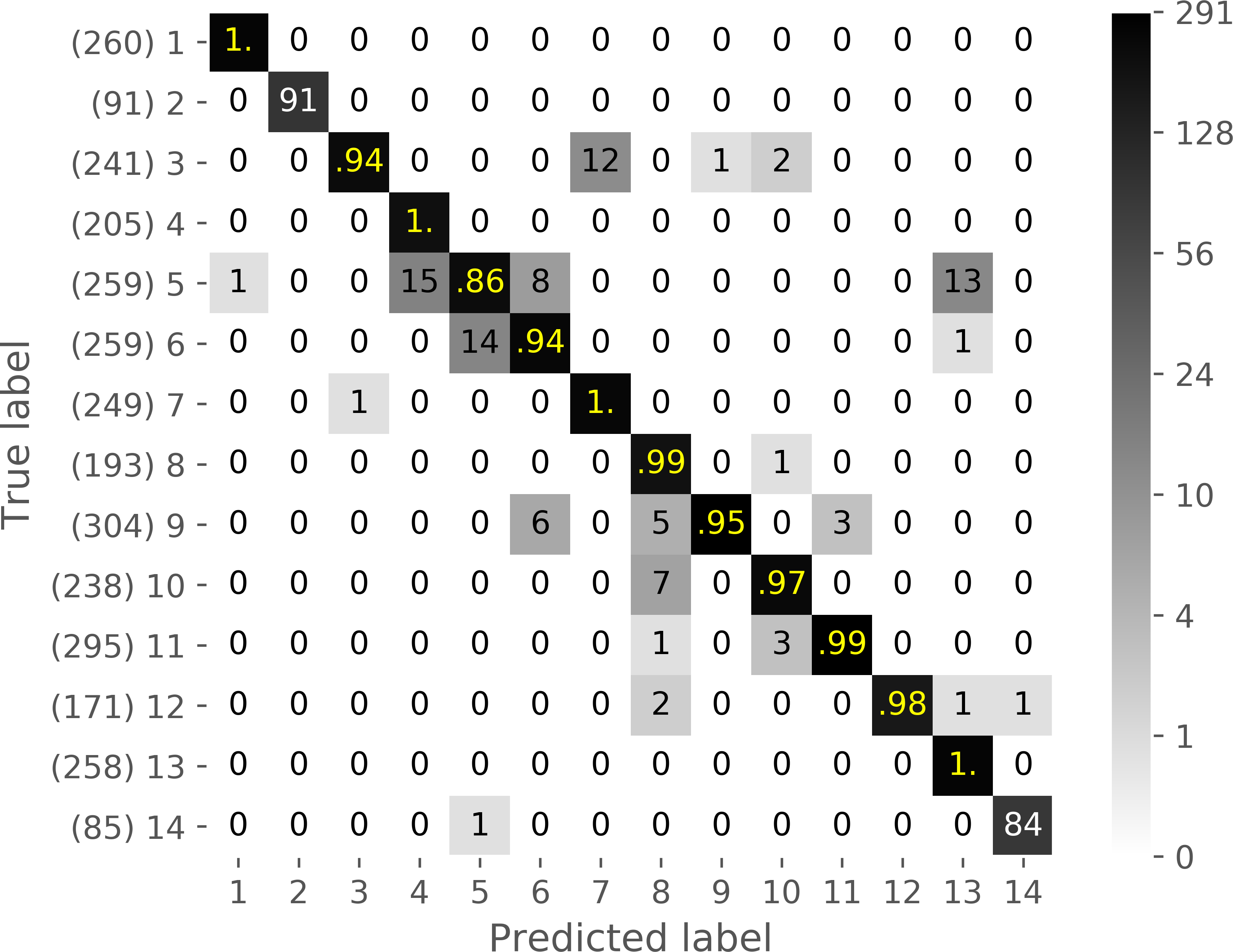}
	\caption{Botswana Gabor}
\end{subfigure}
\hfill
\label{fig:allConfusion}
\caption{Confusion matrices. In parentheses on the y-axis is the total number of samples per class in the test set. The number in the $(i,j)$-th cell is the number of pixels predicted to be class $j$ when the true label is class $i$, when this number is greater than 99 it is instead expressed as a percentage of all the pixels in that class in the test set and is colored yellow. The colorbar is on a log scale. For Indian Pines 5\% of training data was used, for PaviaU 10 samples, for KSC 20 samples, and for Botswana 10 samples. All confusion matrices are made from the best performing trial of the two best methods at the stated amount of randomly distributed training data.}
\end{figure*}

The SSS scenario is much more challenging as evident when comparing the performance of various methods on PaviaU at 90 samples per class (roughly 2\% of training data) in \Cref{fig:Paviaperf}.
The performance in the SSS scenario varies much more wildly between training sets, even though the number of trials performed stays the same as with the randomly sampled datasets.
We see that the deep learning methods DFFN and EAP are most affected by the change from a distributed randomly sampled training set to a single site training set. EAP stops correctly classifying the lower parking lot of PaviaU, and the shapes in DFFN become more distorted. 3D FST is the least affected by the shift to a SSS training set. The features in the red, yellow, and blue boxes of \Cref{fig:PUcolor} change only slightly, and the thin asphalt roads are maintained.
No single method edges out ahead by much consistently for the PaviaU SSS datasets. We again see that performance does not increase by much as the size of the single site grows.
The greatest misclassification error comes from missing large components like the bare soil block in the middle and the continuous meadows.

For KSC and Botswana the SSS sampling method mirrors the structure of the ground truth much more closely (compare the ground truths in \Cref{fig:KSCgt,fig:Botsground_truth} with the training set in \Cref{fig:PUsss}).
In effect when we create our training sets for KSC and Botswana we slice away a piece of just one of the connected components per class in the ground truth.
Perhaps for this reason for KSC and Botswana we see a slightly clearer margin of better performance for 3D FST. 
The simple SVM on the Raw spectrum remains competitive, however the smoothing of 3D FST and 3D Gabor are advantageous over the raw SVM for Botswana.
For KSC we see among the methods a variety of different levels of smoothing around the blue rectangle of \Cref{fig:KSCcolor}, where the separation between mud flats and water is ambiguous.
DFFN with its large receptive field blurs the bridge shapes, and EAP mostly erases the islands in the middle body of water.

It is interesting to see how the extra layers of 3D FST add more detail into the 3D Gabor images for KSC and Botswana for both distributed and SSS training.
For KSC we can see more articulation in the SSS dataset especially.
For Botswana we see a reed border added to the main body of water.
But in general the structure of the ground truth besides the water is especially hard to tell from the false color images of KSC and Botswana, and all but the deepest DFFN give abstract results that look appealing but have a different amount of smoothness.

Spatial smoothing in our proposed method is largely influenced by the choices of $M,M',M''$.
The amount of smoothing our grid search found for FST is on the whole less than other state of the art methods for distributed data, even though the spatial receptive field is similar to the literature.
For SSS data, the spatial smoothing suggested by our gridsearch is at a near minimum, and occasionally edges above the performance of Raw features for this challenging setting.

Finally we look at metrics beyond overall classification scores. \Cref{fig:allConfusion} shows confusion matrices on our proposed method, and a runner up method on all four datasets at a median amount of data. We see no deviation in any conclusions drawn from overall accuracies since the performance is so high in this setting. No small classes suffer a great amount. Looking across the diagonal of per class accuracies between the two most competitive methods we see a great deal of similarity of performance, though on the aggregate 3D FST inches ahead.

\subsection{Computational Cost}
\label{sec:perf}

\begin{table}[!t]
\renewcommand{\arraystretch}{1.3}
\caption{Feature extraction and SVM computation time on the whole image.}
\centering
\begin{tabular}{|l|l|l|l|l|}
\hline
                & IP       & PaviaU   & KSC   & Botswana       \\ \hline
3D FST Feat. (s total) & 9    & 23    & 75    & 71         \\ \hline
3D FST Feat. (pixels/s) & 2,300    & 9,000    & 4,100    & 5,300         \\ \hline
SVM for 3D FST (s total)     & 5 & 41 & 40 & 90   \\ \hline
SVM for 3D FST (pixels/s)     & 4,200 & 5,000 & 7,800 & 4,200   \\ \hline
\end{tabular}
\label{tab:perf}
\end{table}

The runtime performance of our 3D FST is in \cref{tab:perf}.
In our implementation we classified one patch of pixels ($51\times51$) at a time, which greatly accelerated processing compared to processing one pixel at a time. The average rate that we extracted features from the 4 datasets tested was 5,200 pixels per second. In the classification setting, after all the pixels were processed, a linear SVM was trained, and then the test samples were classified. The average rate that we classified pixels across the 4 datasets was 5,300 pixels per second. The SVM performance numbers in \cref{tab:perf} are computed from the test SVM, and the training was always faster (by at least one order of magnitude) than the testing of the SVM.
The major factor in performance was not the filter size directly, but the number of filters used in total, which is influenced by our windows size and downsampling choices.
In our experiments the number of filters per layer was proportional to the product of the size of the supports of the fiters in each dimension, for example in the first layer the number of filters was $M_1\times M_2\times M_3$.
Since the SVM did not take much time total, we saw no need to run PCA or any other form of dimension reduction reduction between feature extraction and classification, though it could have reduced memory usage by throwing away some coefficients.
Needless to say the supervised training of the neural networks of EAP and DFFN took much longer to train than the feature extraction of 3D FST and 3D WST.
Supervised neural networks also need to be retrained for different training/validation/testing set partitions, while the 3D FST of an HSI cube can be saved once for each dataset, and sampled later.
Our downsampling strategy for 3D FST reduced the feature size by about 20x in comparison to the original 3D WST in \cite{tang2015hyperspectral}.
All our methods were performed on a NVidia Titan X GPU with 12 GB of memory.

\section{Conclusion}
\label{sec:conc}

In this paper we have proposed a three-dimensional Fourier scattering transform for HSI classification.
This method has the neural network like benefits of hierarchical feature extraction while bypassing the training process which is computationally expensive in both the amount of required training data and training time.
Our three dimensional time-frequency features are well suited for HSI data since they decompose the HSI into multi-frequency bands and remove small perturbations such as noise.
The 3D FST is particularly effective when there is limited training data.
As supported by the experimental results, 3D FST achieved SoA performance on benchmark datasets, all while executing within a few minutes on a conventional GPU, and using a simple linear SVM for classification.

An advantage of our method is its compatibility with conventional deep learning implementations. This readily allows for a shift from the pre-processing based classification with a linear SVM we presented to an end-to-end feature extraction with a classification deep network.
This has the potential to improve classification performance further as both the classification and feature filters will be learned for each dataset, and opens the door for integrating our method with other deep learning techniques like transfer learning or meta learning. 
Our future work investigates this hybridization of scattering transforms with deep learning where the classification is performed with a neural network following a tune-able scattering transform that serves as a feature extractor, and both are trained jointly.
We also leave to future work a task even more challenging than classification from limited data: classification when training data is only available from other datasets. In this case the 3D FST and the classifier will be adapted to a new dataset where training data may not be available at all. 



\section*{Acknowledgment}

Ilya Kavalerov and Rama Chellappa acknowledge the support of the MURI from the Army Research Office under the Grant No. W911NF-17-1-0304. This is part of the collaboration between US DOD, UK MOD and UK Engineering and Physical Research Council (EPSRC) under the Multidisciplinary University Research Initiative.
Weilin Li was supported by the Ann G. Wylie Dissertation Fellowship during the Spring 2018 semester at the University of Maryland and by the James C. Alexander Prize for Graduate Research.
Wojciech Czaja is supported in part by LTS through Maryland Procurement Office and the NSF DMS 1738003 grant.

\ifCLASSOPTIONcaptionsoff
  \newpage
\fi



%

\bibliography{main} 
\bibliographystyle{IEEEtran}

%



\vskip 0pt plus -1fil

\begin{IEEEbiography}[{\includegraphics[width=1in,height=1.25in,clip,keepaspectratio]{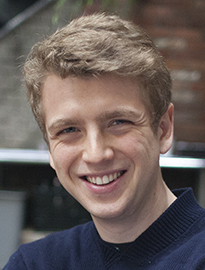}}]{Ilya Kavalerov}
received the B.Sc. degree in Chemistry at the George Washington University in Washington D.C. in 2013. He is currently pursuing the Ph.D. degree in Electrical Engineering at the University of Maryland, College Park.
He was a research intern at Google Cambridge in 2018, and Google Mountain View in 2019.
His research interests include scattering transforms, deep learning, generative adversarial networks, and computer vision.
\end{IEEEbiography}

\vskip 0pt plus -1fil

\begin{IEEEbiography}[{\includegraphics[width=1in,height=1.25in,clip,keepaspectratio]{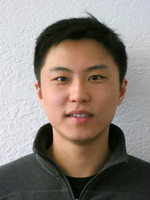}}]{Weilin Li}
received the Ph.D. degree in mathematics from the University of Maryland, College Park, in 2018. He is currently a Courant Instructor with the Courant Institute of Mathematical Sciences at New York University.
His research interests are harmonic analysis, signal processing, and machine learning. 
\end{IEEEbiography}

\vskip 0pt plus -1fil


\begin{IEEEbiographynophoto}{Wojciech Czaja}
received the Ph.D. degree in Mathematics from Washington University in St. Louis. He is currently a Professor of Mathematics at University of Maryland College Park, a member of the Norbert Wiener Center, of the Center for Scientific Computation and  Mathematical Modeling, and a Marie Curie Fellow.
His research interests range from theoretical harmonic analysis and representation theory, to applications of mathematics in data science and machine learning.
\end{IEEEbiographynophoto}

\vskip 0pt plus -1fil

\begin{IEEEbiography}[{\includegraphics[width=1in,height=1.25in,clip,keepaspectratio]{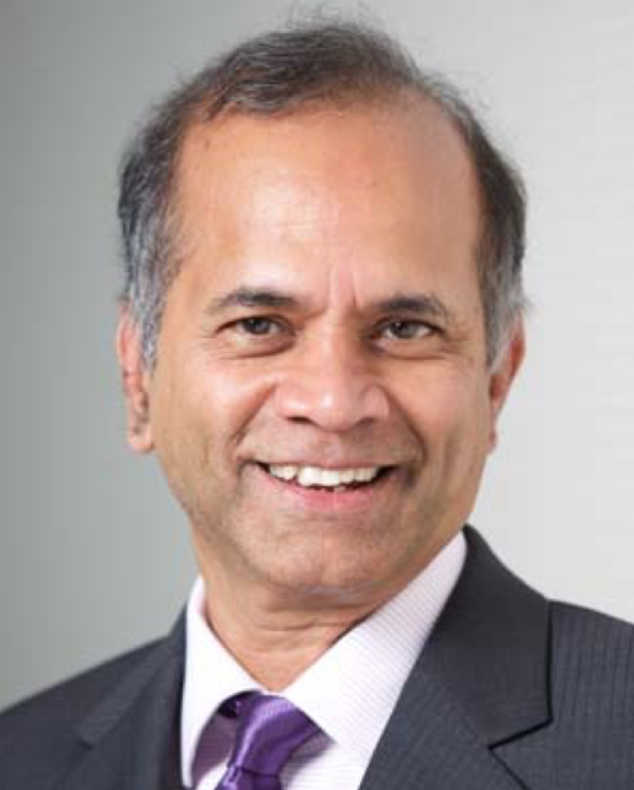}}]{Rama Chellappa}
(F’92) received the Ph.D. degree from Purdue University in 1981. He is a Bloomberg Distinguished Professor with the Department of Biomedical Engineering with a joint appointment in the Department of Electrical and Computer Engineering Department at Johns Hopkins University. His current research interests span computer vision, pattern recognition, machine learning and artificial intelligence. He has received numerous research, teaching, innovation, and service awards from UMD,
IEEE, IAPR, and IBM. He is a fellow of AAAI, AAAS, ACM, IAPR, and OSA.
\end{IEEEbiography}




\end{document}